%% file: COSTCO_R2_arXiv.tex
\documentclass[11pt]{article}
\usepackage{enumerate,natbib,mathrsfs,amsthm,amsmath,bm,amssymb,listings,setspace}
\usepackage[OT1]{fontenc}
\usepackage[colorlinks,linkcolor=red,anchorcolor=blue,citecolor=blue]{hyperref}
\usepackage{fullpage}
\usepackage{tabularx}
\usepackage{tikz}
\def\checkmark{\tikz\fill[scale=0.5](0,.35) -- (.25,0) -- (1,.7) -- (.25,.15) -- cycle;}                                                                                                                                    
\usepackage{array}
\newcolumntype{P}[1]{>{\centering\arraybackslash}p{#1}}
\usepackage{array}
\usepackage[protrusion=false,expansion=true]{microtype}
\usepackage{algorithm,algorithmic,graphicx}
\usepackage{array,multirow,graphicx}
\usepackage{float}
\theoremstyle{plain} 
\input{def-smile}

\def\change{\textcolor{black}}

\def\red{\textcolor{black}}

\usepackage{multirow}
\usepackage{graphicx}
\usepackage{caption}
\usepackage{subcaption}
\usepackage{enumerate}% http://ctan.org/pkg/enumerate

\newcommand{\Ten}{\mathcal{T}}
\newcommand{\Er}{\mathcal{E}}
\newcommand{\Mat}{\mathbf M}
\newcommand{\Am}{\mathbf A}
\newcommand{\Um}{\mathbf U}
\newcommand{\I}{\mathbf I}
\newcommand{\Bm}{\mathbf B}
\newcommand{\Cm}{\mathbf C}
\newcommand{\Dm}{\mathbf D}
\newcommand{\Em}{\mathbf E}
\newcommand{\Fm}{\mathbf F}
\newcommand{\Gm}{\mathbf G}
\newcommand{\Hm}{\mathbf H}
\newcommand{\Jm}{\mathbf J}
\newcommand{\Pm}{\mathbf P}

\newcommand{\ac}{\mathbf a}
\newcommand{\bc}{\mathbf b}
\newcommand{\cc}{\mathbf c}
\newcommand{\dc}{\mathbf d}
\newcommand{\uc}{\mathbf u}

\newcommand{\wc}{\mathbf w}

\newcommand{\vc}{\mathbf v}
\newcommand{\ace}{a}

\newcommand{\bce}{b}
\newcommand{\cce}{c}
\newcommand{\Vm}{\mathbf V}
\newcommand{\Rbb}{\mathbb R}

\newcommand{\bks}{\symbol{92}}

\newcommand{\Tfrac}[2]{%
  \ooalign{%
    $\genfrac{}{}{1.2pt}1{#1}{#2}$\cr%
    $\color{white}\genfrac{}{}{0.6pt}1{\phantom{#1}}{\phantom{#2}}$}%
}

\title{\Large{\textbf{Covariate-assisted Sparse Tensor Completion}}}
\author{
\bigskip
 Hilda S Ibriga and Will Wei Sun \\ 
\normalsize{Purdue University}
}
\date{}

\begin{document}

\maketitle

\begin{footnotetext}
{Hilda S Ibriga is a Research Scientist at Eli Lilly and Company. Most of work was done when Hilda S Ibriga was a PhD student at Purdue University. Will Wei Sun is an Assistant Professor of Krannert School of Management at Purdue University. Contact email: sun244@purdue.edu}
\end{footnotetext}

\begin{abstract}
\baselineskip=18pt
We aim to provably complete a sparse and highly-missing tensor in the presence of covariate information along tensor modes. Our motivation comes from online advertising where users' click-through-rates (CTR) on ads over various devices form a CTR tensor that has about $96\%$ missing entries and has many zeros on non-missing entries, which makes the standalone tensor completion method unsatisfactory. Beside the CTR tensor, additional ad features or user characteristics are often available. In this paper, we propose Covariate-assisted Sparse Tensor Completion (\texttt{COSTCO}) to incorporate covariate information for the recovery of the sparse tensor. The key idea is to jointly extract latent components from both the tensor and the covariate matrix to learn a synthetic representation. Theoretically, we derive the error bound for the recovered tensor components and explicitly quantify the improvements on both the reveal probability condition and the tensor recovery accuracy due to covariates. Finally, we apply \texttt{COSTCO} to an advertisement dataset consisting of a CTR tensor and ad covariate matrix, leading to $23\%$ accuracy improvement over the baseline. An important by-product is that ad latent components from \texttt{COSTCO} reveal interesting ad clusters, which are useful for better ad targeting. 
\end{abstract}


\newpage
\baselineskip=23pt

\section{Introduction}
\label{sec:introduction}

Low-rank tensor completion aims to impute missing entries of a partially observed tensor by forming a low-rank decomposition on the observed entries. It has been widely used in various scientific and business applications, including recommender systems \citep{Symeonidis2008}, neuroimaging analysis \citep{zhou2013tensor}, signal processing \citep{Sidiropoulos2017}, social network analysis \citep{jing2020community}, personalized medicine \citep{Wang2019}, and time series analysis \citep{chen2019factor}. We refer to the recent surveys on tensors for more real applications \citep{Song2019, bi2020tensors}. In spite of its popularity, it is also well known that when the missing percentage of the tensor is very high, a standalone tensor completion method often fails at yielding desirable recovery results. Fortunately, in many real applications, we also have access to some side covariate information. In this paper, we aim to complete a sparse and highly-missing tensor in the presence of covariate information along tensor modes.  

\begin{figure}[h!]
        \centering
    \begin{subfigure}[t]{0.3\textwidth}
        \centering
        A. Sparse CTR tensor
        \includegraphics[width=\linewidth]{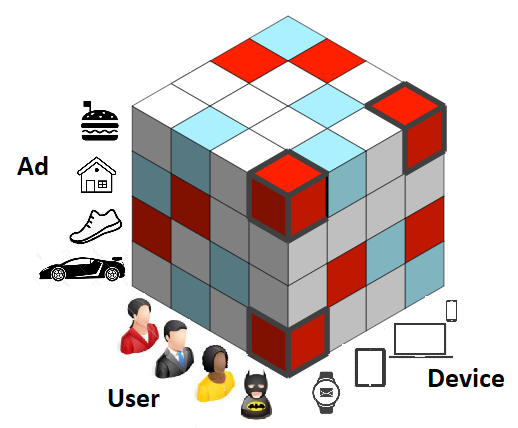} 
         \label{fig:sparse_ctr}
    \end{subfigure}
    ~~~~~~%\hfill
    \begin{subfigure}[t]{0.40\textwidth}
        \centering
         B. Coupled sparse CTR tensor
        \includegraphics[width=\linewidth]{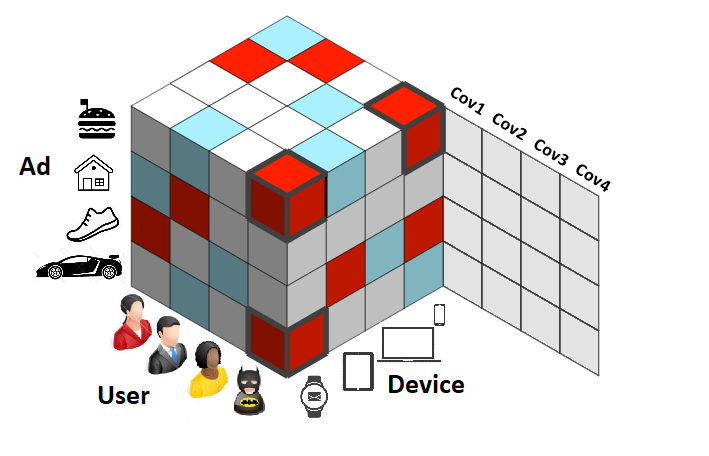} 
         \label{fig:tensor_matrix}
    \end{subfigure}
        \caption{A. sparse (user $\times$ ad $\times$ device) CTR tensor with missing entries; B. sparse CTR tensor with missing entries coupled with matrix of ad covariates. The red cells represent missing entries; blue cells represent zeros, grey cells represent non-zero entries.}
         \label{fig:p_n2} 
\end{figure}

Our motivation originates from online advertising application, where advertisement (ad) information is usually described by both users' click behavior data and ad characteristics data. More formally, the users’ click data refer to as the click-through rate (CTR) of the ads, quantifying the user click behavior on different ads, various platforms, different devices or over time etc. The CTR data is therefore often represented as a tensor of three or four modes, e.g., the user $\times$ ad $\times$ device tensor shown in Figure \ref{fig:p_n2}. The ad characteristic data on the other hand is usually represented in the form of a matrix which contains context information for each ad. Typically in online advertising not all users are presented with all ads, thus creating many missing data in the CTR tensor. Moreover, users typically engage with a small subset of the ads that are presented to them. Low rates of ads engagement is a common phenomenon in online advertising which begets a highly sparse CTR tensor (many zero entries) with high percentage of missing entries. For instance, in our real data shown in Section \ref{sec:realdata}, the ad CTR tensor has $96 \%$ missing entries and is highly sparse with only $40\%$ of the revealed entries being nonzero. We show in Sections \ref{sec:simul} and \ref{sec:realdata} that methods using a standalone tensor completion often fail at recovering the missing entries of a tensor with such missing percentage. On the contrary the ad characteristic data is usually relatively complete and dense. It therefore becomes advantageous to incorporate the ad characteristic information in a model to recover the missing entries of the CTR tensor. The structure of the sparse CTR tensor with missing entries coupled with the ad characteristic data is illustrated in Figure \ref{fig:p_n2}. As shown in Figure \ref{fig:p_n2} the two sources of data; CTR tensor and ad covariates matrix are coupled along the ad mode. 

%%%%%%%%%%%%%%%%%%%%%%%%%%%%%%%
\vspace{-0.5em} 

In this article, we propose Covariate-assisted Sparse Tensor Completion (\texttt{COSTCO}) to recover missing entries in highly sparse tensor with a large percentage of missing entries. Under the low-rank assumption on both the tensor and the covariate matrix, we assume the latent components corresponding to the coupled mode are shared by both the tensor and matrix decomposition. This model encourages a synthetic representation of the coupled mode by leveraging the additional covariate information into tensor completion. \red{Another advantage of our \texttt{COSTCO} is that it naturally handles the cold-start problem. For a new ad, the CTR tensor itself provides no information to estimate the corresponding CTR entries. Hence, existing standalone tensor completion based methods are not directly applicable. In contrast, our \texttt{COSTCO} solves this issue by incorporating additional ad covariate information. The intuition behind it is that the ad covariate matrix provides a reasonable cluster structure of ads. Therefore, the missing clicking behaviors on a new ad can be learnt from the shared latent components estimated based on both the CTR tensor and the ad covariate matrix. Similarly, the cold-start problem can be addressed for a new user when we have a user covariate matrix.} In algorithm, we formulate the parameter estimation as a non-convex optimization with sparsity constraints, and propose an efficient sparse alternating least squares approach with an extra refinement step. Our algorithm jointly extracts latent features from both tensor and the covariate matrix and uses covariate information to improve the recovery accuracy of the recovered tensor components. We showcase through extensive numerical studies that our \texttt{COSTCO} is able to successfully recover entries for a tensor even with $98\%$ missing entries.

In addition to the above methodological contributions, we also make theoretical contributions to the understanding of how side covariate information affects the performance of tensor completion. In particular, we derive the non-asymptotic error bound for the recovered tensor components and explicitly quantify the improvements on both the reveal probability condition and the tensor recovery accuracy due to additional covariate information. We show that \texttt{COSTCO} allows for a relaxation on the lower bound of the reveal probability $p$ compared to that required in tensor completion with no covariates, see Assumption 6 for details. In the extreme case where all tensor modes are coupled with covariate matrices, we can still recover the tensor entries even when the reveal probability of the tensor is close to zero. Moreover, we present the statistical errors for the shared tensor component (corresponding to the coupled mode) and non-shared tensor components separately to demonstrate the gain brought in through the coupling of covariates information in the model. We show that given some mild assumptions on noise levels and condition numbers, our \texttt{COSTCO} guarantees an improved recovery accuracy for the shared component. Unlike existing theoretical analysis on low-rank tensors which assumes the error tensor to be Gaussian, we do not impose any distributional assumption on the error tensor or the error matrix. Our theoretical results depends on the error term only through its sparse spectral norm.

Finally, we apply \texttt{COSTCO} to the advertising data from a major internet company to demonstrate its practical advantages. \texttt{COSTCO} makes use of both ad CTR tensor and ad covariate matrix to extract the latent component which leads to $23\%$ accuracy improvement in recovering the missing entries when compared to the standalone sparse tensor completion and $10\%$ improvement over a covariate-assisted deep learning algorithm. Moreover, an important by-product from our \texttt{COSTCO} is to use the recovered ad latent components for better ad clustering. Ad clustering is an essential task for targeted advertising that helps lead useful ad recommendation for online platform users. Cluster analysis on our ad latent components reveals interesting and new clusters that link different product industries which are not formed in existing clustering methods. Such findings could directly help the marketing team to strategize the ad planing procedure accordingly for better ad targeting.

\subsection{Related work and paper organization}

\textbf{Tensor completion with side information:} The simultaneous extraction of latent information from multiple sources of data can be interpreted as a form of data fusion \citep{acar2011all, acar2013understanding,zhou2017tensor, Wilma2018, Choi2019, Huyan2020, li2020generalized}. Among them, there are a few work related to tensor completion with side information. The most related work to our approach is the gradient-based all-at-once optimization method proposed by \citet{acar2011all} which updates the matrix and tensor components all at once. We compare it in our experiments and find that it is consistently inferior to our \texttt{COSTCO}. \citet{zhou2017tensor} proposed a Riemannian conjugate gradient descent algorithm to solve the tensor completion problem in the presence of side information. However, this procedure does not address the tensor completion problem in the presence of high percentage of missing entries combined with a high sparsity level. \citet{Choi2019} developed a fast and scalable algorithm for the estimation of shared latent features in coupled tensor matrix model. However, their approach does not allow missing entries and only works for complete data. Importantly, all the aforementioned works did not provide any theoretical analysis for their methods. \citet{Wilma2018} proposed a convex coupled tensor-matrix completion method and \citet{Huyan2020} applied the tensor ring decomposition method on the coupled tensor-tensor problem. However, these two works do not account for noise in the tensor or matrix, i.e., their model is noiseless, nor do they consider the sparse tensor case. To the best of our knowledge, our work is the first provably method that is tailored for completing a highly sparse and highly missing tensor in the presence of covariate information.

\textbf{Tensor completion with theoretical guarantees:} Our theoretical analysis is related to a list of recent theoretical work in standalone tensor completion that does not incorporate covariate information \citep{jain2014provable, zhang2019cross, cai2021nonconvex, xia2021statistically}. In particular, \citet{jain2014provable} provided recovery guarantee for symmetric and orthogonal tensors with missing entries, but did not explore recovery for the tensor completion with coupled covariates nor did they address the case of the non-orthogonal, noisy and sparse tensor. \cite{zhang2019cross} established a sharp recovery error for a special tensor completion problem, where the missing pattern was not uniformly missing but followed a cross structure. \citet{xia2021statistically} proposed a two-step algorithm (a spectral initialization method followed by the power method) for the noisy Tensor completion case and established the optimal statistical rate in low-rank tensor completion. Different from our model, they assumed the error tensor to be subgaussian and did not consider sparsity in tensor completion. \citet{cai2021nonconvex} also independently proposed a provable two stage algorithm for the noisy tensor completion problem. Importantly, none of the aforementioned work accommodates the inclusion of covariate information in the tensor completion model. The coupled sparse tensor and matrix formulation in our \texttt{COSTCO} poses unique difficulties in the theoretical analysis. The unequal weights of the tensor and matrix prevent us to obtain a close-form solution for the alternative least squares problem compared to the traditional tensor completion. Moreover, the presence of non-orthogonality, general noise, and sparsity in our model introduce additional challenges. These make our theoretical analysis far from a simple extension to the standard tensor completion problem.

\textbf{Paper organization:} The rest of the paper is organized as follows. Section \ref{sec:prelim} reviews some notations, basic definitions of algebra of tensors. Section \ref{sec:meth} presents our model, the optimization problem and our algorithm along with procedures for initialization and parameter tuning. Section \ref{sec:analtheo} presents the main theoretical results. Section \ref{sec:simul} contains a series of simulation studies. Section \ref{sec:realdata} applies our algorithm to an advertisement data set to illustrate its practical advantages. Interesting extensions, all proof details, lemmas and additional experiments are left in the supplemental material. 

%%%%%%%%%%%%%%%%%%%%%%%%%%%%
\section{Notation and Preliminaries}
\label{sec:prelim}
In this section, we introduce some notation, and review some background on tensors. Throughout the paper we denote tensors by Euler script letters, e.g., $\Ten, \Er$. Matrices are denoted by boldface capital letters, e.g., $\Am, \Bm, \Cm$ ; vectors are represented with boldface lowercase letters, e.g., $\ac, \vc$, and scalars are denoted by lowercase letters, e.g., $a, \lambda$. The $ n \times n$ identity matrix $\I_n$ is simply written as $\I$ when the dimension can be easily implied from the context. 

Following \citet{kolda2009tensor}, we use the term tensor to refer to a multidimensional array; a concept that generalizes the notion of matrices and vectors to higher dimensions. A first-order tensor is a vector, a second-order tensor is a matrix and a third-order tensor is a three dimensional array. Each order of a tensor is referred to as a mode. For example a matrix (second-order tensor) has two modes with mode-$1$ and mode-$2$ being the dimensions represented by the rows and columns of the matrix respectively. Let $\Ten \in \Rbb^{n_1 \times n_2 \times n_3}$ be a third-order non-symmetric tensor. We denote its $(i, j, k)$th entry as $\Ten_{ijk}$. A tensor fiber refers to a higher order analogue of matrix row and column and is obtained by fixing all but one of the indices of the tensor. For the tensor $\Ten$ defined above, the mode-$1$ fiber is given by $\Ten_{:jk}$; the mode-$2$ fiber by $\Ten_{i:k}$ and mode-$3$ fiber by $\Ten_{ij:}$. Next the slices of the tensor $\Ten$ are obtained by fixing all but two of the tensor indices. For example the frontal, lateral and horizontal slices of the tensor $\Ten$ as denoted as $\Ten_{::k}$, $\Ten_{:j:}$ and $\Ten_{i::}$. We define three different types of tensor vector products. For vectors $\uc \in \Rbb^{n_1}, \vc \in \Rbb^{n_2},  \wc  \in \Rbb^{n_3}$, the mode-$1$, mode-$2$ and mode-$3$, tensor-vector product is a matrix defined as a combinations of tensor slices: $\Ten \times_1 \uc= \sum_{i=1}^{n_1}{\uc_{i}\Ten_{i::}}$, $\Ten \times_2 \vc= \sum_{j=1}^{n_2}{\vc_{j}\Ten_{:j:}}$, $\Ten \times_3 \wc= \sum_{k=1}^{n_3}{\wc_{k}\Ten_{::k}}.$ The tensor multiplying two vectors along its two modes is a vector defined as: $\Ten \times_2 \vc \times_3 \wc= \sum_{j,k}{\vc_{j} \wc_{k}\Ten_{:jk}}$, $\Ten \times_1 \uc \times_2 \vc= \sum_{i,j}{\uc_{i} \vc_{j}\Ten_{ij:}}$, $\Ten \times_1 \uc \times_3 \wc= \sum_{i,k}{\uc_{i} \wc_{k}\Ten_{i:k}}.$ Finally the tensor-tensor product is a scalar defined as $\Ten \times_1 \uc \times_2 \vc \times_3 \wc= \sum_{i,j,k}{ \uc_{i} \vc_{j} \wc_{k}\Ten_{ijk}}.$

We denote $\|\Mat\|$ and $ \|\Mat\|_{F}$ to be the spectral norm and the Frobenius norm of a matrix $\Mat$, respectively. The spectral norm of a tensor $\Ten$ is defined as
\begin{align}
\label{eq:spectnorm}
     \|\Ten \| := \sup_{\substack{\| \uc \|_2 = \|\vc\|_2 = \|\wc\|_2 = 1}} \Big|{\Ten \times_1 \uc \times_2 \vc \times_3 \wc }\Big|,
\end{align}
and its Frobenius norm is $ \|\Ten\|_F:= \left( \sum_{i,j,k}{ \Ten_{ijk}^2 } \right) ^{1/2}$. Define the sparse spectral norm of a matrix $\Mat$ as $\|\Mat\|_{<d_1>}:= \sup_{\substack{\| \uc \|_2=1, \|\uc\|_0=d_1}} \|{\Mat \times_1 \uc}\|_2$ and the sparse spectral norm of a tensor $\Ten$ as  
$$	\|\Ten\|_{<d_1,d_2,d_3>}:= \sup_{\substack{\| \uc \|_2 = \|\vc\|_2 = \|\wc\|_2 = 1 \\ \|\uc\|_0=d_1,\|\uc\|_0=d_2,\|\uc\|_0=d_3}} \Big|{\Ten \times_1 \uc \times_2 \vc \times_3 \wc }\Big|,$$
where $d_1 < n_1$, $d_2 <  n_2$, $d_3 < n_3$. When $d_1=d_2=d_3=d$, we simplify $\|\Ten\|_{<d,d,d>}$ as $\|\Ten\|_{<d>}$.

Given a third-order tensor $\Ten \in \Rbb^{n_1 \times n_2 \times n_3}$, we denote its CP decomposition as  
\begin{equation}
\label{eq:CP}
	\Ten =\sum  \limits_{r \in [R]} \lambda_r \ac_r \otimes \bc_r \otimes \cc_r,
\end{equation}
where $[R]$ indicates the set of integer numbers $\{1,\dots,R\}$, and $\otimes$ denotes the outer product of two vectors. For example, the outer product of three vectors $\ac_r \in \Rbb^{n_1}$, $\bc_r \in \Rbb^{n_2}$ and $\cc_r \in \Rbb^{n_3}$ forms a third order tensor of dimension $n_1 \times n_2 \times n_3$ whose $(i,j,k)^{\text{th}}$ entry is equal to $\ace_{ri} \times \bce_{rj} \times \cce_{rk}$ where $\ace_{ri}$ is the $i^{\text{th}}$ entry of $\ac_r$. In \eqref{eq:CP}, $\ac_r, \bc_r, \cc_r$ are of unit norm; that is $\|\ac_r\|_2 =\|\bc_r\|_2= \|\cc_r\|_2 =1$ for all $r \in [R]$; $\lambda_r \in \Rbb^{+}$ is the $r^{th}$ decomposition weight of the tensor. We denote matrices $\Am \in \Rbb^{n_1 \times R}$, $\Bm \in \Rbb^{n_2 \times R}$ and $\Cm \in \Rbb^{n_3 \times R}$ whose columns are $\ac_r,\bc_r$ and $\cc_r$ for $r \in [R]$ respectively as,
\[\Am = [\ac_1, \ac_2, \dots,\ac_R] \quad \Bm = [\bc_1, \bc_2, \dots,\bc_R] \quad \Cm = [\cc_1, \cc_2, \dots,\cc_R].\]

\section{Methodology}
\label{sec:meth}
In this section we introduce our sparse tensor completion model when covariate information is available and propose a non-convex optimization for parameter estimation. Our algorithm employs an alternative updating approach and incorporates a refinement step to boost the performance. 

\subsection{Model}
\label{sec:model}
We observe a third-order tensor $\Ten \in \Rbb^{n_1\times n_2\times n_3}$ and a covariate matrix $\Mat \in \Rbb^{n_1\times n_v}$ corresponding to the feature information along the first mode of the tensor $\Ten$. Here, without loss of generality, we consider the case where the tensor has three modes and the tensor and the matrix are coupled along the first mode. Our method can be easily extended to the case where more than one mode of the tensor has a covariates matrix. Section \ref{sup:General} of the supplement presents a general case where all tensor modes are coupled to covariate matrices.

\red{We consider a widely used random sampling model \citep{jain2014provable, Barak2016, Song2019, Xia2019, cai2020, zhang2020sparse, xia2021statistically, cai2021nonconvex} where the partially observed entries in the tensor are assumed to be uniformly random sampled from the original tensor. That is, let $\Omega$ be the subset of indexes of the tensor $\Ten$ for which entries are not missing. Each index $(i,j,k)$ of the tensor $\Ten$ is included in $\Omega$ independently with reveal probability $p$.} Next we define a projection function $P_{\Omega}(\Ten)$ that projects the tensor onto the observed set $\Omega$, such that
\begin{align}
\label{eq:project}
[P_{\Omega}(\Ten)]_{ijk}= \left\{
\begin{array}{ll}
      \Ten_{ijk} & \text {if } (i,j,k)\in \Omega\\          0 & \text{otherwise.}\    \end{array} 
\right.
\end{align}
In other words $P_\Omega(\cdot)$ is a function that is applied element-wise to the tensor entries and indicates which entries of the tensor are missing.
In this paper, we assume a noisy observation model, where the observed tensor and matrix are noisy versions of their true counterparts. That is,
\begin{align}
  \label{eq:model}
  P_{\Omega}(\Ten) = P_{\Omega}(\Ten^{*} + \Er_T); \quad \Mat = \Mat^{*} + \Er_M,
\end{align}
where $\Er_T$ and $\Er_M$ are the error tensor and the error matrix respectively; $\Ten ^{*}$ and $\Mat^{*}$ are the true tensor and the true matrix, which are assumed to have low-rank decomposition structures \citep{kolda2009tensor};
\begin{equation}
\label{eqn:CPmodel}
\Ten^* =  \sum  \limits_{r \in [R]} \lambda_r^* \ac_r^* \otimes \bc_r^* \otimes \cc_r^*; \quad \Mat^* =  \sum  \limits_{r \in [R]} \sigma_r^* \ac_r^* \otimes \vc_r^*,
\end{equation}
where $\lambda_r^{*} \text{ and } \sigma_r^{*} \in \mathbb{R}^{+}$, and $\ac_r^*\in \mathbb{R}^{n_1} , \bc_r^*\in \mathbb{R}^{n_2} , \cc_r^*\in \mathbb{R}^{n_3} \text{ and } \vc_r^* \in \mathbb{R}^{n_v}$ with $\|\ac_r^*\|_2 =\|\bc_r^*\|_2= \|\cc_r^*\|_2 = \|\vc_r^*\|_2= 1$ for all $r \in [R]$ with $R$ representing the rank of the tensor and matrix. \change{In this article we consider the case that the ranks of both tensor and matrix are the same in order to simplify the presentation and theoretical studies. In this case, the uniqueness of the decomposition is guaranteed \citep{sorensen2015coupled}. However, when the tensor rank and the matrix rank are different, the recovery of low-rank components would become more challenging due to some indeterminacy issue \citep{de2017coupled}.}

As motivated from the online advertisement application, we impose an important sparsity structure on the tensor and matrix components
$\ac_r^* , \bc_r^* , \cc_r^*  \text{ and } \vc_r^*$ such that they belong to the set $\cS(n,d_i)$ with $i=1,2,3,v$, where
\begin{eqnarray}
\label{eq:spars}
\cS(n, d_i) & := & \left\{ \uc \in \mathbb R^{n_i} \Big | \|\uc \|_2=1, \sum_{j=1}^{n_i} \ind_{\{\uc_{j}  \ne 0\}} \le d_i\right\}.
\end{eqnarray}
The values $d_i$ for $i=1,2,3,v$ ar{e considered to be the true sparsity parameters for the tensor and matrix latent components. \change{Note that since the rank $R$ is typically very small in low-rank tensor models, the sum of sparse rank-1 tensors in $(\ref{eqn:CPmodel})$ still leads to a sparse tensor. To illustrate it, suppose each component $\ac_r^*, \bc_r^*, \cc_r^*$ is sparse with only $10\%$ non-zero elements, i.e., $d_i = 0.1 n_i$, then the tensor $\Ten^*$ has at most $R \times 0.001\times n_1n_2n_3$ non-zero entries. In this case, $\Ten^*$ is sparse as long as the rank $R$ is not too large. }

Given a tensor $\Ten$ with many missing entries and a covariate matrix $\Mat$, our goal is to recover the true tensor $\Ten^*$ as well as its sparse latent components. We formulate the model estimation as a joint sparse matrix and tensor decomposition problem. This comes down to finding a sparse and low-rank approximation to the tensor and matrix that are coupled in the first mode.
%{
\begin{align*}
	& \min_{\Am, \Bm, \Cm, \Vm, \blambda, \bsigma}  \Big \{ \| P_{\Omega} \big( \Ten ) -  P_{\Omega} \big( \sum \limits_{r \in [R]} \lambda_r \ac_r \otimes \bc_r \otimes \cc_r \big) \|_F^2 + \| \Mat - \sum \limits_{r \in [R]} \sigma_r \ac_r \otimes \vc_r \|_F ^2   \Big \} \stepcounter{equation}\tag{\theequation} \label{eq:opt2}\\
	& \text{subject to } \|\ac_r\|_2 =\|\bc_r\|_2= \|\cc_r\|_2 = \|\vc_r\|_2= 1, \|\ac_{r} \|_0\le s_1, \|\bc_{r} \|_0 \le s_2 ,\|\cc_{r} \|_0 \le s_3,\|\vc_{r}\|_0 \le s_v.  
\end{align*}
%}
Here $s_i$, $i=1,2,3,v$, are the sparsity parameters and can be tuned via a data-driven way. It is worth mentioning that in this paper we consider the case where the covariate matrix $\Mat$ is fully observed. When $\Mat$ also contains missing entries, we can employ a similar projection function to solve the optimization problem on the observed entries of $\Mat$. In particular, let $\Omega_M$ be the subset of indexes of the matrix $\Mat$ for which entries are not missing, and define a projection function $P_{\Omega_M}(\Mat)$ that projects the matrix onto the observed set $\Omega_M$. When both the tensor $\Ten$ and the covariate matrix $\Mat$ contain missing entries, the objective function in $(\ref{eq:opt2})$ can be adjusted as $\| P_{\Omega} \big( \Ten ) -  P_{\Omega} \big( \sum \limits_{r \in [R]} \lambda_r \ac_r \otimes \bc_r \otimes \cc_r \big) \|_F^2 + \| P_{\Omega_M} \big(\Mat\big) - P_{\Omega_M} \big(\sum \limits_{r \in [R]} \sigma_r \ac_r \otimes \vc_r \big) \|_F ^2 $. The problem in $(\ref{eq:opt2})$ is a non-convex optimization when considering all parameters at once, however the objective function is convex in each parameter while other parameters are fixed. Such multi-convex property motivates us to consider an efficient alternative updating algorithm.
%%%%%%%%%%%%%%%%%%%%%%%%%%%%%%%%%%%%%%%%%%%%%%%%%%%
%% Algorithm
%%%%%%%%%%%%%%%%%%%%%%%%%%%%%%%%%%%%%%%%%%%%%%%%%%%%
\subsection{Algorithm}
\label{sec:alg}
In order to solve the optimization problem formulated in \eqref{eq:opt2}, we use an Alternating Least-Squares (ALS) approach and incorporate an extra refinement step as introduced in \citet{jain2014provable}. In each iteration of ALS, all but one of the components are fixed and the optimization problem reduces to a convex least-squares problem. In order to enforce $\ell_0$ norm penalization in the optimization, we apply a truncation step after each component update similar to that used in \citet{sun2017, zhang2019optimal, hao2020sparse}. For a vector $\uc \in \mathbb{R}^n$ and an index set $F \subseteq [n]$ we define Truncate$(\uc,F)$ such that its $i$-th entry is

\[
     [\text{Truncate}(\uc,F)]_i = 
\begin{cases}
     \uc_i & \text {if } i \in F\\
    0,              & \text{otherwise}.
\end{cases}
\]
For a scalar $s<n$, we denote Truncate($\uc$, $s$)=Truncate($\uc$, supp($\uc$, $s$)), where supp($\uc$, $s$) is the set of indices of $\uc$ which have the largest $s$ absolute values. For example, consider $\uc = (0.1, 0.2, 0.5, -0.6)^{\top}$, we have supp($\uc$, 2) = $\{3,4\}$ and Truncate($\uc$, 2) = $(0,0,0.5,-0.6)^{\top}$. Note that existing sparse tensor models encourage the sparsity either via a Lasso penalized approach \citep{pan2019covariate}, dimension reduction approach \citep{li2017parsimonious}, or sketching \citep{xia2021effective}. We extend the truncation-based sparsity approach in traditional high-dimensional vector models \citep{wang2014high, wang2014optimal} and tensor factorization \citep{sun2017, zhang2019optimal, hao2020sparse} to the tensor completion problem. As shown in \cite{wang2014optimal, sun2017}, the truncation-based sparsity approach often leads to improved estimation performance in practice. 

 %%%%%%%%%%%%%%%%%%%%%%%%%%%%%%%%%%%
%%%Algorithm
%%%%%%%%%%%%%%%%%%%%%%%%%%%%%%%%%%

%Algorithm
\begin{algorithm}[h!]
\caption{COSTCO: Covariate-assisted Sparse Tensor Completion for Solving \eqref{eq:opt2}}
\label{alg:jALS}
\begin{algorithmic}[1]
\STATE \textbf{Input:} Observed tensor $P_\Omega(\Ten) \in \Rbb^{n_1 \times n_2 \times n_3}$, observed matrix $\Mat \in \Rbb^{n_1 \times n_v}$, maximal number of iterations $\tau$, tolerance $tol$, rank $R$, and cardinality $(s_1, s_2,s_3,s_v)$. 
\STATE  Initialize $(\lambda_1, \dots , \lambda_r) , (\Am, \Bm, \Cm)$, $(\sigma_1,\dots \sigma_r), \Vm$. 
\STATE $\ac_r, \bc_r, \cc_r, \vc_r \gets$ the $r^{\text{th}}$ columns of $\Am, \Bm, \Cm \text{ and } \Vm$ respectively, $\forall r\in[R]$
%\vspace{0.1in}
\STATE \textbf{While} {$t \leq \tau$} and {$ \left( \frac{\| \Am_{old} - \Am \|_F }{ \| \Am_{old} \|_F} + \frac{\| \Bm_{old} - \Bm \|_F }{ \| \Bm_{old} \|_F} + \frac{\| \Cm_{old} - \Cm \|_F }{ \| \Cm_{old} \|_F} \right) \geq tol$}, 
\STATE \hspace{0.178in}  $\Am_{old} \gets \Am, \quad \Bm_{old} \gets \Bm, \quad \Cm_{old} \gets \Cm, \quad \Vm_{old} \gets \Vm$
%\vspace{0.1in}
\STATE \hspace{0.178in}	\textbf{For} {$r = 1, \dots, R$}
\STATE  \hspace{0.486in}$\text{res}_T \gets P_\Omega(\Ten ) - P_\Omega( \sum \limits_{m \neq r} \lambda_m \ac_m \otimes \bc_m \otimes \cc_m)$ \quad and \quad   $\text{res}_M \gets \Mat - \sum \limits_{m \neq r} \sigma_m \ac_m \otimes \vc_m$
\STATE    \hspace{0.486in}$\tilde{\ac}_r \gets \Tfrac{\lambda_r\text{res}_T(\mathbf{I}, \bc_r, \cc_r) + \sigma_r \text{res}_M \vc_r}{\lambda_r ^2 P_\Omega(\mathbf{I}, {\bc}_r^2, {\cc_r}^2 ) + \sigma_r^2  }$
%\vspace{0.1in}
\STATE   \hspace{0.486in}$\tilde{\ac}_r \gets \text{Truncate}(\tilde{\ac}_r , s_1)$,\quad $\ac_r \gets \tilde{\ac}_r / \| \tilde{\ac}_r\|_2$
%\vspace{0.1in}
\STATE  \hspace{0.486in}$\tilde{\bc}_r \gets \Tfrac{ \text{res}_T(\ac_r,\mathbf{I}, \cc_r)}{ P_\Omega({\ac}_r^2,\mathbf{I},{\cc}_r^2)}$, \quad
    		 $\tilde{\cc}_r \gets \Tfrac{ \text{res}_T(\ac_r, \bc_r, \mathbf{I})}{ P_\Omega({\ac}_r^2, {\bc}_r^2, \mathbf{I})}$ \quad and \quad $\tilde{\vc}_r \gets \text{res}_M^\top  \ac_r  $
% \vspace{0.1in}   		
\STATE  \hspace{0.486in}$\tilde{\bc}_r \gets \text{Truncate}(\tilde{\bc}_r , s_2) $  	\quad  $\tilde{\cc}_r \gets \text{Truncate}(\tilde{\cc}_r , s_3)$, 	\quad  $\tilde{\vc}_r \gets \text{Truncate}(\tilde{\vc}_r , s_v)$
%\vspace{0.1in}
\STATE  \hspace{0.486in} $\lambda_r \gets \|  \tilde{\cc}_r \|_2$, \quad $\sigma_r \gets \|  \tilde{\vc}_r \|_2$ \\ 
%\vspace{0.1in}
\STATE \hspace{0.486in} $\bc_r \gets \tilde{\bc}_r / \| \tilde{\bc}_r \|_2$, \quad $\cc_r \gets \tilde{\cc}_r / \| \tilde{\cc}_r \|_2$, \quad $\vc_r \gets \tilde{\vc}_r / \| \tilde{\vc}_r\|_2$
\STATE \hspace{0.178in} \textbf{End For}
%\vspace{0.1in}
\STATE \textbf{End While}
\end{algorithmic}
\end{algorithm}
Our \texttt{COSTCO} in Algorithm~\ref{alg:jALS} takes a matrix $\Mat$ and a tensor $\Ten $ with missing entries as input and computes the components of the matrix and tensor. Due to the non-convexity of the optimization problem, there could be multiple local optima. In our algorithm we initialize the tensor and matrix components using the procedure in Section \ref{sup:initialization} which is shown through extensive simulations to provide good starting values for the tensor and matrix components. Line 6 of the algorithm has an inner loop on $r \in [R]$ which loops on each tensor rank. This inner loop on $r$ performs an ``extra refinement'' step that was first introduced in \citet{jain2014provable} for tensor completion; and is, therein, proved to improve the error bounds of tensor recovery. 

The main component updates are performed in Lines 8 and 10 which are solutions to the least-squares problem while other parameters are fixed. Note that the horizontal double line in Lines 8 and 10 indicate element-wise fraction and the squaring in the denominator applies entry-wise on the vectors. After obtaining these non-sparse components, Lines 9 and 11 perform the truncation operator to encourage the sparsity on the latent components. The detailed derivation of this algorithm is shown in Lemma \ref{lem:update} in the supplementary material. Finally, the algorithm stops if either the maximum number of iterations $\tau$ is reached or the normalized Frobenius norm difference of the current and previous components are below a threshold $tol$.

\change{Algorithm~\ref{alg:jALS} handles two possible sources of identifiability issues. First, after obtaining the sparse update $\tilde{\ac}_r, \tilde{\bc}_r, \tilde{\cc}_r, \tilde{\vc}_r$, it normalizes these components by its Euclidean norm so that all factor vectors $\ac_r, \bc_r, \cc_r, \vc_r$ (Lines 9 and 13 of Algorithm 1) are scaling-identifiable. Second, when there are a few entries of the same largest absolute values in a vector, the Truncate operator in Lines 9 and 11 ensures that the same entries will be kept. To illustrate it, consider $\uc = (0.5, 0.5, 0.5, 0.4, 0.3)^{\top}$ and the sparsity parameter $s = 2$, $\textrm{Truncate}(\uc, 2)$ always returns a sparse vector $(0.5, 0.5, 0, 0, 0)^{\top}$, i.e., only the first appear $s$ largest absolute values are kept. }

\begin{figure}
   \centering
        \includegraphics[width=0.7\textwidth]{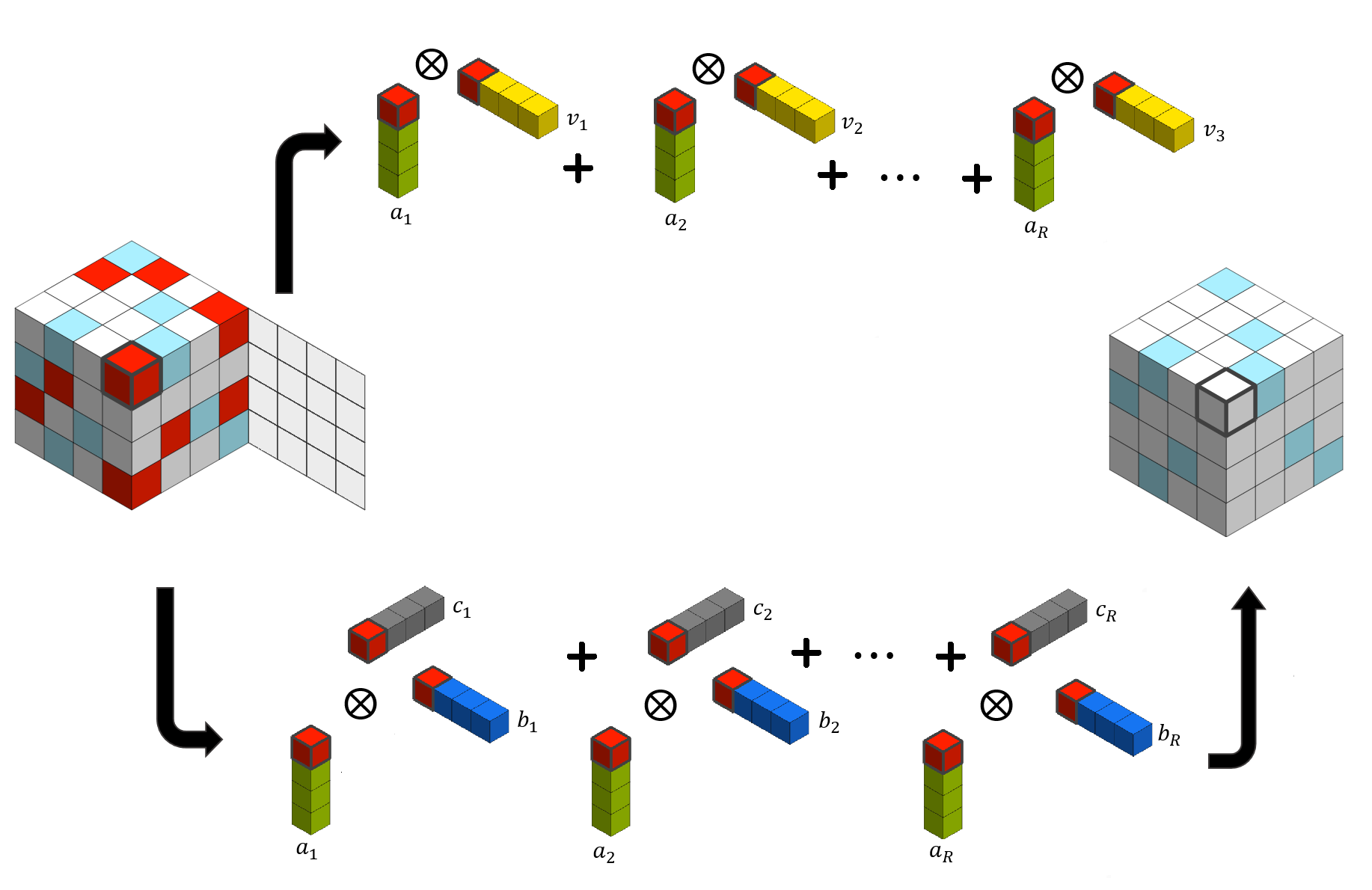}
        \caption{Illustration of \texttt{COSTCO} showing recovery procedure for missing entries through joint tensor matrix decomposition; red cells represent missing entries. The tensor and matrix are coupled along the first mode and the components ${\ac}_r$, $r\in[R]$ are shared by the tensor and matrix decomposition.}
        \label{fig:costco}
\end{figure}

Figure \ref{fig:costco} is an illustration of \texttt{COSTCO} that reveals the intuition behind the working of Algorithm \ref{alg:jALS}. As the percentage of missing entries in the tensor increases, recovering the tensor components using only the observed tensor entries leads to a reduction in the accuracy of the recovered tensor components. However, with \texttt{COSTCO}, we leverage the additional latent information coming from the matrix of covariates on the shared mode. The signal obtained from the matrix contributes in improving the recovery of the shared components and indirectly that of the non-shared components as well. This observation is reflected on Line 8 of Algorithm \ref{alg:jALS} for the shared component update, where we see in the denominator that even when $P_\Omega(\mathbf{I}, {\bc}_r^2, {\cc_r}^2 )$ is close to zero (meaning most entries of the tensor are missing) the denominator remains a non-zero value due to the signal from the covariate matrix. In this case we are still able to estimate the shared component $\ac_r$. This would not be the case without the addition of the covariates matrix information, where the denominator for the update would only be $P_\Omega(\mathbf{I}, {\bc}_r^2, {\cc_r}^2 )$ which is close to zero. Therefore, a standalone tensor completion algorithm would become unstable. In the more general case where all three modes of the tensor are coupled to their own covariates matrices, it is easy to see from the illustration in Figure \ref{fig:costco} that the missing percentage of the tensor could be close to $100\%$. This is because in such case, the covariates matrix components could still be used in the algorithm to recover the tensor components for all three modes and therefore recover the tensor entries. 
%%%%%%%%%%%%%%%%%%%%%%%%%%%%%%%%%%%
%%%Algorithm initialization
%%%%%%%%%%%%%%%%%%%%%%%%%%%%%%%%%%
\subsubsection{Initialization Procedure}
\label{sup:initialization}
This section presents details about the method used for the initialization procedure on Line 2 of Algorithm \ref{alg:jALS}. Unlike matrix completion, success in designing an efficient and accurate algorithm for the tensor completion problem is contingent to starting with a good initial estimates. In fact, the convergence rate of low-rank tensor algorithms is typically written as a function of the tensor components weights as well as the initialization error \citep{anandkumar2014tensor, jain2014provable, sun2017, cai2021nonconvex, xia2021statistically}. It is therefore imperative to design an initialization procedure efficient enough to help rule out local stationary points. 

We use to our advantage, the fact that in our model, the tensor and matrix share at least one mode and use the singular value decomposition (SVD) \citep{stewart1990,Ipsen} of the observed matrix $\Mat$ to initialize the shared components of the tensor $\Am$ along with the matrix weights $\sigma_1, \cdots, \sigma_R$ and matrix component $\Vm$ respectively. We then use the robust tensor power method (RTPM) from \citet{anandkumar2014tensor} to initialize the non-shared components $\Bm$ and $\Cm$ and the tensor weights. This is done by setting all missing entries in the tensor to be zero before running RTPM. In practice we show in our simulations in Section \ref{sec:simul} that this is an adequate initialization procedure and produces much better initials compared to a random initialization scheme. In the more general case where all tensor modes have covariate matrices, the SVD on the covariate matrices can be used to initialize all the tensor components. In this case, the RTPM for non-shared components initialization would not be needed.

%%%%%%%%%%%%%%%%%%%%%%%%%%%%%%%%%%%
%%%Algorithm tuning cardinality and rank
%%%%%%%%%%%%%%%%%%%%%%%%%%%%%%%%%%
\subsubsection{Rank and Cardinality Tuning}
\label{sup:tuning}
Our \texttt{COSTCO} method relies on two key parameters: the rank $R$ and the sparsity parameters. It has been shown that exact tensor rank calculation is a NP-hard problem \citep{kolda2009tensor}. In this section, following the tuning method in \cite{allen2012, sun2017}, we provide a BIC-type
criterion to tune these parameters. Given a pre-specified set of rank values $\mathcal{R}$ and a pre-specified set of cardinality values $\mathcal{S}$,
we choose the parameters which minimizes 
\begin{align*}
  BIC= &\log{\left(\frac{\|P_{\Omega} \big( \Ten- \sum \limits_{r \in [R]} \lambda_r \ac_r \otimes \bc_r \otimes \cc_r \big)\|_{F}^{2}}{n_1 n_2n_3} + \frac{\|\Mat- \sum \limits_{r \in [R]} \sigma_r \ac_r \otimes \vc_r\|_{F}^{2}}{n_1 n_v}\right)} \stepcounter{equation}\tag{\theequation}\label{eq:BIC}\\ 
  & +\frac{\log{(n_1n_2n_3 +n_1 n_v)}}{(n_1 n_2 n_3+n_1n_v)\sum \limits_{r \in [R]} (\|\ac\|_0 + \|\bc\|_0+ \|\cc\|_0+ \|\vc\|_0) }.
\end{align*}
To further speed up the computation, in practice, we tune these parameters sequentially. That is, we first fix $s_i=n_i$ and tune the rank $R$ via $(\ref{eq:BIC})$. Then given the tuned rank, we tune the sparsity parameters. This tuning procedure works very well through simulation studies in Section \ref{sec:simul}.

%%%%%%%%%%%%%%%%%%%%%%%%%%%%%%%%%%%%%%%%%%
%%%%%%%% THEORETICAL ANALYSIS
%%%%%%%%%%%%%%%%%%%%%%%%%%%%%%%%%%%%%%%%%%%
\section {Theoretical Analysis}
\label{sec:analtheo} 
In this section, we derive the error bound of the recovered tensor components obtained from Algorithm~\ref{alg:jALS}. \red{We only provide the results for tensor components as our ultimate goal is to recover the missing entries in the tensor.} We present the recovery results for the estimated shared components $\ac_r$ and non-shared tensor components $\bc_r$ and $\cc_r$ separately to highlight the sharp improvement in recovery accuracy of the tensor resulting from incorporating the covariate information.    

The theory is presented in two phases, first we focus on a simplified case in which the true tensor and matrix components $\ac_r^*, \bc_r^*,\cc_r^*$ and $\vc_r^*$ are non-sparse and both tensor and matrix weights are equal (i.e, $\sigma_r^* =\lambda_r^*$, $\forall r \in [R]$). Presenting this simplified case allows us to showcase clearly the interplay between the reveal probability, the tensor and matrix dimensions as well as how the noises in the tensor and matrix affect the statistical and computational errors of the algorithm. In the second case, we then present the results for the general scenario where the tensor and matrix weights are allowed to be unequal and the tensor and matrix components are assumed to be sparse. 

\subsection{Case 1: Non-sparse Tensor and Matrix with Equal Weights} 
Before presenting the theorem for the simplified case, we introduce assumptions on the true tensor $\Ten^*$ and matrix $\Mat^{*}$ and then discuss their utility. Denote $ n := \max{(n_1, n_2 ,n_3,n_v)}$.
%%%%%%%%%%%%%%%%%%%%%%%%%%%%%%%%%%%%%%%%%%%%%%%%%%%%

\textbf{Assumption 1:} (Tensor and matrix structure)
\begin{enumerate}[i.]
    \item Assume $\Ten^{*}$ and $\Mat^{*}$ are specified as in \eqref{eqn:CPmodel} with unique low-rank decomposition up to a permutation, and assume rank $R =o(n^{1/2})$ and $\lambda_r^*=\sigma_r^*$ (equal weight), $\forall r \in[R]$.% and are bounded away from $0$ and $\infty$.
    \item The entries of the decomposed components for both $\Ten^{*}$ and $\Mat^{*}$ satisfy the $\mu$-mass condition,
\[ \max_r \{\|\ac_r^{*}\|_{\infty},\|\bc_r^{*}\|_{\infty},\|\cc_r^{*}\|_{\infty}, \|\vc_r^{*}\|_{\infty}\}\leq \frac{\mu}{\sqrt{n}}, \]
where $\mu$ is a constant.
 \item The components across ranks for both $\Ten^{*}$ and $\Mat^{*}$ meet the incoherence condition,
\[ \max \limits_{i\neq j} \left \{ |\langle \ac_{i}^{*} {,} \ac_{j}^{*}\rangle|,  |\langle \bc_{i}^{*} {,} \bc_{j}^{*}\rangle| ,  |\langle \cc_{i}^{*} {,} \cc_{j}^{*}\rangle| , |\langle \vc_{i}^{*} {,} \vc_{j}^{*}\rangle|  \right \} \leq \frac{c_0}{\sqrt{n}},\]
where $c_0$ is a constant.
\end{enumerate}
Assumption (1\textit{i}) is a common assumption in the tensor decomposition literature to ensure identifiability \cite{kolda2009tensor,anandkumar2014tensor,  jain2014provable, sun2017}. It imposes the condition that the tensor admits a low rank CP decomposition that is unique. This is the case of the undercomplete tensor decomposition, where the rank of the tensor is assumed to be lower than the dimension of the component. \change{The condition $\lambda_r^*=\sigma_r^*$ is a simplification of the problem that allows us to simplify the derivation and showcase clearly the interplay between important parameters. The same results (up to a constant) in Theorem 1 would hold if $\sigma_r^*$ is of the same order as $\lambda_r^*$. The general weight case is described in Section \ref{sec:case2}.} Assumption (1\textit{ii}) ensures that the mass of the tensor is not contained in only a few entries and is necessary if one hopes to recover any of the non-share components of the tensor with acceptable accuracy.
Assumption (1\textit{iii}) is related to the non-orthogonality of the tensor components and imposes a soft orthogonality condition on the tensor and matrix components. That is, the tensor components are allowed to be correlated only to a certain degree. \citet{2014guaranteed} and \citet{sun2017} show that such a condition is met when the tensor and matrix component are randomly generated from a Gaussian distribution. Both the $\mu$-mass condition and the incoherence conditions have been commonly assumed in low-rank tensor models \citep{anandkumar2014tensor,  jain2014provable, sun2017, cai2021nonconvex, Xia2019, cai2020}.

\textbf{Assumption 2:} (Reveal probability) Denote $\lambda_{min}^*:=\min\limits_{r\in[R]}\{\lambda_r^*\}$ and $\lambda_{max}^*:=\max\limits_{r\in[R]}\{\lambda_r^*\}$. We assume that each entry $(i,j,k)$ of the tensor $\Ten^*$ for all  $i \in [n_1]$, $j \in [n_2]$ and $k \in [n_3]$ is observed with equal probability $p$ which satisfies,
\[p \geq  \frac{CR^2\mu^3 \lambda_{max}^{*2} \log^2(n) }{(\lambda_{min}^*+\sigma_{min}^*)^2n^{3/2}},\]
where $C$ is a constant.

Assumption 2 guarantees that the tensor entries are revealed uniformly at random with probability $p$. The lower bound on $p$ is an increasing function of the tensor rank since recovering tensors with a larger rank is a harder problem which requires more observed entries. The bound on $p$ is also an increasing function of the $\mu$-mass parameter since a larger $\mu$-mass parameter in Assumption (1\textit{ii}) indicates a smaller signal in each tensor entry and hence more reveal entries for accurate component recovery would be needed. Moreover, the bound on $p$ is a decreasing function of the tensor component dimension $n$ and relates as $n^{-3/2}$ up to a logarithm term. This is the optimal dependence on the dimension in tensor completion literature \citep{jain2014provable, Xia2019}. Most importantly, the lower bound on $p$ is relaxed when the minimal weight $\lambda_{\min}^*$ of the tensor or the minimal weight $\sigma_{\min}^*$ of the matrix increases. This reflects a critical difference when compared to the lower bound condition required in traditional tensor completion \citep{jain2014provable, Xia2019} which corresponds to the case $\sigma_{\min}^* = 0$. It shows the advantage of coupling the matrix of covariates for the tensor completion. This new lower bound on $p$ translates to requiring less observed entries for the tensor recovery in the presence of covariates. \change{Note that in the present simplified case $\sigma_r^* = \lambda_r^*$, we still choose to write $\sigma_{\min}^*$ explicitly in the lower bound condition to showcase the effect of the covariate information. The improvement on $p$ over existing literature will be clearer in Assumption 6 for the general weight case.}

\textbf{Assumptions 3} (Initialization error) Define the initialization errors for the tensor components as 
$ \epsilon_{0_T}:= \max_{r \in [R]}\{\|\ac_r^{0}-\ac_{r}^{*} \|_2,\|\bc_r^{0}-\bc_{r}^{*} \|_2, \|\cc_r^{0}-\cc_{r}^{*} \|_2, \frac{|\lambda_r^{0} -\lambda_r^{*}|}{\lambda_r^{*}} \}$ and the initialization error for the matrix components as $ \epsilon_{0_M}:=\max_{r \in [R]} \{\|\vc_r^{0}-\vc_{r}^{*} \|_2, \frac{|\sigma_r^{0} -\sigma_r^{*}|}{\sigma_r^{*}} \}$. Assume that 
\begin{align*}
   \epsilon_{0}:=\max\{\epsilon_{0_T},\epsilon_{0_M}\}\leq & \frac{\lambda_{min}^*}{100R\lambda_{max}^*}-\frac{c_0}{3\sqrt{n}}. \stepcounter{equation}\tag{\theequation}\label{eq:et01}
\end{align*}

Here the component $c_0/\sqrt{n}$ is due to the non-orthogonality of the tensor factors. When the components are orthogonal, we allow a larger initialization error. This observation aligns with the common knowledge in tensor recovery as the problem is known to be harder for non-orthogonal tensor factorization \citep{2014guaranteed}. Similarly, a larger rank $R$ of the tensor leads to a harder problem and a stronger condition on the initialization error. Under Assumption (1i) $R =o(n^{1/2})$, when the condition number $\lambda_{max}^*/\lambda_{min}^* = \mathcal{O}(1)$, this initial condition reduces to $\epsilon_{0}=\mathcal{O}(1/R)$. As shown in \cite{2014guaranteed,  jain2014provable}, the robust tensor power method initialization procedure used in our Algorithm satisfies $\mathcal{O}(1/R)$ error bound.

\textbf{Assumption 4} (Signal-to-noise ratio condition) Denote $\|\mathcal{E}_T \|$, $\|\mathcal{E}_M \|$ as the spectral norm of the error tensor and error matrix, respectively. We assume that 
\begin{align}
\frac{\| \Er_T \|}{\sqrt{p}\lambda_{min}^*} =o(1) \quad \text{and} \quad \frac{\| \Er_M\|}{ (p+1)\lambda_{min}^*} =o(1). \label{eqn:SNR}
\end{align}
Assumption 4 can be considered as the commonly used signal-to-noise ratio condition in noisy tensor decomposition \citep{sun2017, cai2021nonconvex, Sun2019, xia2021statistically}. It ensures that the estimators for both shared and non-shared components contract in each iteration and the corresponding final statistical errors converge to zero. Note that when all mode of the tensors are coupled with covariate matrices, the condition on $\| \Er_T \|$ can be relaxed to $\frac{\sqrt{p}\| \Er_T \|}{ (p+1)\lambda_{min}^*} =o(1)$ due to the incorporation of covariate matrices for all shared components. 

%%%%%%%%%%%%%%%%%%%%%%%%%%%%%%%%%%%%%%%%%%%%%%%%%%%%%%%%%%%%%%%%
\begin{theorem}[Non-sparse tensor and matrix components with equal weights]
\label{theorem:global1}
\space 
Assuming Assumptions 1, 2 , 3 and 4 are met. After running $\Omega\left(\log_{2}{\left(\frac{(p+1)\lambda_{min}^{*}\epsilon_{0} }{\sqrt{p}\|\mathcal{E}_T\| +\|\mathcal{E}_M \|} \vee \frac{\sqrt{p}\lambda_{min}^{*}\epsilon_{0}}{\|\mathcal{E}_T \|}\right)}\right)$ iterations of Algorithm \ref{alg:jALS} with $s_i = n_i$, for $i=1,2,3,v$, we have 
\begin{itemize}
\item \textbf{Shared Component $\ac_{r}$}:
\begin{equation}
\label{eq:conva-1}
   \max_{r \in [R]}\left( \|\ac_r-\ac_{r}^{*} \|_2\right) = {\mathcal{O}_p\left( \frac{\sqrt{p}\|\mathcal{E}_T \| +\|\mathcal{E}_M \|}{(p+1)\lambda_{min}^{*}}\right)}.
\end{equation}
\item \textbf{Non-Shared Components $\bc_{r}$, $\cc_{r}$}:
\begin{equation}
\label{eq:convc-1}
  \max_{r\in[R]} \left( \|\bc_r-\bc_{r}^{*} \|_2,\|\cc_r-\cc_{r}^{*} \|_2 ,  \frac{|\lambda_r -\lambda_{r}^{*}|}{  \lambda_r^{*}}\right)= {\mathcal{O}_p\left(\frac{\|\mathcal{E}_T \|}{\sqrt{p}\lambda_{min}^{*}}\right)}.
\end{equation}
 \end{itemize}
\end{theorem}

Theorem \ref{theorem:global1} indicates that the shared component error is a weighed average of the spectral norm of the error tensor and error matrix. Whereas the non-shared component error is simply a function of the error tensor. In the extreme case in which the covariates matrix $\Mat$ is noiseless, then the recovery error of the shared component becomes $\frac{\sqrt{p}\|\mathcal{E}_T \|}{(p+1)\lambda_{min}^{*}},$ which is much smaller than the recovery error of the non-shared component $\frac{\|\mathcal{E}_T \|}{\sqrt{p}\lambda_{min}^{*}}$, especially when the observation probability $p$ is very small. Moreover even in the case in which the coupled covariates matrix is not noiseless, since $p\leq 1$ we notice an improvement in the statistical error of the recovered shared component compared to that of the non-shared components as long as the spectral norm of the error matrix is no larger than the spectral norm of the error tensor. 

\begin{remark}\label{remark:gaussian}
\change{(Sub-Gaussian noise) In Theorem \ref{theorem:global1}, we consider the noisy model with a general error tensor and error matrix. When the entries of the error tensor $\mathcal{E}_T$ and the error matrix $\mathcal{E}_M$ are i.i.d sub-Gaussian with mean zero and variance proxy $\sigma^2$, we can further simply the statistical error. For simplicity, consider $\mathcal{E}_T \in \mathbb R^{n\times n\times n}$ and $\mathcal{E}_M \in \mathbb R^{n\times n}$. According to \cite{tomioka2014spectral} and \cite{vershynin2018high}, $\|\mathcal{E}_T \| = \mathcal{O}_p(\sigma \sqrt{n\log(n)})$ and $\|\mathcal{E}_M \| = \mathcal{O}_p(\sigma \sqrt{n\log(n)})$. Therefore, the errors of the shared component in $(\ref{eq:conva-1})$ and that of the non-shared component in $(\ref{eq:convc-1})$ can be simplified as 
 $$
(\ref{eq:conva-1}) = \mathcal{O}_p\left( \frac{\sigma}{\lambda_{min}^{*}} \frac{(\sqrt{p}+1) \sqrt{n\log(n)}}{(p+1)}  \right); ~~  (\ref{eq:convc-1}) = \mathcal{O}_p\left( \frac{\sigma}{\lambda_{min}^{*}} \sqrt{\frac{n\log(n)}{p} } \right).
 $$
The estimation error for the non-shared component matches with that in the standalone tensor completion \citep{cai2021nonconvex}, while the estimation error for the shared component largely improves due to the incorporation of the covariate matrix. The improvement is more significant especially when the observation probability $p$ is small as $(\sqrt{p}+1)/(p+1) \prec 1/\sqrt{p}$.}
\end{remark}

%%%%%%%%%%%%%%%%%%%%%%%%%%%%%%%%%%%%%%%%%%%%%%%%%%%%%%%%%%%%%%%%
\subsection{Case 2: Sparse Tensor and Matrix with General Weights}
\label{sec:case2}
We now present the result for the general case with sparse tensor and matrix $\Ten^{*}$ and $\Mat^{*}$ and the weights of the tensor and matrix are allowed to be unequal. The theoretical analysis for the general case is much more challenging than that covered in Theorem \ref{theorem:global1}. For example, unlike the setting in Case 1, we are no longer able to derive the closed form solution to the optimization problem in \eqref{eq:opt2} for the shared tensor component. Instead, we construct an intermediate estimate in the analysis of the shared component recovery. Fortunately, this general result allows us to explicitly quantify the improvement due to the covariates on the missing percentage requirement and the final error bound.

The following conditions are needed for the general scenario. Recall that $d = \max\{d_1, d_2, d_3,d_v\}$ is the maximal true sparsity parameter defined in \eqref{eq:spars} and define $s:=\max\{s_1,s_2,s_3,s_v\}$.
%%%%%%%%%%%%%%%%%%%%%%%%%%%%%%%%%%%%%%%%%%%%%%%%%%
%\subsection{Assumptions}

\textbf{Assumption 5} (sparse tensor and matrix structure) 
\begin{enumerate}[i.]
    \item Assume $\Ten^{*}$ and $\Mat^{*}$ have the sparse structure in \eqref{eqn:CPmodel} and \eqref{eq:spars} with unique low-rank decomposition up to a permutation, and assume rank $R =o(d^{1/2})$.
    \item The entries of the decomposed components for $\Ten^{*}$  satisfy the following $\mu$-mass condition
\[ \max_r \{\|\ac_r^{*}\|_{\infty},\|\bc_r^{*}\|_{\infty},\|\cc_r^{*}\|_{\infty}, \|\vc_r^{*}\|_{\infty}\}\leq \frac{\mu}{\sqrt{d}}. \]
\item The components across ranks for both $\Ten^{*}$ and $\Mat^{*}$ meet the incoherence condition,
\[ \max \limits_{i\neq j} \left \{ |\langle \ac_{i}^{*} {,} \ac_{j}^{*}\rangle|,  |\langle \bc_{j}^{*} {,} \bc_{i}^{*}\rangle| ,  |\langle \cc_{j}^{*} {,} \cc_{i}^{*}\rangle| , |\langle \vc_{j}^{*} {,} \vc_{i}^{*}\rangle|  \right \} \leq \frac{c_0}{\sqrt{d}}.\]
\end{enumerate}
Notice that since the components of tensor and matrix are assumed to be sparse, the $\mu$-mass and incoherence condition are functions of the maximum number of non-zero elements $d$ in the tensor and matrix components rather than the dimension $n$. In the case in which $d\ll n$, this constitutes a milder assumption compared to Assumptions 1\text(ii) and 1\text(iii).

\textbf{Assumption 6} (Reveal probability) We assume that each tensor entry $(i,j,k)$ for all  $i \in [n_1]$, $j \in [n_2]$ and $k \in [n_3]$ is observed with equal probability $p$ which satisfies,
\begin{align}
\label{eq:revealp2}
p \geq  \frac{CR^2\mu^3 \lambda_{max}^{*2} \log^2(d) }{(\lambda_{min}^*+\sigma_{min}^*)^2d^{3/2}}.
\end{align} 

Similar to the equal-weight case, the required lower bound on the reveal probability in \eqref{eq:revealp2} improves the established lower bound for the tensor completion with no covariates matrix. \change{Specifically, \cite{jain2014provable, Montanari2018, Xia2019} show that the lower bound for non-sparse tensor completion is of the order $\frac{\lambda_{max}^{*2}\log^2(n)}{\lambda_{min}^{*2}n^{3/2}}$ while our lower bound is of the order $\frac{\lambda_{max}^{*2}\log^2(n)}{(\lambda_{min}^*+\sigma_{min}^*)^2n^{3/2}}$ when the components are not sparse ($d=n$). This highlights the fact that a weaker assumption on the reveal probability is required in the presence of covariates matrix than in the case with no covariates. An interesting phenomenon is that when the minimal weight of the matrix $\sigma_{\min}^*$ is very large, we could allow the reveal probability to be even close to zero. For example, in the non-sparse case, when $\lambda_{max}^* = O(\lambda_{\min}^*)$ and $\sigma_{min}^* / \lambda_{max}^* = \sqrt{n}$, our lower bound on $p$ is relaxed to $O(n^{-5/2})$ up to a logarithm order. In fact, as long as $\lambda_{max}^* = o(\sigma_{min}^*)$ and $\lambda_{max}^* = O(\lambda_{\min}^*)$, the lower bound would be smaller than $O(n^{-3/2})$. This is a major advantage of our method and this property does not exist in existing standalone tensor completion which requires $n^{-3/2}$ lower bound on $p$. As demonstrated in our simulations, our \texttt{COSTCO} is still satisfactory even when $98\%$ of the tensor entries are missing, while the traditional tensor completion method start to fail when there are more than $90\%$ missing entries.} Moreover, in the sparse case, the lower bound is a decreasing function of the sparsity parameter $d$. This is intuitive as when $d$ decreases, the non-zero tensor components will concentrate on fewer dimensions which makes the tensor recovery problem harder.

\textbf{Assumption 7} (Initialization error) Assume that 
\begin{align*}
  \epsilon_{0}:=\max\{ \epsilon_{0_T}, \epsilon_{0_M}\} \leq & \frac{95/96\lambda_{min}^{*2}+\sigma_{min}^{*2}}{144R(\lambda_{max}^{*2}+\sigma_{max}^{*2})}-\frac{c_0}{3\sqrt{d}}, \stepcounter{equation}\tag{\theequation}\label{eq:et02}
\end{align*}
with $\epsilon_{0_T}$ and $\epsilon_{0_M}$ as defined in Assumption 3.

Compared to that in Assumption 3, the initialization condition for Case 2 is slightly stronger. This is reflected on two parts. First, the term $c_0/\sqrt{d}$ is due to the non-orthogonality of sparse tensor components and is larger in the sparse case. This requires a stronger condition on the rank $R$ as shown in Assumption (1i) in order to ensure the positivity of the right-hand side of $(\ref{eq:et02})$. Second, the ratio $(95/96\lambda_{min}^{*2}+\sigma_{min}^{*2}) / 144(\lambda_{max}^{*2}+\sigma_{max}^{*2}))$ is smaller than $\lambda_{min}^{*}/(100\lambda_{max}^{*})$ in Assumption 3. Even when $\lambda_r^{*} = \sigma^*_r$ and $d=n$, this condition is still slightly stronger than Assumption 3 since $\lambda_{min}^{*2}/\lambda_{max}^{*2} < \lambda_{min}^{*}/\lambda_{max}^{*}$. This additional term is due to handling the non-equal weights. Fortunately, when condition numbers $\lambda_{max}^*/\lambda_{min}^* = \mathcal{O}(1)$ and $\sigma_{max}^*/\sigma_{min}^* = \mathcal{O}(1)$, we have $\epsilon_{0}=\mathcal{O}(1/R)$, which is again satisfied by the initialization procedure in our algorithm.

{
\textbf{Assumption 8} (Signal-to-noise ratio condition) Denote $\|\mathcal{E}_T \|_{<s>}$, $\|\mathcal{E}_M \|_{<s>}$ as the sparse spectral norm of the error tensor and error matrix defined in Section \ref{sec:prelim}. We assume that 
\begin{align}
\frac{\| \Er_T \|_{<s>}}{\sqrt{p}\lambda_{min}^*} =o(1) \quad \text{and} \quad \frac{\sigma^*_{max}\| \Er_M\|_{<s>}}{ p\lambda_{min}^{*2} + \sigma_{min}^{*2}} =o(1). \label{eqn:SNR_general}
\end{align}
Assumption 8 extends the signal-to-noise ratio condition in Assumption 4 to the sparse and general non-equal weight case. 
}
%%%%%%%%%%%%%%%%%%%%%%%%%%%%%%%%%%%%%%%%%%%%%%
\begin{theorem}[Sparse tensor and matrix components with general weights]
\label{theorem:global2}  
Assuming assumptions 5, 6, 7 and 8 are met. After running {$\Omega\left(\log_{2}{\left(\frac{ (p\lambda_{min}^{*2}+\sigma_{min}^{*2})\epsilon_{0}}{\sqrt{p}\lambda_{max}^{*}\|\mathcal{E}_T \|_{<s>} +\sigma_{max}^{*}\|\mathcal{E}_M \|_{<s>}} \vee  \frac{\sqrt{p}\lambda_{min}^{*}\epsilon_{0}}{\|\mathcal{E}_T \|_{<s>}\epsilon_T}\right)}\right)$} iterations of Algorithm \ref{alg:jALS} with $s_i \ge d_i$, for $i=1,2,3,v$, we have
\begin{itemize}
\item \textbf{Shared Component $\ac_r$: }
\begin{equation}
\label{eq:conva-2}
\max_{r \in [R]}\left( \|\ac_r-\ac_{r}^{*} \|_2\right) = {\mathcal{O}_p\left(\frac{\sqrt{p}\lambda_{max}^{*}\|\mathcal{E}_T \|_{<s>} +\sigma_{max}^{*}\|\mathcal{E}_M \|_{<s>}}{p\lambda_{min}^{*2}+\sigma_{min}^{*2}}\right)}.
\end{equation}
\item \textbf{Non-Shared Components $\bc_r, \cc_r$:}
\begin{equation}
\label{eq:convc-2}
\max_{r\in[R]} \left( \|\bc_r-\bc_{r}^{*} \|_2,\|\cc_r-\cc_{r}^{*} \|_2 ,  \frac{|\lambda_r -\lambda_{r}^{*}|}{  \lambda_r^{*}}\right) = {\mathcal{O}_p\left(\frac{\|\mathcal{E}_T \|_{<s>}}{\sqrt{p}\lambda_{min}^{*}}\right)}.
\end{equation}
 \end{itemize}
\end{theorem}
Similar to that in Theorem \ref{theorem:global1}, the statistical error for the shared tensor component in Theorem \ref{theorem:global2} is a weighed average of the sparse spectral norm of the error tensor $\Er_T$ and error matrix $\Er_M$. The key difference is that the weight is now related to $\lambda^*_{\max}$ and $\sigma^*_{\max}$ and the spectral norm is now much smaller than the non-sparse counterparts in Theorem \ref{theorem:global1} since typically $s \prec n$ and hence $\|\mathcal{E}_T \|_{<s>} \prec \|\mathcal{E}_T \|$ and $\|\mathcal{E}_M \|_{<s>} \prec \|\mathcal{E}_M \|$. Similarly, the recovery error for the non-shared tensor component in the general case is also smaller than that in \eqref{eq:convc-1} due to a smaller spectral norm. This observation highlights the advantage of considering sparse tensor components. In addition, we highlight a few important scenarios in Table \ref{tab:theoresults} where the error of shared tensor component is smaller than that of the non-shared component. Such scenario indicates when the additional covariate information is useful to reduce the estimation error of the tensor components. In summary, such improvement is observed when the sparse spectral norm of the error matrix is smaller than or comparable to that of the error tensor. 

\begin{table}[h!]
\centering
\begin{small}
\caption{Statistical error of shared tensor component in Theorem \ref{theorem:global2} under various conditions. ``Improved" means there is an improvement over the error of the non-shared components.}
\label{tab:theoresults}
%\vskip 1 em
\begin{tabular}{|>{\centering\arraybackslash}p{2cm}|>{\centering\arraybackslash}p{3.5cm}|>{\centering\arraybackslash}p{5.8cm}|>{\centering\arraybackslash}p{2.1cm}|}
\hline
\textbf{Condition Number} & \textbf{Noise} & \textbf{Statistical Error} & \textbf{Improved?}
\hspace{0.05in}\\
 \hline
 & $\|\Er_M\|_{<s>}=0 $ & $\mathcal{O}_p\left(\frac{p\|\mathcal{E}_T \|_{<s>} }{\sqrt{p}\lambda_{min}^*+\sigma_{min}^*}\right)$ & \checkmark \\
\cline{2-4}
 $\frac{\lambda_{max}^{*}}{\lambda_{min}^{*}}=\mathcal{O}(1)$   &  $\|\Er_M\|_{<s>}= \|\Er_T\|_{<s>}$ & $\mathcal{O}_p\left(\frac{(\sqrt{p}+1)\|\mathcal{E}_T \|_{<s>} }{p\lambda_{min}^*+\sigma_{min}^*}\right)$ & \checkmark \\
\cline{2-4}
 $\frac{\sigma_{max}^{*}}{\sigma_{min}^{*}}=\mathcal{O}(1)$ & $\|\Er_M\|_{<s>} \prec \|\Er_T\|_{<s>}$ & $\mathcal{O}_p\left(\frac{\sqrt{p}\lambda_{max}^{*}\|\mathcal{E}_T \|_{<s>} +\sigma_{max}^{*}\|\mathcal{E}_M \|_{<s>}}{p\lambda_{min}^{*2}+\sigma_{min}^{*2}}\right)$ & \checkmark \\
\cline{2-4}
 & $\|\Er_M\|_{<s>} \succ \|\Er_T\|_{<s>}$ & $\mathcal{O}_p\left(\frac{\sqrt{p}\lambda_{max}^{*}\|\mathcal{E}_T \|_{<s>} +\sigma_{max}^{*}\|\mathcal{E}_M \|_{<s>}}{p\lambda_{min}^{*2}+\sigma_{min}^{*2}}\right)$ & inconclusive \\
\hline
\end{tabular}
\end{small}
\end{table}

\begin{remark} \change{(Sub-Gaussian noise) Similar to Remark \ref{remark:gaussian}, when the entries of the error tensor $\mathcal{E}_T$ and the error matrix $\mathcal{E}_M$ are i.i.d sub-Gaussian with mean zero and variance proxy $\sigma^2$, we can further simply the statistical error in Theorem \ref{theorem:global2}. Utilizing a similar covering number argument in \cite{tomioka2014spectral}, \cite{zhou2021partially} show that the sparse spectral norm of $\mathcal{E}_T$ and $\mathcal{E}_M$ satisfies $\|\mathcal{E}_T \|_{<s>} = \mathcal{O}_p(\sigma \sqrt{s\log(n)})$ and $\|\mathcal{E}_M \|_{<s>} = \mathcal{O}_p(\sigma \sqrt{s\log(n)})$. Therefore, the errors of the shared component in $(\ref{eq:conva-2})$ and that of the non-shared component in $(\ref{eq:convc-2})$ can be simplified as 
 $$
(\ref{eq:conva-2}) = \mathcal{O}_p\left( \frac{(\sqrt{p}\lambda_{max}^{*} + \sigma_{max}^{*}) \sigma\sqrt{s\log(n)} }{p\lambda_{min}^{*2}+\sigma_{min}^{*2}}  \right); ~~  (\ref{eq:convc-2}) = \mathcal{O}_p\left( \frac{\sigma}{\lambda_{min}^{*}} \sqrt{\frac{s\log(n)}{p} } \right).
 $$
The estimation error for the non-shared component matches with the rate in the sparse tensor model \citep{zhou2021partially}, while the estimation error for the shared component again largely improves due to the incorporation of the covariate matrix. }
\end{remark}

%%%%%%%%%%%%%%%%%%%%%%%%%%%%%%%%%%%
%%% Simulations
%%%%%%%%%%%%%%%%%%%%%%%%%%%%%%%%%%
\section{Simulations}
\label{sec:simul}
In this section we evaluate the performance of our \texttt{COSTCO} algorithm via a series of simulations. We compare it with two competing state of the arts methods \texttt{tenALSsparse} by \citet{jain2014provable} and  \texttt{OPT} by \citet{acar2011all}. \texttt{tenALSsparse} is an alternating minimization based method for tensor completion which incorporates a refinement step in the standard ALS method. In contrast to our method, \texttt{tenALSsparse} does not incorporate side covariate information in tensor completion. Comparing our algorithm to \texttt{tenALSsparse} helps to highlight the impact of incorporating addition information through coupling with a covariate matrix. It is worth noting that the original algorithm from \citet{jain2014provable} was built for the recovery of non-sparse tensors. In order to allow a fair comparison between our algorithm and theirs, we modify their original algorithm by introducing the same truncation scheme presented in Algorithm \ref{alg:jALS} to generate the sparse version of their algorithm. The second comparison method is the \texttt{OPT} algorithm by \cite{acar2011all}, which approaches the coupled matrix and tensor component recovery by solving for all components simultaneously using a gradient-based optimization approach. The all-at-once optimization method is known to be robust to rank mis-specification \citep{Song2019}, however it is computationally less efficient then ALS based methods specially when the tensor is highly missing \citep{tomasi2006}.

In the aforementioned sections, we discuss our models and theories via a third-order tensor to simply the presentation. Note that our \texttt{COSTCO} is applicable to the tensor with more than three modes. In the simulation, we generate a fourth-order tensor $\Ten^{*} \in \Rbb^{ d_1 \times 30 \times 30 \times 30}$ and a matrix $\Mat^{*} \in \Rbb^{d_1 \times 30}$. We assume that the matrix and the tensor share components across the first mode just as is the case in the aforementioned sections. In order to form the tensor $\cal T^{*}$ and the matrix $\Mat^{*}$, we draw each entry of $\Am^{*} \in \Rbb^{d_1 \times R}, \Bm^{*} \in \Rbb^{30 \times R}, \Cm^{*} \in \Rbb^{30 \times R}, \Dm^{*} \in \Rbb^{30 \times R}$ and $\Vm^{*} \in \Rbb^{30 \times R}$, from the iid standard normal distribution. We enforce sparsity to the tensor components by keeping only the top $40\%$ of the entries in each column in $\Bm^{*}, \Cm^{*} $ and $\Dm^{*} $ and set the rest of the entries to zero. In all of our simulations we consider the coupled modes $\Am^{*}$ to be dense to mimic the real data scenario in Section \ref{sec:realdata} where the coupled matrix is dense. We define $\lambda_1^{*}, \dots, \lambda_R^{*}$ and $\sigma_1^{*}, \dots, \sigma_R^{*}$ as the product of the non-normalized component norms in each mode, that is, $\lambda_r^*=\|\ac_r^{*}\|_2\times \|\bc_r^{*}\|_2\times\|\cc_r^{*}\|_2\times\|\dc_r^{*}\|_2$ and $\sigma_r^*=\|\ac_r^{*}\|_2\times \|\vc_r^{*}\|_2$. We then normalize each of the columns of $\Am^{*}$, $\Bm^{*}$ ,$\Cm^{*}$ ,$\Dm^{*}$ ,$\Vm^{*}$  to unit norm. To illustrate, the first mode component matrix $\Am^{*}$ becomes $\Am^{*}=[\frac{\ac_r^*}{\|\ac_r^*\|_2},\cdots,\frac{\ac_R^*}{\|\ac_R^*\|_2}]$. The sparse tensor $\Ten^{*}$ and matrix $\Mat^{*}$ are then formed as $ \Ten^{*} = \sum \limits_{r \in [R]} \lambda_r^{*} \ac_r^{*} \otimes \bc_r^{*} \otimes \cc_r^{*} \otimes \dc_r^{*} $ and $
	\Mat^{*} = \sum \limits_{r \in [R]} \sigma_r^{*} \ac_r^{*} \otimes \vc_r^{*}$. We then add noise to the tensor and matrix using the following setup $\Ten = \Ten^{*} +\eta_T \mathcal{N}_T  \frac{\| \Ten^{*} \|_F}{\| \mathcal{N}_T \|_F}$ and $ \Mat =\Mat^{*} + \eta_M \mathcal{N}_M  \frac{\| \Mat^{*} \|_F}{\| \mathcal{N}_M \|_F}$, where $\mathcal{N}_T$ and $\mathcal{N}_M$ are a tensor and a matrix of the same size as $\Ten^{*}$ and $\Mat^{*}$ respectively, whose entries are generated from the standard normal distribution. A similar noise generation procedure has been considered in \citet{acar2011all}.
 We simulate the uniformly missing at random pattern in the tensor data by generating entries of the reveal tensor $\bm{\Omega} \in \Rbb^{d_1 \times 30 \times 30 \times 30}$ from the binomial distribution with reveal probability $p$.  The sparse and noisy tensor $P_{\Omega}(\Ten)$  with missing data is finally obtained as $P_{\Omega}(\Ten) = \Ten \ast \bm{\Omega}$, where $\ast$ is the element-wise multiplication.

%\noindent\textbf{Performance Metrics:} 
To assess the goodness of fit for the tensor and tensor components recovery, we use the normalized Frobenius norm of the difference between the recovered component and the true component. We compute the tensor estimation error, the tensor component error and tensor weights error as:
\begin{align*}
     \text{tensor error }:= {\| \Ten^{*} - \Ten \|_F} / {\| \Ten^{*} \|_F};\quad &\text{component error }:= {\| \Um^{*} - \Um \|_F/\| \Um^{*} \|_F}; \\ 
     \text{weight error}:= &\| \bm{\lambda^{*}} - \bm{\lambda} \|_2/\|\bm{\lambda^{*}} \|_2,\stepcounter{equation}\tag{\theequation}\label{eq:estm_error}
\end{align*}
where $\Ten, \Um,$ are the estimated tensor and tensor components with $\Um \in \{\Am, \Bm, \Cm, \Dm\}$, and  $\bm{\lambda}:=(\lambda_1, \cdots, \lambda_R)^{\top}$ is the vector of estimated tensor weights returned by Algorithm \ref{alg:jALS}. In all simulations we return the mean error of 30 replicas of each experiment. Throughout all the experiments, we set the maximum number of iterations $\tau$ to be $200$, the tolerance $tol$ in Algorithm \ref{alg:jALS} is set to be $1 e^{-7}$. To avoid bad local solutions, we conduct $10$ initializations for each replicate in all methods. We set the tuning range for the rank $R$ to be $\{1,2,3,4,5\}$. The tuning range for the sparsity is set to be $\{20\%,40\%,60\%,80\%,90\%, 100\%\}$, each value representing the percentage of non-zero entries in the latent components as performed on Lines 9 and 11 of Algorithm \ref{alg:jALS}. Note that in addition to a series of simulations considered here, in Section \ref{sup:simulation} of the supplementary material, we provide two additional simulations to investigate the practical effect of dimension size of the shared component and the rank on our \texttt{COSTCO} algorithm.

\subsection{Missing Percentage}
In this first simulation we consider the case with varying levels of missing percentages. We set the dimension of the couple mode to be $d_1=30$ and therefore generate $P_{\Omega}(\Ten) \in \Rbb^{30 \times 30 \times 30 \times 30}$ . We set the rank to be $R=2$ and the noise level $\eta_T$, $\eta_M$ to be both 0.001. We measure the recovery error under four different settings of the reveal probability parameter $p = \{0.2, 0.1, 0.05, 0.01 \}$. In other words, $ 80 \%$, $ 90\%$, $95\%$ and $99\%$ of the tensor entries are missing in each setting. Table \ref{tab:miss} indicate that under all varying missing probability, our \texttt{COSTCO} algorithm provides a better fit in tensor recovery relative to \texttt{tenALSsparse} and \texttt{OPT}. Notably, with a higher level of missing data, missing percentage $\geq 90$ \texttt{COSTCO} significantly outperforms both \texttt{tenALSsparse} and \texttt{OPT} methods of tensor recovery. This is more evident when we compare our algorithm to \texttt{tenALSsparse} for the case where missing percentage ranges from $90\%$ to $98\%$; in these scenarios the recovery error of \texttt{COSTCO} is at least 10 folds better than that of \texttt{tenALSsparse}. This agrees with the two advantages of incorporating covariate information into tensor completion as we discussed in the theoretical results: (1) allowing higher missing percentage; (2) reducing estimation errors. Moreover, we notice that the estimation error for the shared component Comp $\ddot {\Am}$ is better than that of the non-shared components. This also aligns with the theoretical result which shows that the recovery of the couple component improves over that of non-coupled components due to additional covariate information. Finally, although \texttt{OPT} also uses coupling, it underperforms compared to \texttt{COSTCO} because the all at once optimization method suffers with unstable gradient when the missing entry percentage is large.

%%%%%%%%%%%%%%%%%%%%%%%%%%%%%%%
%% Missing data simulation test  table 
%%%%%%%%%%%%%%%%%%%%%%%%%%%%%%%%%%%%%%%
\begin{table}[h!]
\centering
\begin{small}
\caption{Estimation errors with varying missing percentages. Reported values are the average and standard error (in parentheses) of tensor, tensor components and weight recovery error based on 30 data replications. \texttt{COSTCO}: the proposed method; \texttt{tenALSsparse}: sparse version of the tensor completion method by \cite{jain2014provable}; \texttt{OPT}: the gradient based all at once optimization method of \cite{acar2011all}; symbol $(\ddot{A})$ used to put shared tensor-matrix component $\Am$ in emphasis.}
\label{tab:miss}
%\vskip 1 em
\begin{tabular}{cc|ccc} \hline
 & &  \multicolumn{3}{c}{Estimation Error} \\ 
 \cline{3-5}
  Missing Percent & Component &  \texttt{COSTCO} & \texttt{tenALSsparse}& \texttt{OPT} \\ \hline
80\% & $\Ten$ &  \textbf{3.38e-05 (2.36e-12)} &	3.66e-05 (2.73e-12) &	3.56e-05 (2.31e-12) \\
 & Comp $\ddot {\Am}$ &  \textbf{1.52e-05 (2.37e-12)} & 2.22e-05 (3.93e-12) & \textbf{1.52e-05 (2.36e-12)} \\ 
 & Comp $\Bm$ & \textbf{2.12e-05 (4.39e-12)}& 2.13e-05 (3.64e-12) & 2.26e-05 (5.05e-12) \\ 
 & Comp $\Cm$ &  \textbf{1.98e-05 (4.69e-12)} & 1.99e-05 (4.83e-12) & 2.24e-05 (4.35e-12) \\
 & Comp $\Dm$ &  \textbf{2.17e-05 (2.92e-12)} & 2.18e-05 (2.78e-12) & 2.26e-05 (2.99e-12) \\ 
 & $\bm{\lambda}$ &  1.18e-06 (4.67e-13) & \textbf{1.17e-06 (4.95e-13)} & 1.18e-06 (4.67e-13) \\
  \hline
%   &  &  &  &  \\
90\% & $\Ten$ & \textbf{3.93e-05 (6.12e-12)} & 4.47e-02 (2.71e-11) & 4.94e-05 (6.07e-12) \\
 & Comp $\ddot {\Am}$ & \textbf{1.80e-05 (2.79e-12)} & 5.65e-02 (2.74e-11) & \textbf{1.80e-05 (2.82e-12)} \\ 
 & Comp $\Bm$ & \textbf{2.16e-05 (1.31e-11)} & 4.84e-02 (2.02e-11) & 3.17e-05 (1.31e-11) \\ 
 & Comp $\Cm$ & \textbf{2.12e-05 (9.54e-12)} & 4.96e-02 (3.22e-11) & 3.13e-05 (9.75e-12) \\ 
 & Comp $\Dm$ & \textbf{2.17e-05 (1.38e-11)} & 5.79e-02 (2.00e-11) & 3.18e-05 (1.39e-11) \\ 
 & $\bm{\lambda}$ & \textbf{1.65e-06 (7.98e-13)} & 4.84e-02 (8.31e-13) & 1.65e-06 (7.98e-13) \\ 
\hline
% &  &  &  &  \\
 95\% & $\Ten$ & \textbf{5.69e-05 (1.92e-11)} & 1.19e-01 (8.70e-03) & 6.93e-05 (1.90e-11) \\
 & Comp $\ddot {\Am}$ & 1.92e-05 (5.60e-12) & 1.44e-01 (2.01e-02) & \textbf{1.50e-05 (6.30e-12)} \\ 
 & Comp $\Bm$ & \textbf{3.44e-05 (2.29e-11)} & 1.28e-01 (1.61e-02) & 4.45e-05 (2.30e-11) \\ 
 & Comp $\Cm$ & \textbf{3.39e-05 (3.36e-11)} & 1.30e-01 (1.02e-02) & 4.39e-05 (3.34e-11) \\ 
 & Comp $\Dm$ & \textbf{3.74e-05 (1.84e-11)} & 1.40e-01 (1.39e-02) & 4.74e-05 (1.80e-11) \\ 
 & $\bm{\lambda}$ & \textbf{1.26e-06 (8.99e-13)} & 1.25e-01 (1.08e-02) & 1.76e-06 (8.99e-13) \\ 
  \hline
% &  &  &  &  \\
 98\% & $\Ten$ & \textbf{2.36e-02 (3.50e-11)} &	5.05e-01 (1.75e-02) &	5.02e-02 (1.98e-02)\\
 & Comp $\ddot {\Am}$ & \textbf{2.17e-02 (1.18e-11)} & 6.58e-01 (2.03e-02) & 6.87e-02 (2.61e-03) \\ 
 & Comp $\Bm$ & \textbf{2.63e-02 (5.60e-11)} & 6.18e-01 (1.29e-02) & 6.31e-02 (2.95e-02) \\ 
 & Comp $\Cm$ & \textbf{2.58e-02 (5.81e-11)} & 5.89e-01 (1.49e-02) & 6.27e-02 (3.86e-02) \\ 
 & Comp $\Dm$ & \textbf{2.16e-02 (5.39e-11)} & 5.94e-01 (2.16e-02) & 6.96e-02 (2.03e-02) \\ 
 & $\bm{\lambda}$ & \textbf{2.14e-02 (5.67e-13)} & 5.19e-01 (1.75e-02) & 5.00e-02 (2.14e-02) \\ 
  \hline
% &  &  &  &  \\
99\% & $\Ten$ & \textbf{7.13e-01 (5.93e-11)} & 9.99e-01 (5.35e-02) & 	8.80e-01 (2.33e-02) \\
 & Comp $\ddot {\Am}$ & \textbf{3.60e-01 (1.28e-10)} & 1.17e+00 (1.17e-01) & 4.17e-01 (4.39e-02) \\ 
 & Comp $\Bm$ & \textbf{7.40e-01 (1.04e-10)} & 1.14e+00 (9.65e-02) & 7.94e-01 (3.70e-02) \\ 
 & Comp $\Cm$ & \textbf{8.25e-01 (3.75e-11)} & 	1.17e+00 (9.15e-02) & 9.14e-01 (3.65e-02) \\ 
  & Comp $\Dm$ & \textbf{5.90e-01 (4.57e-11)} & 9.77e-01 (9.83e-02) & 7.12e-01 (4.51e-02) \\ 
  & $\bm{\lambda}$ & \textbf{6.48e-01 (5.73e-11)} & 9.77e-01 (6.04e-02) & 8.68e-01 (2.33e-02) \\ 
  \hline 
\end{tabular}
\end{small}
\end{table}

%%%%%%%%%%%%%%%%%%%%%%%%%%%%%%%
\subsection{Noise Level}
In the next set of experiments we vary the noise level parameter for the tensor $\eta_T$ and noise level for the matrix $\eta_M$ to test algorithms' robustness to noise. These two parameters control the signal-to-noise ratio in the model. The missing probability for these experiments is set to $90\%$ and tensor rank and sparsity of the true tensor are set to $R = 2$ and $60\%$ respectively.

%%%%%%%%%%%%%%%%%%%%%%%%%%%%%%%
%% Tensor noise simulation test  table 
%%%%%%%%%%%%%%%%%%%%%%%%%%%%%%%%%%%%%%%
\begin{table}[h!]
\centering
\begin{small}
\caption{Estimation errors with varying noise levels of error matrix and error tensor. Reported values are the average and standard error (in parentheses) of estimation errors. \texttt{COSTCO}: the proposed method; \texttt{tenALSsparse}: sparse version of the tensor completion method by \cite{jain2014provable}; \texttt{OPT}: the gradient based all at once optimization method of \cite{acar2011all}.}%; symbol $(\ddot{A})$ used to put shared tensor-matrix component $\Am$ in emphasis.}
\label{tab:noise}
%\vskip 1 em
\begin{tabular}{cc|ccc} \hline
 & &  \multicolumn{3}{c}{Estimation Error} \\ 
 \cline{3-5}
 Noise Level & Component &  \texttt{COSTCO} & \texttt{tenALSsparse}& \texttt{OPT} \\ \hline
&	 $\Ten$ & \textbf{2.74e-04	(7.31e-10)}	&	5.37e-04	(1.00e-09)	&	4.74e-04	(7.31e-10)	\\
&	 Comp $\ddot {\Am}$ & \textbf{1.05e-04	(2.24e-10)}	&	3.17e-04	(1.13e-09)	&	1.05e-04	(2.24e-10)	\\
$\eta_M=0.001$&	 Comp ${\Bm}$ & \textbf{2.13e-04	(8.03e-10)}	&	3.10e-04	(4.72e-10)	&	3.13e-04	(8.03e-10)	\\
$\eta_T= 0.01$&	 Comp ${\Cm}$ & \textbf{2.15e-04	(1.33e-09)}	&	3.14e-04	(1.35e-09)	&	3.15e-04	(1.33e-09)	\\
&	 Comp ${\Dm}$ & \textbf{2.21e-04 (1.43e-09)}	&	3.22e-04	(1.69e-09)	&	3.21e-04	(1.43e-09)	\\
&	 $\bm{\lambda}$ & \textbf{1.41e-05	(6.77e-11)}	&	1.48e-05	(7.44e-11)	&	\textbf{1.41e-05	(6.77e-11)}	\\
%   &  &  &  &  \\
  \hline
%   &  &  &  &  \\
&$\Ten$ & \textbf{	2.73e-03	(5.50e-08)}	&	5.36e-03	(8.04e-08)	&	4.73e-03	(5.50e-08)	\\
&	 Comp $\ddot {\Am}$ & \textbf{	1.06e-03	(2.39e-08)}	&	3.16e-03	(1.87e-07)	&	1.06e-03	(2.39e-08)	\\
$\eta_M=0.001$&	 Comp ${\Bm}$ & \textbf{	2.03e-03	(1.25e-07)}	&	3.00e-03	(1.66e-07)	&	3.03e-03	(1.25e-07)	\\
$\eta_T= 0.1$ &	 Comp ${\Cm}$ & \textbf{	2.15e-03	(6.21e-08)}	&	3.10e-03	(3.68e-08)	&	3.15e-03	(6.21e-08)	\\
&	 Comp ${\Dm}$ & \textbf{	2.20e-03	(1.02e-07)}	&	3.23e-03	(1.01e-07)	&	3.20e-03	(1.02e-07)	\\
&	 $\bm{\lambda}$ & 	1.52e-04	(7.07e-09)	&	\textbf{1.46e-04	(6.09e-09)	}&	1.52e-04	(7.07e-09)	\\
%   &  &  &  &  \\
  \hline
%   &  &  &  &  \\
&	 $\Ten$ & \textbf{3.88e-04	(5.55e-10)}	&	5.35e-04	(6.41e-10)	&	4.88e-04	(5.55e-10)	\\
&	 Comp $\ddot {\Am}$ & \textbf{1.74e-04	(3.79e-10)}	&	3.21e-04	(8.24e-10)	&	1.74e-04	(3.82e-10)	\\
$\eta_M=0.01$&	 Comp ${\Bm}$ & \textbf{2.17e-04	(9.18e-10)}	&	3.14e-04	(1.10e-09)	&	3.17e-04	(9.18e-10)	\\
$\eta_T= 0.001$&	 Comp ${\Cm}$ & \textbf{2.16e-04	(1.13e-09)}	&	3.16e-04	(1.44e-09)	&	3.16e-04	(1.13e-09)	\\
&	 Comp ${\Dm}$ & \textbf{2.07e-04	(8.39e-10)}	&	3.02e-04	(8.70e-10)	&	3.07e-04	(8.39e-10)	\\
&	 $\bm{\lambda}$ & \textbf{1.49e-05	(7.21e-11)}	&	1.53e-05	(6.63e-11)	&	\textbf{1.49e-05	(7.21e-11)}	\\
%   &  &  &  &  \\
  \hline
%   &  &  &  &  \\
 & $\Ten$ 	&	9.75e-04	(1.60e-08)	&	\textbf{5.37e-04	(1.36e-09)}	&	1.28e-03	(1.60e-08)	\\
&	 Comp $\ddot {\Am}$ 	&	1.39e-03	(2.27e-08)	&	\textbf{3.17e-04	(1.16e-09)}	&	1.39e-03	(2.27e-08)	\\
 $\eta_M=0.1$&	 Comp ${\Bm}$ 	&	\textbf{2.20e-04	(1.11e-09)}	&	3.09e-04	(1.02e-09)	&	3.21e-04	(1.12e-09)	\\
$\eta_T= 0.001$&	 Comp ${\Cm}$ 	&	\textbf{2.29e-04	(1.30e-09)}	&	3.19e-04	(1.01e-09)	&	3.23e-04	(1.32e-09)	\\
&	 Comp ${\Dm}$ 	&	\textbf{2.24e-04	(1.20e-09)	}&	3.12e-04	(1.27e-09)	&	3.25e-04	(1.20e-09)	\\
&	 $\bm{\lambda}$ 	&	\textbf{1.26e-05	(7.94e-11)}	&	1.27e-05	(7.62e-11)	&	\textbf{1.26e-05	(7.94e-11)}	\\
  \hline 
\end{tabular}
\end{small}
\end{table}

As can be seen in Table \ref{tab:noise}, when the tensor noise $\eta_T$ is greater than that of the matrix noise $\eta_M$, our algorithm outperforms the two competing methods with a large gap in recovery error. Even when the matrix has a slightly larger noise level than the tensor $(\eta_M=0.01,\eta_T=0.001)$, \texttt{COSTCO} still outperforms the other two algorithms. It shows that in high missing data regime coupling a matrix that has a slightly larger noise than the tensor still provides enough information to improve the tensor recovery rate. On the other hand, when the matrix noise level is much higher than that of the tensor ($\eta_M=0.1,\eta_T=0.001$ in Table \ref{tab:noise}), we observe that our algorithm \texttt{COSTCO} and the other coupled algorithm \texttt{OPT} are inferior compared to \texttt{tenALSsparse}. In this case, the recovery of the shared component $\Am$ suffers the most in \texttt{COSTCO} and \texttt{OPT} and is responsible for the inferior tensor recovery error compared to \texttt{tenALSsparse} which does not use the coupled matrix. This is expected as a matrix with much larger noise than that of a tensor no longer brings in enough signals in the coupling and therefore makes the tensor completion problem harder than when the matrix is completed omitted from the model. Finally, an interesting phenomenon is that the noise level of the error matrix $\eta_M$ only affects the estimation error of the shared component but not those of the non-shared components. To see it, in the last two settings in Table \ref{tab:noise}, when $\eta_T$ is fixed and $\eta_M$ increases, only the recovery accuracy of the shared component $\Am$ significantly drops, but those of the non-shared components have no significant changes. However, in the first two settings in Table \ref{tab:noise}, when $\eta_M$ is fixed and $\eta_T$ increases, the recovery accuracy of both shared and non-shared components significantly drops. These findings agree well with our theoretical results in Theorem \ref{theorem:global2}.

%%%%%%%%%%%%%%%%%%%%%%%%%%%%%%
%Real data analysis
%%%%%%%%%%%%%%%%%%%%%%%%%%%%%%%
\section{Real Data Analysis}
\label{sec:realdata}

We apply our \texttt{COSTCO} method to an advertisement (ad) data to showcase its practical advantages. \texttt{COSTCO} makes use of multiple sources of ad data to extract the ad latent component which is a comprehensive representation of ads. We demonstrate that the obtained ad latent components are able to deliver interesting ad clustering results that are not achievable by a stand-alone method.

Online advertising is a type of marketing strategy which uses internet to promote a given product to potential customers. Extracting patterns in data gathered from online advertisement allows ad platforms and companies to churn data into knowledge which is then used to improve customer satisfaction. Clustering algorithms have been applied to the ad data to discover ad or user clusters for better ad targeting. After computing the similarity between the new ad and each ad cluster, the ad agency can determine whether a new ad should be assigned to a specific user group. Most ad-user clustering research focuses on a single correlation data. What makes our method different is that we not only have a third-order user-by-ad-by-device click tensor data but we also possess additional information which describe specific features of ads. Our \texttt{COSTCO} algorithm uses both click tensor data and ad matrix data to extract the ad latent component for better ad clustering.

The data we analyze in this section is advertising data collected from a major internet company for 4 weeks in May-June 2016. A user preference tensor was obtained by tracking the behavior of 1000 users on 140 ads accessed through 3 different devices. \change{The $1000 \times 140 \times3$ tensor is formed by computing the click-through-rate (CTR) of each (user, ad, device) triplet over the four weeks period; which is the number of times a user has clicked an ad from a certain device divided by the number of times the user has seen that ad from the specific device. Each CTR tensor entry was aggregated over multiple publishers (homepage, news, sports, finance, weather, fashion, etc) during these 4 weeks for the same (user, ad, device) triplet.} As illustrated in Figure \ref{fig:reveal1}, this ad CTR tensor has $96 \%$ missing entries and is highly sparse with only $40\%$ of the revealed entries being nonzero. A missing entry in the ad CTR data occurs when a given user is not presented with a certain ad from a specific device, while zeros (sparsity) in the ad CTR data are used to represent user choosing not to interact with an ad that was presented to them on a specific device.

%\iffalse
\begin{figure}[h!]
        \centering
    \begin{subfigure}[t]{0.32\textwidth}
        \centering
        \includegraphics[width=\linewidth]{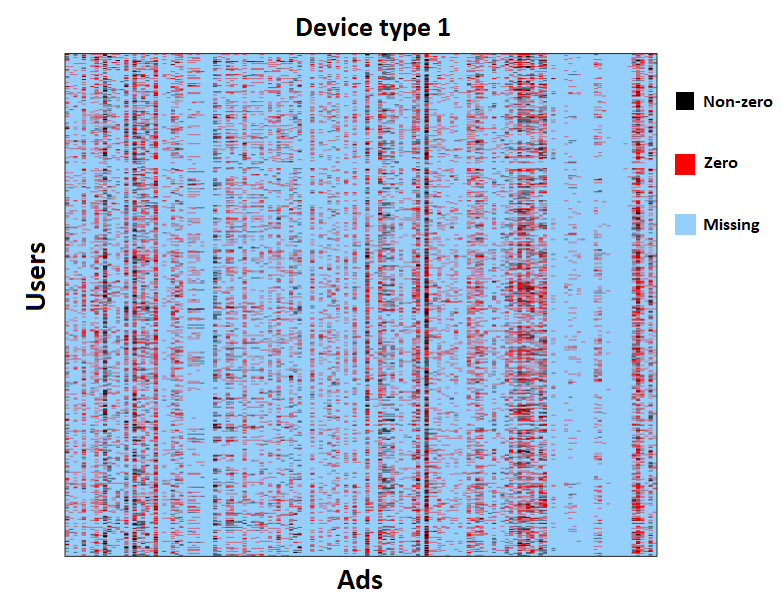} 
         %\label{fig:reveal}
    \end{subfigure}
    \hfill
    \begin{subfigure}[t]{0.32\textwidth}
        \centering
        \includegraphics[width=\linewidth]{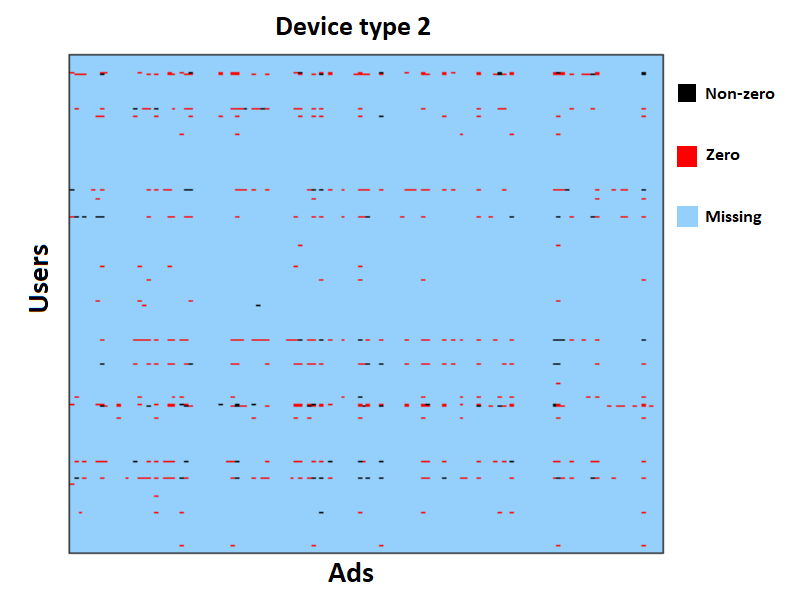} 
         \label{fig:reveal2}
    \end{subfigure}
    \hfill
    \begin{subfigure}[t]{0.32\textwidth}
        \centering
        \includegraphics[width=\linewidth]{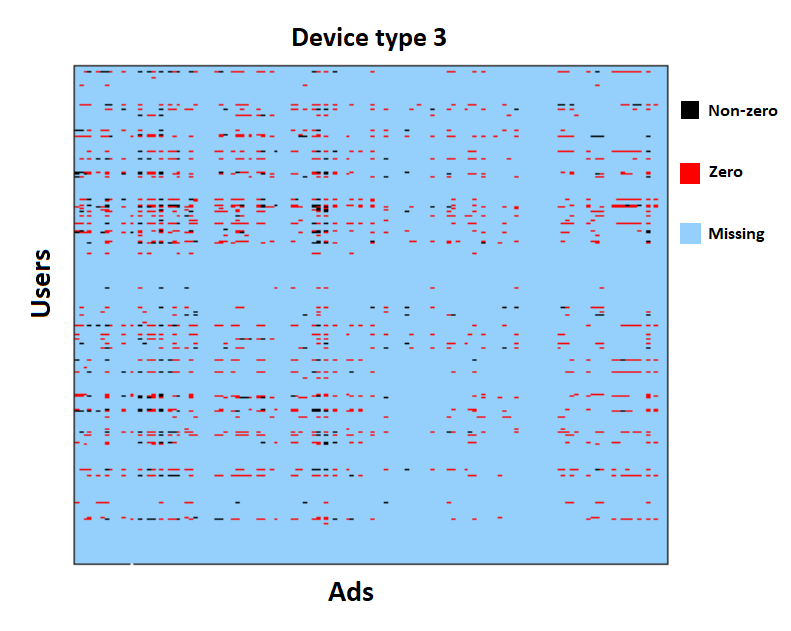} 
         \label{fig:reveal3}
    \end{subfigure}
        \caption{Illustration of missing data and sparsity in our ad CTR tensor.}
         \label{fig:reveal1} %(\subref{fig:timing1}) you can see a  green square....}
\end{figure}
%\fi

Beside the ad CTR tensor, we also have access to the ad text raw data that store the content of all ads. We use Latent Dirichlet Allocation (LDA) \citep{blei2003latent} to process the ad text data. LDA is an unsupervised topic modeling algorithm that attempts to describe a set of text observations as a mixture of different topics. We first follow \citet{blei2003latent} to tune the parameters of LDA such as the number of topics and the Dirichlet distribution parameter that give the best trade-off between low perplexity value and efficient computing time. The best perplexity is obtained for $20$ topics. This means that all the $140$ advertisement data can be considered as a combination of 20 topics. Due to space constraints, we illustrate an example of 7 out of 20 topics in Table \ref{tab:topWords}, and only display the top 10 words for each of the 7 topics returned by LDA. Each topic column was labeled based on overall meaning of the top words. Once trained, LDA returns a matrix that contains the proportion of topics in each ad. We use this matrix of proportions of dimension $\Rbb^{140\times 20}$ as the ad covariate matrix that will be used jointly with the ad CTR tensor to obtain ad latent components in our \texttt{COSTCO} algorithm.

%%%%%%%%%%%%%%%%%%%%%%%%%%%%%%%%
%%%%%%%top words
%%%%%%%%%%%%%%%%%%%%%%%%%%%%%%
\begin{centering}
\begin{table}[h!]
%\centering
\caption{Top ten words for 7 chosen topics. Top words were obtained through LDA.}
\label{tab:topWords}
\scalebox{0.85}{
\begin{tabular}{c|c|c|c|c|c|c|c}
 \hline
\textbf{Topics} & \multicolumn{1}{c|}{\textbf{Ride} } & \multicolumn{1}{c|}{\textbf{Gaming }} & \multicolumn{1}{c|}{\textbf{Security}} & \multicolumn{1}{c|}{\textbf{Mortgage}} & \multicolumn{1}{c|}{\textbf{Insurance}} &  \multicolumn{1}{c|}{\textbf{Online dating}}& \multicolumn{1}{c}{\textbf{Fashion retail }}\\
 \hline
 \parbox[t]{2mm}{\multirow{9}{*}{\rotatebox[origin=c]{90}{\textbf{Top Words}}}} 
&	 uber 	&	 game 	&	 vivint 	&	 mortgage 	&	 get 	&	 	 single 	&	 buy 	\\
&	 pay 	&	 controller 	&	 home 	&	 apr 	&	 insurance 	&		 pic 	&	 sale 	\\
&	 car 	&	 experience	&	 front 	&	 payment 	&	 less 	&	 	 man 	&	 gilt 	\\
&	 people 	&	 gameplay 	&	 security 	&	 free 	&	 see 	&		 profile 	&	 zulily 	\\
&	 weekly 	&	 accessory 	&	 smart 	&	 new 	&	 month 	&	 	 click 	&	 lulus 	\\
&	 fare	&	 ebay 	&	 call 	&	 arm 	&	 drive 	&	 	 meet 	&	 charlotterusse 	\\
&	 ride 	&	 level 	&	 control 	&	 quotes 	&	 day 	&	 	 browse 	&	 neimanmarcus 	\\
&	 give 	&	 time 	&	 camera 	&	 calculate 	&	 miles 	&	 look 	&	 maurices 	\\
&	 work	&	 joystick 	&	 adt 	&	 easy 	&	 low 	&	 free 	&	 lastcall	\\
&	 drive 	&	 wide	&	 look	&	 process 	&	 qualify 	&	 	 pay 	&	 spring 	\\
\hline
	\end{tabular}
	}
\end{table}
\end{centering}

We first evaluate the tensor recovery error by randomly splitting the observed tensor entries into $80\%$ training and $20\%$ testing. Let $\hat{\cal T}$ indicate the recovered tensor from the training set. We use $\hat{\cal T}$ for training and compute the recovery error on the testing set. The metrics used to access the recovery error of the tensor is defined as ${\| P_{\Omega_{Test}}( \Ten - \hat{\Ten} ) \|_F} / {\|P_{\Omega_{Test}} (\Ten) \|_F}$, where $P_{\Omega_{Test}} (\Ten) = \bm{\Omega}_{Test} \ast \Ten $ with $\bm{\Omega}_{Test}$ being a binary tensor of the same size as $\Ten$ that has ones on the test entries and zeros elsewhere. The tensor recovery error for \texttt{COSTCO} is $0.825$, leading to $23\%$ accuracy improvement over the baseline \texttt{tenALSsparse} whose error is $1.083$. \change{We also implement a covariate-assisted version of the neural tensor factorization \citep{wu2019neural} via Tensorflow. Specifically, user id, ad id, and device id are first converted to one-hot encodings, which are then fed into three parallel embedding layers. The concatenation of these and the covariates of the corresponding advertisement is then fed into a 3-layer perceptron to learn its representation, which is subsequently used as features to predict the associated CTR entries. The implementation details are included in Section \ref{sec: sup_neural} in the supplementary. The tensor recovery error of this covariate-assisted neural tensor factorization method is $0.910$, which is better than the baseline \texttt{tenALSsparse} but is still inferior to our \texttt{COSTCO}.} This highlights the benefit of fusing the ad content matrix to the ad CTR tensor. The \texttt{OPT} algorithm was not used for comparison as the algorithm optimization package failed with error messages after multiple trials on this data. We conjecture this is due to the unstable performance of the all at once optimization when the missing percentage is very high.

We then compare the ad latent components returned from \texttt{COSTCO} and \texttt{tenALSsparse} in Figure \ref{fig:clustReal}. As a comparison, we also include the result of SVD which directly decomposes the ad covariate matrix data. The ad clusters shown in Figure \ref{fig:clustReal} are obtained by applying the K-means clustering algorithm to the ad latent component data from each method. As shown in Figure \ref{fig:clustReal}, the first two columns of the latent components returned from our \texttt{COSTCO} show a clear clustering structure with 5 clusters. On the other hand, the ad components extracted from \texttt{tenALSsparse} are all clustered around zeros. This is because the ad CTR tensor is highly sparse and the latent components based on decomposing the tensor itself contain many small values. Therefore, ad clusters generated using \texttt{tenALSsparse} tend to have very large and very small clusters.% that do not contain any understandable relationship between the ads. 

\begin{figure}[h!]
        \centering
    \begin{subfigure}[t]{0.32\textwidth}
        \centering
        \includegraphics[width=\linewidth]{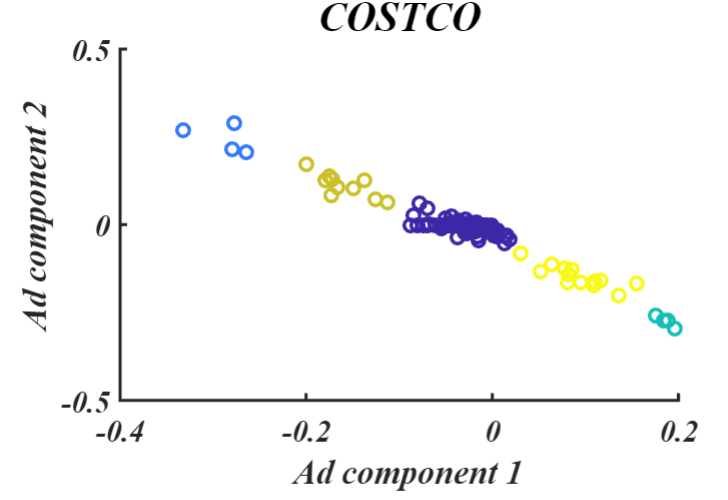} 
         \label{fig:costco_cluster}
    \end{subfigure}
    \hfill
    \begin{subfigure}[t]{0.32\textwidth}
        \centering
        \includegraphics[width=\linewidth]{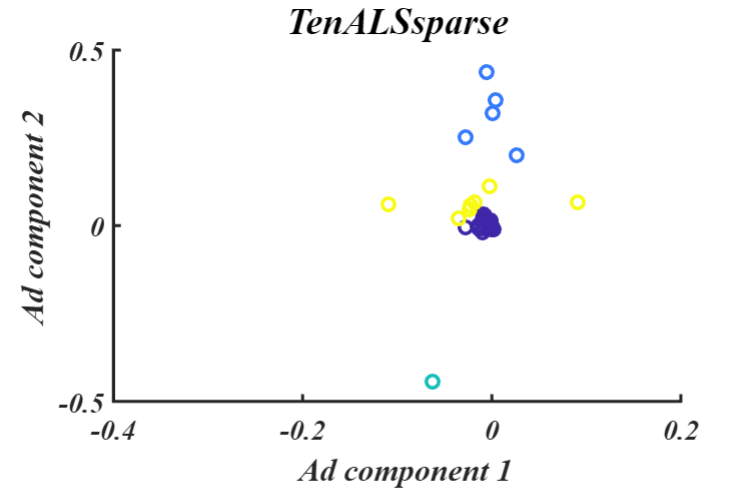} 
         \label{fig:ALSsp_cluster}
    \end{subfigure}
    \hfill
    \begin{subfigure}[t]{0.32\textwidth}
        \centering
        \includegraphics[width=\linewidth]{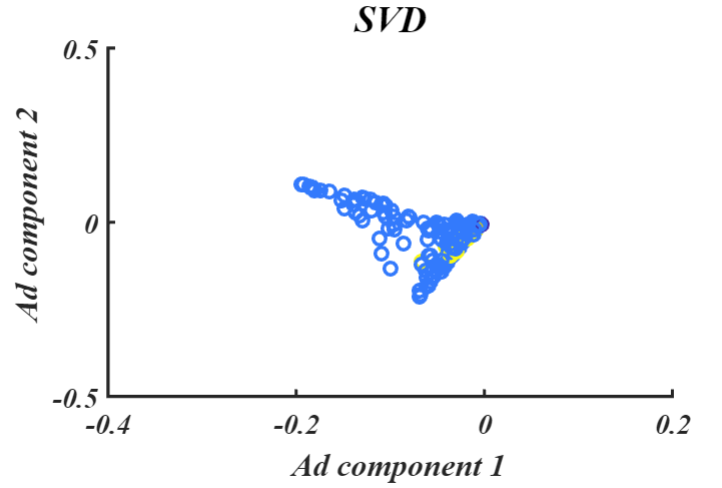} 
         \label{fig:SVD_cluster}
    \end{subfigure}
          \caption{Scatter plot of the ad latent components obtained from three methods. Different clusters are represented via different colors.}
         \label{fig:clustReal} %(\subref{fig:timing1}) you can see a  green square....}
\end{figure}

\red{Finally, after obtaining the ad clusters, we visualize the ad topics from the each cluster in Figure \ref{fig:Clust}. Specifically, for all ads assigned in each cluster, we apply the topic modeling method LDA to these ad texts to obtain their topics. For example, the ad cluster 1 from our \texttt{COSTCO} algorithm consists of four interesting topics, represented as four boxes in the first row of Figure \ref{fig:Clust}. Within each topic, the top five words are highlighted in green in our \texttt{COSTCO} method.} 
 \begin{figure}[h!]
   \centering
         \includegraphics[width=0.7\textwidth]{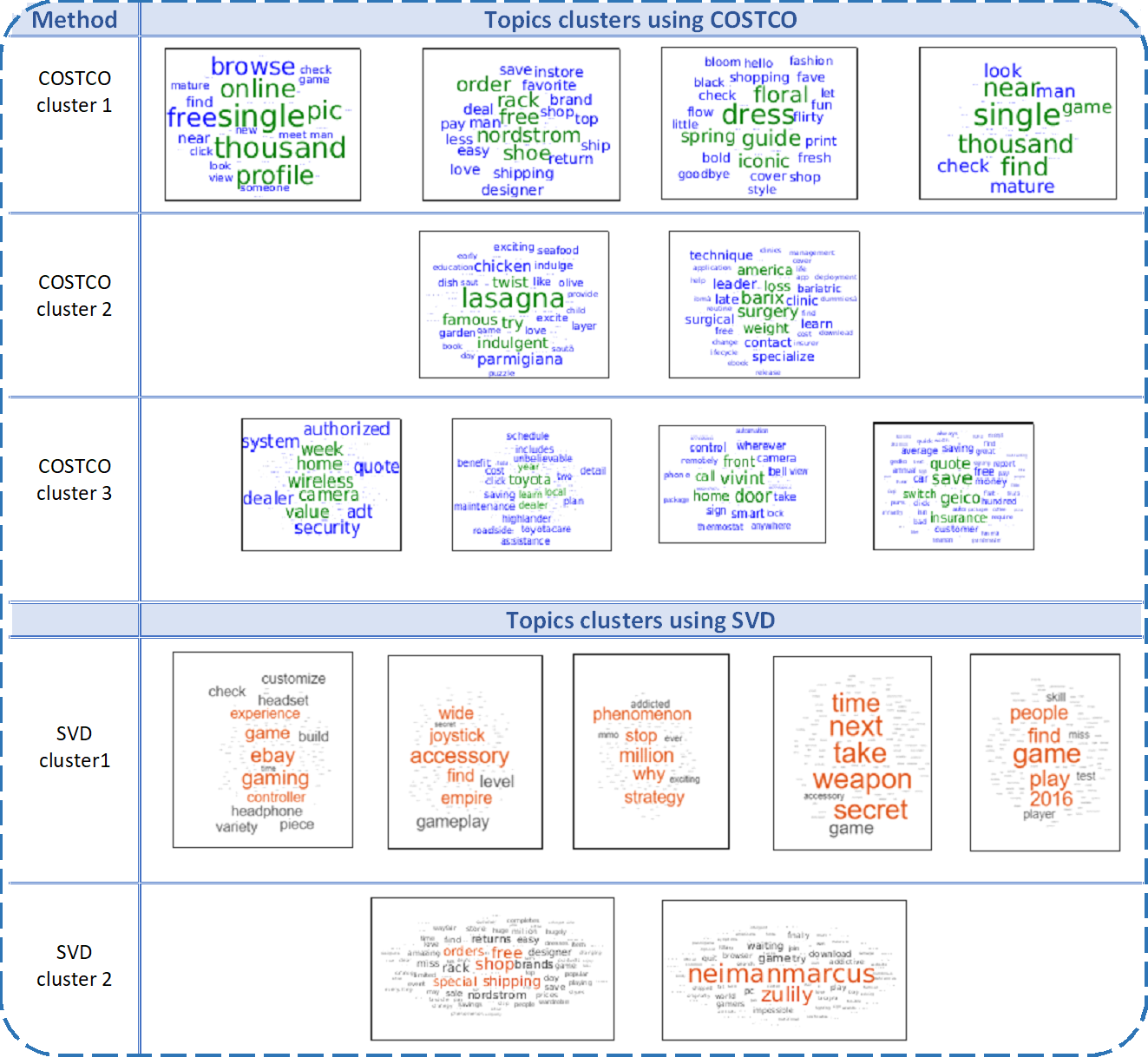}
             \caption{Result of ad clusters obtained using different methods}
         \label{fig:Clust}
 \end{figure}
 
Figure \ref{fig:Clust} demonstrates some interesting ad clustering results obtained from our \texttt{COSTCO} algorithm which links different ad industries into the same cluster. For example based on cluster 1 from \texttt{COSTCO}, ads about male and female online dating are clustered together with ads about women retail stores and man clothing accessories. In cluster 2 from \texttt{COSTCO}, ads about weight lost and weight lost surgery are clustered together with ads about gourmet cuisine and restaurant which indicates that users who interact with weight loss ads are also interested in nutrition related ads. Cluster 3 of \texttt{COSTCO} contains ads about house mortgage, home security devices, auto, home and auto insurance, house weather control devices which indicates that users that are homeowners tend to be interested in home and auto related things. These interesting clusters are not obtained in the SVD method nor the \texttt{tenALSsparse} method. The clusters from SVD are solely related to the topic of each ad as shown in Figure \ref{fig:Clust} and the clusters from \texttt{tenALSsparse} are highly unbalanced and do not contain any understandable relationship between ads. These clustering results illustrate the practical value of our \texttt{COSTCO} method. By incorporating ad covariate matrix into the completion of the ad CTR tensor, we are able to obtain a more synthetic description of ads and find interesting links between different advertising industries, which directly helps the marketing team to strategize the ad planing procedure accordingly for better ad targeting.

\section*{Acknowledgment}
The authors thank the editor Professor Ian McKeague, the associate editor and two anonymous reviewers for their valuable comments and suggestions which led to a much improved paper. Will Wei Sun's research was partially supported by ONR grant N00014-18-1-2759. Any opinions, findings, and conclusions or recommendations expressed in this material are those of the authors and do not necessarily reflect the views of the Office of Naval Research.

%%%%%%%%%%%%%%%%%%%%%%%%%%%%%%%%%%%%%%%%%%%%%%%%%%%%%%%%%%
%%%%%%%%%%%%%%%%%%%%%%%%%%%%%%%%%%%%%%%%%%%%%%%%%%%

\baselineskip=10pt
\bibliographystyle{ims}
\bibliography{costco}

\newpage
\baselineskip=18pt

\setcounter{page}{1}
\setcounter{section}{0}
\setcounter{subsection}{0}
\setcounter{equation}{0}
\pagenumbering{arabic}
\renewcommand{\thesubsection}{S.\arabic{subsection}}
\renewcommand\theequation{S\arabic{equation}}
\renewcommand\thetable{S\arabic{table}}
\renewcommand\thefigure{S\arabic{figure}}

\begin{center}
\Large{\bf Supplementary Material for \\Covariate-assisted Sparse Tensor Completion}
\end{center}
\smallskip

\noindent
This supplementary material contains six parts. Section \ref{sup:extension} contains interesting extensions to our current framework. Section \ref{sup:mainproof} provides proofs of two main theorems, Section \ref{sup:prooflem} proves main lemmas, Section \ref{sup:aulemma} lists auxiliary lemmas and their proofs, Section \ref{sup:simulation} discusses additional simulation results, and Section \ref{sec: sup_neural} includes the implementation details of a competitive covariate-assisted neural tensor factorization compared in the real data analysis.

 %%%%%%%%%%%%%%%%%%%%%%%%%%%%%%%%%%%
%%%Extensions
%%%%%%%%%%%%%%%%%%%%%%%%%%%%%%%%%%
\subsection{\change{Some Extensions}} 
\label{sup:extension}
%\change{
%In this section we discuss two interesting extensions to our current framework. Section \ref{sup:General} extends to the case where all tensor modes are coupled to covariate matrices and Section \ref{sup:noiseless} considers an interesting extension when we know in advance that the coupled covariate matrix is noiseless. 
%}

\subsubsection{\change{All Tensor Modes are Coupled with Matrices}}
\label{sup:General}
\change{
In Section \ref{sec:meth}, we consider the special case where the tensor and the covariate matrix are coupled along the first mode. In this subsection, we present an extension where all tensor modes are coupled to covariate matrices. Let $\Ten \in \Rbb^{n_1\times n_2\times n_3}$ and $\Mat_{\ac} \in \Rbb^{n_1\times n_{va}}$, $\Mat_{\bc} \in \Rbb^{n_2\times n_{vb}}$, $\Mat_{\cc} \in \Rbb^{n_3\times n_{vc}}$ be the observed third-order tensor and covariate matrices corresponding to the feature information along the three modes of the tensor $\Ten$. The noisy observation model considered in Section \ref{sec:model} becomes
\begin{align*}
  P_{\Omega}(\Ten) = P_{\Omega}(\Ten^{*} + \Er_T); \quad \Mat_a = \Mat_a^{*} + \Er_{Ma}; \quad \Mat_b = \Mat_b^{*} + \Er_{Mb}; \quad \Mat_c = \Mat_c^{*} + \Er_{Mc},
\end{align*}
where $\Er_T$, $ \Er_{Ma}$, $ \Er_{Mb}$  and $ \Er_{Mc}$ are the error tensor and the error matrices respectively; $\Ten ^{*}$, $\Mat_a^{*}$, $\Mat_b^{*}$  and $\Mat_c^{*}$  are the true tensor and the true matrices, which are assumed to have each a low-rank CP decomposition structure \citep{kolda2009tensor} represented as $\Ten^* = \sum_{r\in [R]}\lambda_r^* \ac_r^* \otimes \bc_r^* \otimes \cc_r^*$ and
\begin{align*}
\Mat_{\ac}^* =  \sum_{r\in [R]}\sigma_{ar}^* \ac_r^* \otimes \vc_{ar}^*; \quad \Mat_{\bc}^* =  \sum_{r\in [R]}\sigma_{br}^* \bc_r^* \otimes \vc_{br}^*; \quad \Mat_{\cc}^* =  \sum_{r\in [R]}\sigma_{cr}^* \cc_r^* \otimes \vc_{cr}^* ,
\end{align*}
where $\lambda_r^{*} , \sigma_{ar}^{*} ,  \sigma_{br}^{*},  \sigma_{cr}^{*} \in \mathbb{R}^{+}$, $\ac_r^*\in \mathbb{R}^{n_1}, \bc_r^*\in \mathbb{R}^{n_2} , \cc_r^*\in \mathbb{R}^{n_3}, \vc_{ar}^* \in \mathbb{R}^{n_{va}}, \vc_{br}^* \in \mathbb{R}^{n_{vb}}$ and $\vc_{cr}^* \in \mathbb{R}^{n_{vc}}$ with $\|\ac_r^*\|_2 =\|\bc_r^*\|_2= \|\cc_r^*\|_2 = \|\vc_{ar}^*\|_2= \|\vc_{br}^*\|_2=\|\vc_{cr}^*\|_2=1$ for $r\in [R]$.}

{Given an observed tensor $\Ten$ with missing entries and covariate matrices $\Mat_a$, $\Mat_b$ and $\Mat_c$, in order to recover the true tensor $\Ten^*$ as well as its latent components, the objective function in $(\ref{eq:opt2})$ now becomes
$\| P_{\Omega} \big( \Ten ) -  P_{\Omega} \big( \sum_{r\in [R]}\lambda_r \ac_r \otimes \bc_r \otimes \cc_r \big) \|_F^2 + \| \Mat_a - \sum_{r\in [R]}\sigma_{ar} \ac_r \otimes \vc_{ar}\|_F ^2+  \| \Mat_b - \sum_{r\in [R]} \sigma_{br} \bc_r \otimes \vc_{br}\|_F ^2  +\| \Mat_c - \sum_{r\in [R]} \sigma_{cr} \cc_r \otimes \vc_{cr}\|_F^2$. A similar alternative updating algorithm can be developed to solve this new optimization problem. Figure \ref{ifig:costco} illustrates the rank-one \texttt{COSTCO} procedure when all tensor modes are coupled to covariate matrices. It reveals how \texttt{COSTCO} leverages the additional latent information coming from the covariate matrices on the shared modes. }
\begin{figure}[h!]
   \centering
        \includegraphics[width=0.6\textwidth]{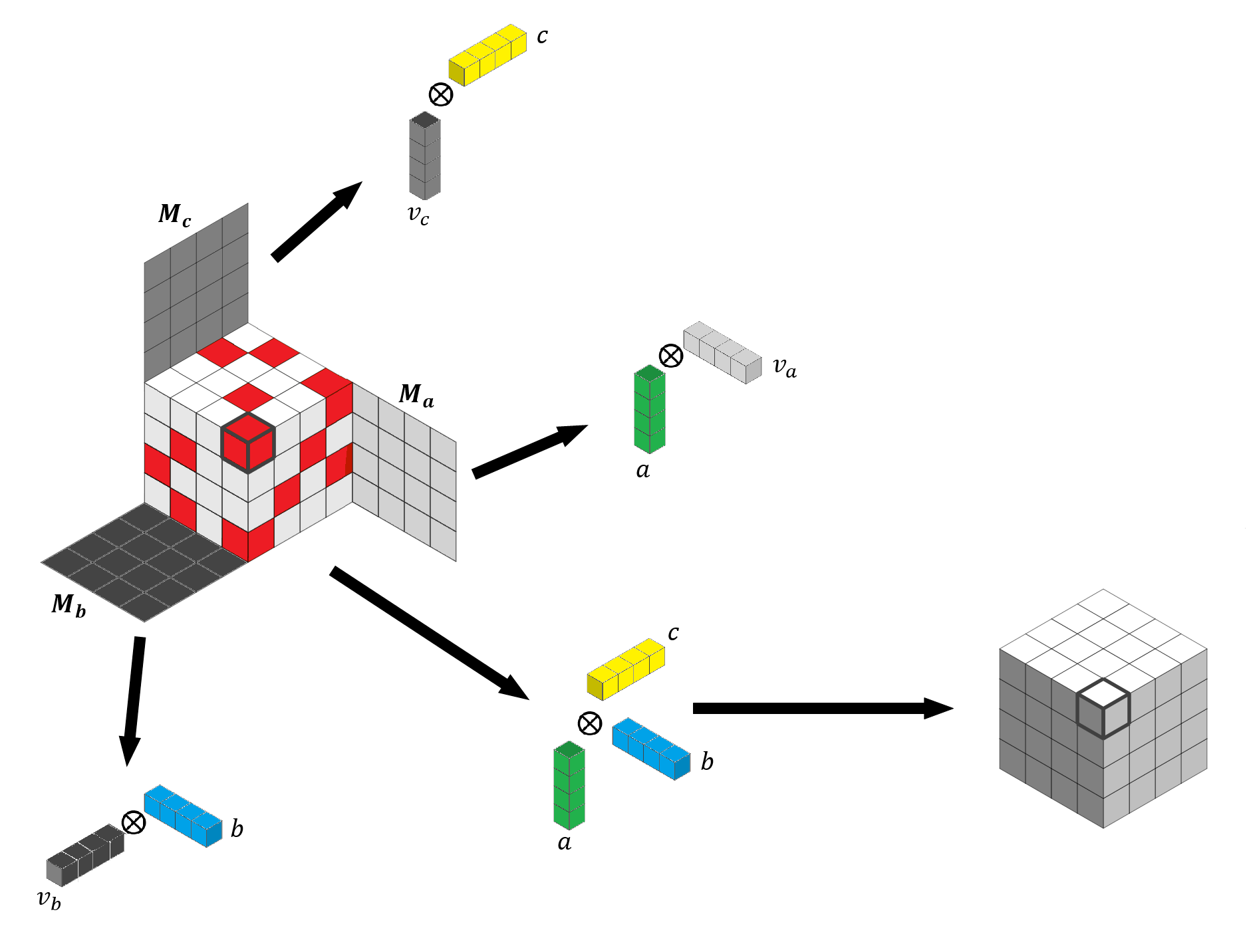}
        \caption{{A rank-one illustration of \texttt{COSTCO} when all the tensor modes are coupled with covariate matrices; red cells represent missing entries. The components ${\ac}$, $\bc$ and $\cc$ are shared by the tensor and matrices $\Mat_a$, $\Mat_b$ and $\Mat_c$, respectively.}}
        \label{ifig:costco}
\end{figure}

{ When all the tensor modes are coupled with covariate matrices, the initialization procedure actually becomes easier. Remind that in Section \ref{sup:initialization}, when there is only one mode of the tensor is coupled with a covariate matrix, we use SVD decomposition of the covariate matrix as the initialization method for the shared tensor components and the robust tensor power method \citep{anandkumar2014tensor} for the non-shared tensor components. When all the tensor modes are coupled with covariate matrices, we can apply SVD decomposition of all these three covariate matrices to obtain the initialization of all latent components directly. }

\red{
In this paper, we consider a coupled tensor and matrix factorization where the shared latent components from the tensor and the corresponding covariate matrix are same. As one reviewer suggests, it is possible to consider a soft generalization of this constraint using a fusion-type penalization. In particular, we can consider a new objective function 
$\| P_{\Omega} \big( \Ten ) -  P_{\Omega} \big( \sum_{r\in [R]}\lambda_r \ac_r \otimes \bc_r \otimes \cc_r \big) \|_F^2 + \| \Mat_a - \sum_{r\in [R]}\sigma_{ar} \ac_r^{'} \otimes \vc_{ar}\|_F ^2+  \| \Mat_b - \sum_{r\in [R]} \sigma_{br} \bc_r^{'} \otimes \vc_{br}\|_F ^2  +\| \Mat_c - \sum_{r\in [R]} \sigma_{cr} \cc_r^{'} \otimes \vc_{cr}\|_F^2 + \lambda_1 \|\ac_r - \ac_r^{'}\|_2 + \lambda_2 \|\bc_r - \bc_r^{'}\|_2+ \lambda_3 \|\cc_r - \cc_r^{'}\|_2$, where $\lambda_1,\lambda_2,\lambda_3$ are some non-negative tuning parameters. When $\lambda_1,\lambda_2,\lambda_3$ are very large, it will eventually lead to our current coupled tensor and matrix factorization framework where $\ac_r = \ac_r^{'}$, $\bc_r = \bc_r^{'}$, $\cc_r = \cc_r^{'}$. On the other hand, when $\lambda_1 = \lambda_2 =\lambda_3 = 0$, no covariate information is incorporated into the tensor completion and the tensor completion is independent of the matrix factorization. Hence, this soft-penalization formulation provides additional flexibility for the amount of information can be borrowed from covariate matrices. We leave a rigorous study on this interesting generalization as future work.}

\subsubsection{\change{Noiseless Covariate Matrices}}
\label{sup:noiseless}
\change{
In this subsection, we discuss an interesting extension when we know in advance that the coupled covariate matrix is noiseless. In this case, improved error rate and sample size condition could be achieved via a small modification to our \texttt{COSTCO} algorithm.
}

\change{Our current \texttt{COSTCO} algorithm is designed to jointly extract latent components from both the tensor and the covariate matrix to learn a synthetic representation. This is achieved via our optimization problem in $(\ref{eq:opt2})$. In order to solve this, we develop an alternative update algorithm which updates one parameter at one time while fixing others. When we know in advance that the coupled covariate matrix is noiseless, i.e., $\Er_M = 0$ and the incoherence parameter $c_0 = 0$, applying SVD on the covariate matrix would lead to the perfect shared components $\ac_r = \ac_r^*$ for $r\in [R]$. In this case, we can fix these shared components $\ac_r = \ac_r^*$ and solve a modified optimization $\min_{\Bm, \Cm, \blambda}  \| P_{\Omega} \big( \Ten ) -  P_{\Omega} \big( \sum \limits_{r \in [R]} \lambda_r \ac_r^* \otimes \bc_r \otimes \cc_r \big) \|_F^2$. In this case, the final error rate of the shared component would be zero, which is much improved over our current rates in Theorems \ref{theorem:global1}-\ref{theorem:global2}. 
}

\change{Moreover, in this case, this modified algorithm could also lead to an improved sample size condition. Based on Assumption 6, the sample size requirement for the non-sparse case ($d=n$) is $n^3p \succeq \frac{\lambda_{max}^{*2}n^{3/2}\log^2(n)}{(\lambda_{min}^*+\sigma_{min}^*)^2}$. When $\Er_M = 0$ and the incoherence parameter $c_0 = 0$, the SVD on the covariate matrix would lead to perfect $\ac_r^*$ for $r\in [R]$. If we fix them in the algorithm, we would need a weaker sample size condition. An extreme case is when all three tensor modes are coupled with a noiseless covariate matrix. Then all the tensor components $\ac^*_r, \bc^*_r, \cc^*_r$ can be perfectly recovered via the SVD operations on three noiseless covariate matrices. Therefore, we can recover the whole tensor without observing any entry in the tensor, i.e., $p=0$. }

However, this modified algorithm would require the knowledge that the covariate matrix is noiseless. As it is challenging to judge whether the coupled covariate matrix is noiseless or not in practice, in this paper we will focus on the current \texttt{COSTCO} algorithm and leave a thorough study of this interesting extension as future work.

\subsubsection{\red{General Sampling Mechanism}}
\label{sup:sampling}
\red{
In this paper we consider a widely used random sampling model where the partially observed entries in the tensor are assumed to be uniformly random sampled from the original tensor. Such random sampling model has been commonly considered in tensor completion \citep{jain2014provable, Barak2016, Song2019, Xia2019, cai2020, zhang2020sparse, xia2021statistically, cai2021nonconvex}. 
}   

\red{
It would be interesting to study how to extend our framework to a general sampling mechanism. \cite{zhang2019cross} considers a special ``cross" sampling mechanism for efficiently compressing a tensor and derives its minimax optimal sample complexity. This approach focuses on how to use minimal samples to provably reconstruct a tensor and is not applicable for other sampling mechanisms. Recently, \cite{yang2021tenips} considers tensor completion with a missing not at random setting where the reveal probabilities of tensor entries are not equal and may depend on the entry values themselves. By assuming that both the true tensor and the reveal probability tensor to be low-rank, they propose a two-step algorithm to first estimate the propensities in the tensor and then predict the missing values of the tensor via a weighted tensor completion procedure. Since considering a different sampling model requires completely new algorithm developments and theoretical analysis tools, in this paper we choose to focus on the common random sampling model and leave a rigorous study of the extension to general sampling mechanism as interesting future work.
}

%%%%%%%%%%%%%%%%%%%%%%%%%%%%%%%%%%%%%%%%%%%
%% Proofs
%%%%%%%%%%%%%%%%%%%%%%%%%%%%%%%%%%%%%%%
\subsection{Proof of Main Theorem} 
\label{sup:mainproof}
In this section we provide the proofs of the main theoretical results presented in \ref{theorem:global1} and \ref{theorem:global2}. As elaborated in the discussion paragraphs in Section \ref{sec:analtheo} proving first the particular case in Theorem \ref{theorem:global1} allows for a better presentation and explanation for the proof technique used for the general case in Theorem \ref{theorem:global2}. For simplicity, in the
following proofs we consider the case where all tensor and matrix modes have the same dimensions $n$ that is $n_1 = n_2 = n_3 = n_v=n$. We also assume that the sparsity parameters for each mode are equal ($d_1 = d_2 =d_3 =d_v =d$). It follows from the two simplification aforementioned that in Algorithm \ref{alg:jALS} we let $s_1 = s_2 = s_3 =s_v=s$. Proving the case, in which the dimensions of the tensor and matrix' modes are allowed to be unequal is a trivial yet notation heavy extension of the technique we use in the proof of Theorem \ref{theorem:global1} and Theorem \ref{theorem:global2}. As defined in equation \eqref{supeq:distance}, we use the euclidean distance between the component estimates and true components to measure the error for component recovery. We also use the relative absolute difference between estimated and true weights to capture the recovery error for the weights as defined in equation \eqref{supeq:relatdistance}. Define $\dc_{u_r}$ to be,
\begin{equation}
\label{supeq:distance}
\dc_{u_r}=: \uc_r-\uc^*_r, \quad \text{and} \quad \|\dc_{u_r}\|_2= \|\uc_r-\uc_r^*\|_2, 
\end{equation}
and 
\begin{equation}
\label{supeq:relatdistance}
\Delta_{\lambda_r}:=|\frac{\lambda_r-\lambda_{r}^{*}}{\lambda_{r}^{*}}| \quad
\text{ and } \quad  \Delta_{\sigma_r}:=|\frac{\sigma_r-\sigma_{r}^{*}}{\sigma_{r}^{*}}|,
\end{equation}
where $\uc_r$ could be any of $\ac_r, \bc_r, \cc_r, \vc_r$, $\forall r \in [R]$. 
 \subsubsection{Proof of Theorem \ref{theorem:global1}}
 \label{proof:global1}
 
 Theorem \ref{theorem:global1} provides the sufficient conditions which guarantee that the shared tensor components $\ac_r$ and non-shared components $\bc_r$,$\cc_r$ recovered in Algorithm \ref{alg:jALS} converge to the truth $\ac_r^*$ and $\bc_r^*$, $\cc_r^*$ respectively with the assumption that the tensor and matrix are dense and their decomposition weights are equal in each mode i.e $\lambda_r^*=\sigma_r^* \quad \forall r\in [R]$. The theorem also provides the explicit convergence rates for the tensor components in Algorithm \ref{alg:jALS} and highlights the difference in rates between the shared and non-shared components. 
 
 Our proof consists of three steps. In Step 1 we use Lemma \ref{lem:update} to derive the close form for the optimization problem presented in equation \eqref{eq:opt2}. This step is only specific to the dense tensor and equal weights case as it makes it possible to derive a close form solution to the optimization formula presented in equation \eqref{eq:opt2}. In Step 2, we derive a general bound for the share and non-shared tensor estimates by proving Lemmas \ref{cor:contrac} and \ref{cor:contraa} given that the components obtained from the initialization method satisfy a specific error constraint. In Step 3, we simplify the error bound obtained in Lemma \ref{cor:contrac} and \ref{cor:contraa} to ensure that the share and non-shared tensor component estimate contract at a geometric rate in one iteration. Theorem \ref{theorem:global1} is then completed by showing that after enough iterations the contraction error vanishes to only leave a statistical error.
 
\textbf{Step 1:} The next lemma accomplishes the first step in proving Theorem  \ref{theorem:global1}. Since the tensor and matrix weights are assumed to be equal, without loss of generality we use $\lambda_r^*$ and $\lambda_r$ $\forall r \in [R]$ to represent true and estimated weights respectively for both tensor and matrix.
\begin{lemma}
\label{lem:update}
Let $\text{res}_M = \Mat - \sum \limits_{m \neq r} \lambda_m \ac_m \otimes \vc_m$ and $\text{res}_T = P_\Omega(\Ten) - P_\Omega( \sum \limits_{m \neq r} \lambda_m \ac_m  \otimes \bc_m  \otimes \cc_m)$ be the residual matrix and residual tensor, respectively defined on line (7) of Algorithm \ref{alg:jALS}. In each ALS update of Algorithm \ref{alg:jALS}, the solution to the optimization problem in equation \eqref{eq:opt2} for the shared and non-shared components of the tensor and matrix in the $r^{th}$ iteration of the inner loop are,
\begin{align}
\label{subeq:updatea}
&\textbf{Share Components: } \ac_r  = \Tfrac{\lambda_r\text{res}_T(\mathbf{I}, \bc_r, \cc_r) + \sigma_r \text{res}_M \vc_r}{\lambda_r ^2 P_\Omega(\mathbf{I}, {\bc}_r^2, {\cc_r}^2 ) + \sigma_r^2  },\\
\label{subeq:updatebc}
&\textbf{Tensor non-shared components: }	\bc_r = \tilde{\bc}_r/\|\tilde{\bc}_r\|_2, \quad 	\cc_r = \tilde{\cc}_r/\|\tilde{\cc}_r\|_2, \quad 	\lambda_r = \|\tilde{\cc}_r\|_2,\\
\label{subeq:updatev}
&\textbf{Matrix non-shared components: } \vc_r = \tilde{\vc}_r/\|\tilde{\vc}_r\|_2 \quad and \quad \sigma_r = \|\tilde{\vc}_r\|_2,
\end{align}
where $\tilde{\bc}_r$, $\tilde{\cc}_r$, $\tilde{\vc}_r$ have the following form
\begin{equation}
\label{eq:unorm_update}
	\tilde{\bc}_r = \Tfrac{\text{res}_T (\ac_r,\mathbf{I}, \cc_r)}{ P_\Omega( \ac_r^2,\mathbf{I},\cc_r^2 )} \quad 	\tilde{\cc}_r = \Tfrac{\text{res}_T(\ac_r, \bc_r,\mathbf{I})}{ P_\Omega( \ac_r^2, \bc_r^2,\mathbf{I})}\text{ and } \tilde{\vc}_r =\text{res}_M^{\top}  \ac_r.
\end{equation}
\end{lemma}
Note that the horizontal double lines in the expressions above indicate element-wise fraction and the squares in the denominator represent the element-wise squaring. The proof of Lemma \ref{lem:update} is provided in Section \ref{sup:prooflem}. It involves deriving the close form of the optimization problem presented in equation \eqref{eq:opt2} in the non-sparse tensor case.

\textbf{Step 2:} The second step builds the error contraction results in one iteration of Algorithm 
\ref{alg:jALS}. We achieve step two through Lemmas \ref{cor:contrac} and \ref{cor:contraa} which address the non-shared and shared component cases respectively.
%%%%%%%%%%%%%%%%%%%%%%%%%%%%%%%%%%%%%%%%%%%%%%%%%%%%%%%%%%%%%%%%%%%%
%%%%%%%%%%%%%%%%%%%%%%%%%%%%%%%%%%%%%%%%%%%%%%%%%%%%%%%%%%%%%%%%%%%%
\begin{lemma} 
\label{cor:contrac}
Assume Assumption 1 holds and $p \geq  \frac{C\mu ^3(1+\gamma/3) \log_2(n^{10}) }{n^{3/2}\gamma^2}$ for some positive $\gamma$. Also assume estimates $\ac_r$, $\bc_r$, $\lambda_r$ of our algorithm with $s_i =n_i$, $i= 1,2,3,v$, satisfy $\max\{\|\dc_{a_r}\|, \|\dc_{b_r}\|,\Delta_{\lambda_r}\}\leq \epsilon_T $  $\forall r \in [R]$ with $\dc_{a_r}, \dc_{b_r}, \Delta_{\lambda_r}$ defined in $(\ref{supeq:distance})$. Then, the update for the non-shared tensor component $\cc_r$ satisfies with probability  $1-2n^{-9}$,
\begin{equation}
 \label{supeq: contractc}
     \max\limits_{r\in[R]}\|\cc_r-\cc_{r}^{*}\|_2 \leq   \frac{16pR \lambda_{max}^{*}  \max\left( c_0/\sqrt{n}+3\epsilon_T,\gamma  \right)\epsilon_T+\sqrt{p}(1+\gamma)\|\mathcal{E}_T \|}{\lambda_{min}^{*} p(1-\gamma)}.
\end{equation}
%with $\gamma \leq C /\log(n)$ where $C$  is a constant.
\end{lemma}
The detailed proof of Lemma \ref{cor:contrac} is presented in Section \ref{sup:prooflem}. We later show in step 3 of the proof of Theorem \ref{theorem:global1} that the upper bound in \eqref{supeq: contractc} can we written as the sum of a contracting term and a non contracting statistical error term.
\begin{lemma} 
\label{cor:contraa}
Assume Assumption 1 holds and $p \geq  \frac{C\mu ^3(1+\gamma/3) \log_2(n^{10}) }{n^{3/2}\gamma^2}$ for some positive $\gamma$. In addition, assume estimators $\cc_r$, $\bc_r$, $\vc_r$, $\lambda_r$, $\sigma_r$ of our algorithm with $s_i =n_i$, $i= 1,2,3,v$, satisfy $\max\{\|\dc_{c_r}\|, \|\dc_{b_r}\|,  \Delta_{\lambda_r}\}\leq \epsilon_T $ and $ \{\|\dc_{v_r}\|,\Delta_{\sigma_r}\}\leq \epsilon_M $ $\forall r \in [R]$. Then the update for the shared tensor component $\ac_r$ satisfies with probability $1-2n^{-9}$,
\begin{equation}
\label{supeq:aa}
    \max\limits_{r\in[R]}\|\ac_r-\ac_{r}^{*}\|_2 \leq  g(p,\epsilon_T, \zeta,R) \epsilon_T  + f(\epsilon_M, \zeta,R)\epsilon_M + \frac{1}{\lambda_{min}^{*} } \frac{\sqrt{p}(1+\gamma)\|\mathcal{E}_T \| +\|\mathcal{E}_M \|}{p(1-\gamma)+1}
\end{equation}  
 with, 
 \begin{align*}
      g(p,\epsilon_T, \zeta,R):= \frac{16pR \lambda_{max}^{*} \left(  \zeta+3\epsilon_T, \gamma  \right)}{\lambda_{min}^{*}(p(1-\gamma)+1)}; \quad 
f(\epsilon_M, \zeta,R):=\frac{6R \lambda_{max}^{*}( \zeta+3\epsilon_M ) \epsilon_M}{\lambda_{min}^{*}(p(1-\gamma)+1)}, \quad  \text{and} \quad  \zeta=c_0/\sqrt{n}.
 \end{align*}
\end{lemma}
The proof of Lemma \ref{cor:contrac} and Lemma \ref{cor:contraa} show that each iteration of Algorithm \ref{alg:jALS} results in an error contraction for the estimates of the non-shared ($\bc_r$ and $\cc_r$) and shared ($\ac_r$) tensor components respectively. Such results imply that after a sufficient number of iterations, Algorithm \ref{alg:jALS} can yield good estimates for these components.
The detailed proof of Lemma \ref{cor:contraa} is discussed in Section \ref{sup:prooflem}.\\
\textbf{Step 3:} To complete the proof of the theorem, we carefully employ the assumptions on the initialization in order to guarantee that expressions \eqref{supeq: contractc} and \eqref{supeq:aa} in Lemmas \ref{cor:contrac} and \ref{cor:contraa} can be written in the form $\epsilon_R + q \epsilon_0$ with $q \leq \frac{1}{2}$. This entails showing that for $f(\epsilon_M, \zeta,R)$ and $g(\epsilon_M, \zeta,R)$ in the Lemma \ref{cor:contraa} adds up to less than $\frac{1}{2}$ given the assumptions in Theorem \ref{theorem:global1}.\\ Denote $\epsilon_0 := \max\{\epsilon_{T_{0}},\epsilon_{M_{0}}\}$, set $\gamma:= \frac{\lambda_{min}^*}{64R\lambda_{max}^*}$ and define $q_1$ and $q_2$
\begin{align*}
 q_1:=\frac{16R \lambda_{max}^{*} (\zeta+3\epsilon_0)(p+\frac{6}{16})}{\lambda_{min}^{*}(p(1-\gamma)+1)} \quad  \text{and} \quad  q_2:=\frac{16R \lambda_{max}^{*} (p\gamma +\frac{6}{16}(\zeta+3\epsilon_0))}{\lambda_{min}^{*}(p(1-\gamma)+1)}.
\end{align*}
According to Assumption 3, we get that $q_1\leq \frac{p+6/16}{2p+2}\leq \frac{1}{2}$. Also $q_2\leq \frac{p}{4(\frac{63}{64}p+1)}+ \frac{3}{16} \leq \frac{1}{4}+\frac{3}{16}<\frac{1}{2}$ since $p\leq 1$. This implies that $q:= \max\{q_1,q_2\}\leq 1/2$.\\ Finally, we bound the error term of $  \max\limits_{r\in[R]}\|\ac_r-\ac_{r}^{*}\|_2 $ by showing that it can be written as a sum
of a contracting term and a constant non-contracting term. Specifically, according to \eqref{supeq:aa} in each iteration we have,
\begin{align*}
    \max\limits_{r\in[R]}\|\ac_r-\ac_{r}^{*}\|_2 &\leq  g(p,\epsilon_{T_0}, \zeta,R) \epsilon_{T_0}  + f(\epsilon_{M_0}, \zeta,R)\epsilon_{M_0} + \frac{1}{\lambda_{min}^{*} } \frac{\sqrt{p}(1+\gamma)\|\mathcal{E}_T \| +\|\mathcal{E}_M \|}{p(1-\gamma)+1}\\
    & \leq \max\{q_1,q_2\}\epsilon_0 + \frac{1}{\lambda_{min}^{*} } \frac{\sqrt{p}(65/64)\|\mathcal{E}_T \| +\|\mathcal{E}_M \|}{p(63/64)+1}\\
    &\leq q \epsilon_0 + \frac{1}{\lambda_{min}^{*} } \frac{\sqrt{p}(65/64)\|\mathcal{E}_T \| +\|\mathcal{E}_M \|}{p(63/64)+1},
\stepcounter{equation}\tag{\theequation}\label{eq:contraction}
\end{align*} 
where $q\epsilon_0$ is a contracting term and the term after it is non contracting. According to the signal-to-noise condition in Assumption 4, we have the non-contracting term satisfies $\frac{1}{\lambda_{min}^{*} } \frac{\sqrt{p}(65/64)\|\mathcal{E}_T \| +\|\mathcal{E}_M \|}{p(63/64)+1} = o(1)$. This together with $q\le 1/2$ and the bounded initialization condition implies that the estimation error after one-iteration in $(\ref{eq:contraction})$ is still bounded by $\epsilon_0$. By iteratively applying the above inequality, after $\tau=\Omega\left(\log_{2}{\left(\frac{(p+1)\epsilon_{0} }{\sqrt{p}\|\mathcal{E}_T\| +\|\mathcal{E}_M \|}\right)}\right)$, we get
\[\max\limits_{r\in[R]}\|\ac_r-\ac_{r}^{*}\|_2\leq \mathcal{O}_p\left(\frac{1}{\lambda_{min}^{*} } \frac{\sqrt{p}\|\mathcal{E}_T \| +\|\mathcal{E}_M \|}{p+1}\right).\]
Similar derivation can be applied on the upper bound of  $\max\limits_{r\in[R]}\|\cc_r-\cc_{r}^{*}\|_2$ in \eqref{supeq: contractc} to get a contracting and non contracting term. Then taking the maximun over all non-shared components and tensor weights lead to getting after running  $\tau= \Omega\left(\log_{2}{\left(\frac{\sqrt{p}\lambda_{min}^{*}\epsilon_{0}}{\|\mathcal{E}_T \|}\right)}\right)$ iterations of Algorithm \ref{alg:jALS},
  \[\max_{r\in[R]} \left( \|\bc_r-\bc_{r}^{*} \|_2,\|\cc_r-\cc_{r}^{*} \|_2 ,  \frac{|\lambda_r -\lambda_{r}^{*}|}{  \lambda_r^{*}}\right) \leq \mathcal{O}_p\left(\frac{\|\mathcal{E}_T \|}{\sqrt{p}\lambda_{min}^{*}}\right),\] 
  which completes the proof of Theorem \ref{theorem:global1}.\eop

%%%%%%%%%%%%%%%%%%%%%%%%%%%%%%%%%%%%%%%%%%%%%%%%%%%%
\subsubsection{Proof of Theorem \ref{theorem:global2}}
 \label{proof:global2}
 In this section we establish the results for the analysis of Theorem \ref{theorem:global2} which is the general and sparse case where the matrix and tensor weights are not assumed to be equal. In order to prove the general case we make use of some of the intermediate results derived in the analysis of Theorem \ref{theorem:global1}. Namely, we follow the 3 three steps analysis approach introduced in the analysis of Theorem \ref{theorem:global1} and highlight the key difference which makes the analysis of Theorem \ref{theorem:global2} non trivial in comparison. As presented in the formulation of the optimization problem in \eqref{eq:opt2} we use the $\ell^0$ norm regularization as a mean to  introduce sparsity in the model. However, deriving a close form solution to this sparse optimization problem becomes very difficult with this choice of regularization function. In step 1 of the analysis, we circumvent this issue by using a greedy truncation method defined on lines (9) and (11) of Algorithm \ref{alg:jALS} to approximate the sparse solution to the optimization problem in \eqref{eq:opt2}. We show that using the truncation method to only preserve the $s$ largest entries of the components with the condition that $s\geq d$ is suitable for accurate components recovery. In practice for Algorithm \ref{alg:jALS} the parameter $s$ can be tuned in a data-driven manner following the sequential tuning schema presented in Algorithm \ref{sup:tuning}. In step 2 of the analysis, we derive a general bound for the shared tensor component through Lemma \ref{cor:contraas}. In step 3 we simplify the general bound derived in step 2 to show that one iteration of the algorithm results in a geometric error contraction. Theorem \ref{theorem:global2} is then completed by showing that after enough iterations the contraction error vanished to only leave a statistical error.
 
\begin{lemma} 
\label{cor:contraas}
Assume Assumptions 5, 6 and 7 hold. In addition, assume estimators $\bc_r$, $\cc_r$, $\vc_r$, $\lambda_r$, $\sigma_r$ of our algorithm satisfy $\max\{\|\dc_{c_r}\|, \|\dc_{b_r}\|,\Delta_{\lambda_r}\}\leq \epsilon_T $ and $ \{\|\dc_{v_r}\|, \Delta_{\sigma_r}\}\leq \epsilon_M $ $\forall r \in [R]$  and $s_i \geq d_i$ for $i= 1,2,3,v$. Then the update for the shared tensor component $\ac_r$ satisfies with probability $1-2n^{-9}$,
\begin{align*}
%\label{supeq:ba}
   \max\limits_{r\in[R]}\|\ac_r-\ac_{r}^{*}\|_2 & \leq g(p,\epsilon_T, \zeta,R) \epsilon_T + f(\epsilon_M, \zeta,R)\epsilon_M\\ &+\frac{\lambda_{max}^{*} \sqrt{p}(1+\gamma)\| \Er_T \|_{<d+s>} +\sigma_{max}^{*}\| \Er_M \|_{<d+s>}}{\lambda_{min}^{*2} p(1-\gamma)+\sigma_{min}^{*2}},
   \stepcounter{equation}\tag{\theequation}\label{supeq:ba}
\end{align*}  
where $ \zeta=c_0/\sqrt{d}$ and 
\begin{align*}
 g(p,\epsilon_T, \zeta,R)\leq \frac{24pR  \lambda_{max}^{*2}  \max(  \zeta+3\epsilon_T, \gamma)  }{\lambda_{min}^{*2} p(1-\gamma)+\sigma_{min}^{*2}}; \quad  f(\epsilon_M, \zeta,R)\leq\frac{9R \sigma_{max}^{*2} (\zeta+3\epsilon_M)}{\lambda_{min}^{*2} p(1-\gamma)+\sigma_{min}^{*2}}.
\end{align*}
\end{lemma}
The detailed proof of Lemma \ref{cor:contraas} is discussed in Section \ref{sup:prooflem}.\\
\textbf{Step 3:} The last step in the proof of Theorem \ref{theorem:global2}, consists in using the assumptions on the initialization error in order to guarantee that expression \eqref{supeq:ba} in Lemmas \ref{cor:contraas} can be written in the form $\epsilon_R + q \epsilon_0$ with $q \leq \frac{1}{2}$. Just like was the case in the proof of Theorem \ref{theorem:global1}, this entails showing that for $f(\epsilon_M, \zeta,R)$ and $g(\epsilon_M, \zeta,R)$ adds up to less than $\frac{1}{2}$ given the assumptions in Theorem \ref{theorem:global2}.

Given the initialization condition in Assumption 7 we get
$$g(p,\epsilon_T, \zeta,R)\leq \frac{24pR  \lambda_{max}^{*2}  \max(  \zeta+3\epsilon_T, \gamma)  }{\lambda_{min}^{*2} p(1-\gamma)+\sigma_{min}^{*2}}; \quad
f(\epsilon_M, \zeta,R)\leq\frac{9R \sigma_{max}^{*2} (\zeta+3\epsilon_M)}{\lambda_{min}^{*2} p(1-\gamma)+\sigma_{min}^{*2}}$$
Denote  $\epsilon_0 := \max\{\epsilon_{T_{0}},\epsilon_{M_{0}}\}$,  $q_1:=\frac{24R (\zeta+3\epsilon_0)(\lambda_{max}^{*2}p+\frac{9}{24}\sigma_{max}^{*2})}{\lambda_{min}^{*2} p(1-\gamma)+\sigma_{min}^{*2}}$ and
$q_2:=\frac{24R ( \lambda_{max}^{*2}
p\gamma + \frac{9}{24}\sigma_{max}^{*2}
(\zeta+3\epsilon_0))}{\lambda_{min}^{*2} p(1-\gamma)+\sigma_{min}^{*2}}$. 
We choose $\gamma = \frac{1/2\lambda_{min}+1/2 \sigma_{min}}{96R\lambda_{max}}$. According to Assumption 7 we get that $q_1\leq \frac{p\lambda_{max}^{*2}+ 3/8\sigma_{max}^{*2}}{2(p\lambda_{max}^{*2}+ \sigma_{max}^{*2})}\leq \frac{1}{2}$. Also $q_2\leq \frac{p\min\{\lambda_{min}^{*2},\sigma_{min}^{*2}\}}{4(\lambda_{min}^{*2}p\frac{95}{96}+\sigma_{min}^{*2})}+ \frac{3 \sigma_{max}^{*2}}{16(p\lambda_{max}^{*2}+\sigma_{max}^{*2})} \leq\frac{p}{4(p\frac{95}{96}+1)}+\frac{3}{16}$. Hence $q_2\leq \frac{1}{4}+\frac{3}{16}<\frac{1}{2}$ since $p\leq 1$. This implies that $q:= \max\{q_1,q_2\}\leq 1/2$.\\ Finally, we bound the error term of $  \max\limits_{r\in[R]}\|\ac_r-\ac_{r}^{*}\|_2 $ by showing that it can be written as a sum
of a contracting term and a constant non-contracting term. Specifically, according to \eqref{supeq:aa} in each iteration we have,
\begin{align*}
    \max\limits_{r\in[R]}\|\ac_r-\ac_{r}^{*}\|_2 &\leq  g(p,\epsilon_{T_0}, \zeta,R) \epsilon_{T_0}  + f(\epsilon_{M_0}, \zeta,R)\epsilon_{M_0}\\  &+\frac{(\lambda_{max}^{*}+\epsilon_T) \sqrt{p}(1+\gamma)\| \Er_T \|_{<d+s>} +(\sigma_{max}^{*}+\epsilon_T)\|\| \Er_M \|_{<d+s>}}{(\lambda_{min}^{*}+\epsilon_T)^2 p(1-\gamma)+(\sigma_{min}^{*}+\epsilon_M)^2}\\
    & \leq \max\{q_1,q_2\}\epsilon_0 + \frac{(97/96)\sqrt{p}\lambda_{max}^{*}\|\mathcal{E}_T \|_{<d+s>} +\sigma_{max}^{*}\|\mathcal{E}_M \|_{<d+s>}}{\frac{95}{96}p\lambda_{min}^{*2}+\sigma_{min}^{*2}}\\
    &\leq q \epsilon_0 + \frac{(97/96)p\lambda_{max}^{*}\|\mathcal{E}_T \|_{<d+s>} +\sigma_{max}^{*}\|\mathcal{E}_M \|_{<d+s>}}{\frac{95}{96}\sqrt{p}\lambda_{min}^{*2}+\sigma_{min}^{*2}},
\stepcounter{equation}\tag{\theequation}\label{eq:contraction2}
\end{align*} 
where $q\epsilon_0$ is a contracting term. According to Assumption 8 and the facts that $\|\mathcal{E}_T \|_{<d+s>} \le \|\mathcal{E}_T \|_{<2s>} = \mathcal{O}(\|\mathcal{E}_T \|_{<s>})$ and $\|\mathcal{E}_M \|_{<d+s>} \le \|\mathcal{E}_M \|_{<2s>} = \mathcal{O}(\|\mathcal{E}_M \|_{<s>})$, the non-contracting term converges to zero. Therefore, the error in $(\ref{eq:contraction2})$ is still bounded by $\epsilon_0$. By iteratively applying the above inequality, after the number of iterations stated in Theorem \ref{theorem:global2}, we get \[\max\limits_{r\in[R]}\|\ac_r-\ac_{r}^{*}\|_2 \leq \mathcal{O}_p\left(\frac{\sqrt{p}\lambda_{max}^{*}\|\mathcal{E}_T \|_{<s>} +\sigma_{max}^{*}\|\mathcal{E}_M \|_{<s>}}{p\lambda_{min}^{*2}+\sigma_{min}^{*2}}\right),\]
The proof for the non-shared component in Theorem \ref{theorem:global2} is very similar to that of the non-share component in Theorem \ref{theorem:global1} we therefore leave it out.
This completes the proof of Theorem \ref{theorem:global2}.\eop

 %%%%%%%%%%%%%%%%%%%%%%%%%%%%%%%%%%%%%%%%%%%%%%%%%%%%%%%%
\subsection{Proofs of Lemmas \ref{lem:update}, \ref{cor:contrac}, \ref{cor:contraa} and \ref{cor:contraas}} 
\label{sup:prooflem}
In this section we provide details of the derivation for the proofs of Lemmas \ref{lem:update}-\ref{cor:contraas}.
\subsubsection{Proof of Lemma \ref{lem:update}}
The dense version of the optimization problem in \eqref{eq:opt2} can be formulated as follows:
\paragraph{Optimization:} Non-Sparse formulation
\begin{equation}
 \min_{\Am, \Bm, \Cm, \Vm, \blambda, \bsigma} \Big \{ \| P_{\Omega} \big( \Ten ) -  P_{\Omega} \big( \sum \limits_{r \in [R]} \lambda_r \ac_r \otimes \bc_r \otimes \cc_r \big) \|_{F}^{2}  +  \| \Mat - \sum \limits_{r \in [R]} \sigma_r \ac_r \otimes \vc_r \|_{F}^{2} \Big \}.\stepcounter{equation}\tag{\theequation} \label{eq:opt1}\\
%& \text{subject to }\sigma_r, \lambda_r \in \Rbb^{+},
\end{equation}
Denote $\text{res}_M = \Mat - \sum \limits_{m \neq r} \sigma_m \ac_m \otimes \vc_m$ and $\text{res}_T = P_\Omega(\Ten) - P_\Omega( \sum \limits_{m \neq r} \lambda_m \ac_m  \otimes \bc_m  \otimes \cc_m)$ as the residual matrix and residual tensor, respectively. In each ALS update of Algorithm \ref{alg:jALS} we need to solve the following least squares optimizations problem.
\begin{equation}
\label{eq:optSol}
	\min_{\ac_r}  \Big \{ \| \text{res}_M - \sigma_r \ac_r \otimes \vc_r \|_F^2   
	 + \| \text{res}_T -  P_{\Omega} ( \lambda_r \ac_r \otimes \bc_r \otimes \cc_r ) \|_F^2 \Big \}.
\end{equation}
The optimization problem in \eqref{eq:optSol} is convex in $\ac_r$. Therefore, we can find $\ac_r$ by taking its derivative and setting it to zero. In order to do this we first derive the equivalent of the optimization function in \eqref{eq:optSol} explicitly in terms of the entries of the tensor and matrix components:
\begin{equation}
\label{eq:optFro}
	 \min_{\ac_r} \Big \{ \sum \limits_{i,j} \left ( {\text{res}_M}_{i,l} - \sigma_r \ac_r (i)\times \vc_r(l) \right)^2 
	+ \sum \limits_{\{i,j,k\} \in \Omega} \left( {\text{res}_T}_{i,j,k} -   \lambda_r \ac_r(i) \times \bc_r(j) \times \cc_r(k) \right)^2 \Big \},
\end{equation}
where ${\text{res}_T}_{i,j,k} $ is the $(i,j,k)^{\text{th}}$ entry of $\text{res}_T$ and ${\text{res}_M}_{i,l}$ is the $(i,l)^{\text{th}}$ entry of $\text{res}_M$. The notation $\{i,j,k\} \in \Omega$ with $\Omega$ defines in \eqref{eq:project}, guarantees that the summation only applies on the observed entries of tensor $\text{res}_T$; $\ac_r(i)$ is the $i^{\text{th}}$ component of $\ac_r$ where $i\in [n]$.

Taking the derivative of \eqref{eq:optFro} with respect to $\ac_r(i)$ for all $i \in [n]$ and setting it to zero we get:
\begin{equation}
\label{eq:eq5}
	\ac_r(i) = \frac{ \lambda_r  \sum \limits_{j,k} ( {\text{res}_T}_{i,j,k} \bc_r(j) \cc_r(k) ) +  \sigma_r \sum \limits_j  {\text{res}_M}_{i,l} \vc_r(l)}{ \lambda_r^2 \sum \limits_{j,k} {\bc}_r^2(j) {\cc}_r^2(k) +  \sigma_r^2 \sum \limits_{l}  {\vc}_r^2(l)}
\end{equation}
for all $i \in [n]$. The first summation in the numerator of equation \eqref{eq:eq5} is the definition of the modes $2$ and $3$ tensor matrix product of $\text{res}_T$ with the matrix obtained from  $\bc_{r} \otimes \cc_{r}$. Following the notation provided in Section \ref{sec:prelim} this product can be rewritten as:
\begin{equation}
\label{eq:prod}
	 \text{res}_T(\mathbf{I}, \bc_r, \cc_r) = {\text{res}_T} \times_2 \bc_r \times_3 \cc_r ,
\end{equation}
for all $i \in [n]$, where $\mathbf{I}$ is the identity matrix. It is worth noting that the vector tensor product in \eqref{eq:prod} is a vector of length $n$. We can write the second term in the numerator as a matrix vector left multiplication. The vector $\ac_r$ can therefore be written as:
\begin{equation}
\label{eq:update}
	\ac_r = \Tfrac{\lambda_r\text{res}_T(\mathbf{I}, \bc_r, \cc_r) + \sigma_r \text{res}_M \vc_r}{\lambda_r ^2 P_\Omega(\mathbf{I}, {\bc}_r^2, {\cc}_r^2 ) + \sigma_r^2  },
\end{equation}
where the double line fraction indicates element-wise division and $(\cdot)^2$ denotes elements-wise power. %The expression in \eqref{eq:update} is the update formula given in Line 12 of Algorithm \ref{alg:jALS}.\\

In order to solve the optimization problem for components other than the first component that are not shared with the matrix we proceed similarly. We start from:
\begin{align}
\label{eq:optSol2}
	& \min_{\bc_r}  \Big \{ \| \text{res}_T -  P_{\Omega} ( \lambda_r \ac_r \otimes \bc_r) \|_F^2 \Big \},
\end{align}
which is equivalent to 
\begin{align}
\label{eq:optFro2}
	& \sum \limits_{\{i,j,k\} \in \Omega} \left( {\text{res}_T}_{i,j,k} - \lambda_r \ac_r(i) \times \bc_r(j) \times \cc_r(k) \right)^2.
\end{align}
Taking the derivative of \eqref{eq:optFro2} with respect to $\bc_r(j)$ or $\cc_r(k)$ then setting to them to zero and solving for $\bc_r(j)$ or $\cc_r(k)$ we get the following update:
\begin{equation}
\label{eq:aSol23}
  \tilde{\bc}_r(j) :=\lambda_r \bc_r(j) = \frac{\sum \limits_{\{i,.,k\} \in \Omega} ( {\text{res}_T}_{i,j,k} \ac_r(i) \cc_r(k) )}{\sum \limits_{\{i,.,k\} \in \Omega} \ac_r^2(i) \cc_r^2(k)},\quad  \tilde{\cc}_r(k) :=\lambda_r \cc_r(k) = \frac{\sum \limits_{\{i,j,.\} \in \Omega} ( {\text{res}_T}_{i,j,k} \ac_r(i) \bc_r(j) )}{\sum \limits_{\{i,j,.\} \in \Omega} \ac_r^2(i) \bc_r^2(j)},
\end{equation}
respectively. In vector form this is written as,
\begin{equation}
	\tilde{\bc}_r = \Tfrac{\text{res}_T (\ac_r,\mathbf{I}, \cc_r)}{ P_\Omega( \ac_r^2,\mathbf{I},\cc_r^2 )} \quad \text{ and } \quad 	\tilde{\cc}_r = \Tfrac{\text{res}_T(\ac_r, \bc_r,\mathbf{I})}{ P_\Omega( \ac_r^2, \bc_r^2,\mathbf{I})}.
\end{equation}
These are the un-normalized updates in line 10 of Algorithm \ref{alg:jALS}.
Since by definition $\bc_r$ and $\cc_r$ are unit vectors then $\|\tilde{\cc}_r\|_2=\|\lambda_r \cc_r\|_2=\lambda_r$ as defined in line 12 of Algorithm \ref{alg:jALS} and
 $\cc_r= \tilde{\cc}_r/ \|\tilde{\cc}_r\|_2$ as in line 13 of the main algorithm. The update for $\bc$ is obtained in a similar manner.
The above derivation corresponds to the non-sparse scenario, i.e., Algorithm 1 without the truncation steps on lines 9 and 11. However for the sparse case, to incorporate sparsity in the resulting update equations, we use the truncation scheme proposed in \citet{sun2017}.
We get the estimate of the matrix component $\vc_r$, using a similar derivation and get,
\begin{equation}
\label{tab:crossVal4}
	\tilde{\vc}_r := \sigma_r\vc_r = \text{res}_M^{\top}  \ac_r,
\end{equation}
and since $\vc_r$ is a unit vector we get $\sigma_r=\|\tilde{\vc}_r\|_2$ and $\vc_r= \tilde{\vc}_r / \|\tilde{\vc}_r\|_2$ as in lines 12 and 13 of Algorithm \ref{alg:jALS}. This complete the proof of Lemma \ref{lem:update}.\eop

\subsubsection{Proof of Lemma \ref{cor:contrac}}
\label{proof:contrac}
 The main challenge in the proof of Lemma \ref{cor:contrac} lies in finding a tight upper bound for the error of $c_r$. In the following derivation only provide the analysis for the non-shared tensor components $\cc_r$ since the proof of the other non-shared component $\bc_r$ is very similar.\\
 In \eqref{subeq:updatebc} we derived the close form formula for the update $\cc_r$ to be $ \tilde{\cc}_r/\|\tilde{\cc}_r\|_2$. To bound the expression $\|\cc_r -\cc_r^*\|_2$ , we make use of the intermediate estimate $\tilde{\cc}_r$ which is define in $(\ref{eq:unorm_update})$ as,
\begin{equation}
\label{subeq:updatebc2}
\tilde{\cc}_r = \Tfrac{\text{res}_T(\ac_r, \bc_r, \mathbf{I})}{ P_\Omega( {\ac}_r^2, {\bc}_r^2 ,\mathbf{I})}.
\end{equation}
From Lemma \ref{lem:update}, notice that $\tilde{\cc}_r$ can be written as $\lambda_r \cc_r$. That is, $\tilde{\cc}_r$ can be thought of as the un-normalized version of the estimate $\cc_r$. Proving Lemma \ref{cor:contrac} therefore consists in deriving an error bound for $\|\tilde{\cc}_r -\lambda_r^{*} \cc_r^*\|_2$, followed by using Lemma \ref{aulem:normunnorm} which shows that $\|\cc_r -\cc_r^*\|_2 \leq \frac{2}{\lambda_r^{*}} \|\tilde{\cc}_r -\lambda_r^{*} \cc_r^*\|_2$. \\
Let $\Dm$, $\Em$, $\Fm$, $\Gm$, be $n\times n$ diagonal matrices with the following diagonal elements,\\
\[ \Dm_{kk}= \sum\limits_{i,j}{\delta_{ijk}\ac_{r}^{2}(i)\bc_{r}^{2}(j)} \text{ ;} \text{ } \text{ } \text{ }
 \Em_{kk}= \sum\limits_{i,j} \delta_{ijk}\ac_{r}^{*}(i) \bc_{r}^{*}(j) \ac_r(i) \bc_r(j) ; \]
\[ \Fm_{kk}= \sum\limits_{i,j} \delta_{ijk} \ac_{m}^{*}(i)\bc_{m}^{*}(j)  \ac_r(i)\bc_r(j)   \text{ ;} \text{ } \text{ } \text{ } \Gm_{kk}= \sum\limits_{i,j} \delta_{ijk} \ac_{m}(i) \bc_{m}(j) \ac_r(i)  \bc_r(j), \]
where $\delta_{ijk}$ is a Bernoulli random variable with success probability $p$ and indicates whether the $ijk$-th tensor entry is observed or not.
Then the vector $\tilde{\cc}_{r}$ obtained after one pass of the inner loop of Algorithm \ref{alg:jALS} can be written as
\begin{equation}
\label{c:expr}
\tilde{\cc}_{r} =  \Dm^{-1} \left(\lambda_{r}^{*}\Em \cc_{r}^{*} + \sum\limits_{m \in [R] \setminus r} (\lambda_{m}^{*}\Fm \cc_{m}^{*} - \lambda_{m}\Gm \cc_{m}) + P_{\Omega}(\mathcal{E}_T(\ac_r,\bc_r,\I)) \right).\\
\end{equation}
We make use of the fact that $\|\tilde{\cc}_{r}-\lambda_{r}^{*}\cc_{r}^{*} \|_2  = \|\tilde{\cc}_{r}-\lambda_{r}^{*}\Dm^{-1}\Dm \cc_r^{*} \|_2$, to yield,
\begin{align*}
\|\tilde{\cc}_r-\lambda_{r}^{*}\cc_{r}^{*} \|_2 & =  \|\underbrace{\lambda_{r}^{*}\Dm^{-1} (\Em -\Dm) \cc_{r}^{*}}_{err_1} + \underbrace{\Dm^{-1} \sum\limits_{m \in [R] \setminus r} (\lambda_{m}^{*}\Fm \cc_{m}^{*}- \lambda_{m}\Gm \cc_{m})}_{err_2}+ \underbrace{\Dm^{-1} P_{\Omega}(\mathcal{E}_T(\ac_r,\bc_r,\I))}_{err_3} \|_2
\end{align*}
Applying the triangle inequality to the above expression is very convenient as it breaks its into the three different error terms shown below, each characterizing different sources of error affecting the non-shared component update,
\begin{align}
\label{supeq:1}
\|\tilde{\cc}_r-\lambda_{r}^{*}\cc_{r}^{*} \|_2 \leq \|err_1\|_2+ \|err_2 \|_2 + \|err_3 \|_2,
\end{align}
where $err_1  =  \lambda_{r}^{*}\Dm^{-1} (\Em -\Dm) \cc_{r}^{*}$ can be characterized as the error due to the power method. This error is well understood and does not require meticulous bound control in order to yield the desire result. Also if $\Ten^*$ was a rank $1$ and noiseless tensor, the proof of Lemma \ref{cor:contrac} would reduce to bounding this error term. \\
Unlike $err_1$ discussed above, bounding $err_2  = \Dm^{-1} \sum\limits_{m \in [R] \setminus r} (\lambda_{m}^{*}\Fm \cc_{m}^{*} - \lambda_{m}\Gm \cc_{m})$ represents the main challenge in the proof. It is worth noting that $err_2$ is the error due to the deflation method applied in Algorithm \ref{alg:jALS}.
Two issues arise with bounding this error, the first resides in the non-orthogonality of the tensor $\Ten^*$. If the tensor $\Ten^*$ was orthogonal then a deflation algorithm would have little to no difficulty differentiating between the ranks of the tensor. However with the non-orthogonality assumption we are left with a non disappearing residual due to fact that for example two component vectors of the tensor $\cc_r$ and $\cc_j$ could be close to parallel making it difficult for the algorithm to differentiate between the two. Moreover $err_2$ exposes the relationship that exists between recovering a component $\cc_r$ and the error for the other mode components $\ac_j, \bc_j$ and with $ j \neq r$. If not carefully controlled, $err_2$ could cause the estimate $\cc_r$ to diverge from $\cc_{r}^{*}$. Assumption (1.\textit{iii}) is therefore used and required to control the magnitude of $err_2$. \\
The third error term $err_3  =  \Dm^{-1} P_{\Omega}(\mathcal{E}_T(\ac_r,\bc_r,\I))$ is simply the error due to the noise of the tensor and can be easily bounded after standard assumptions are made about the spectral norm of $\Er_T$. Another challenge in bounding the error of the $\cc_r$ update comes from the fact that the tensor has missing entries. As represented in equation \eqref{subeq:updatebc2} the operations involved in computing the update $\cc_r$ is only carried on the observed entries of the tensor. This computation caveat forces the use of concentration inequalities in the analysis of the error bound of the component. Choosing the right concentration inequality becomes therefore very important in order to guarantee a given convergence rate while allowing some reasonable constraints on the tensor entry reveal probability to $p$. The rest of the proof consists in finding a bound for each of the three errors discussed above.
We start with bounding the first error term. Using the fact that $\| \cc^{*}_r \|_2= 1$ and since $\Dm^{-1} (\Em -\Dm)$ is a diagonal matrix its spectral norm is the maximum absolute value of its diagonal elements, we get
\begin{align*}
\|err_1\|_2  &\leq \|\lambda_{r}^{*}\Dm^{-1} (\Em -\Dm)\|_2\\
&= \lambda_{r}^{*} \max\limits_{k} |\Dm^{-1} (\Em -\Dm)|_{kk} \\
& \leq \lambda_{r}^{*}  \max\limits_{k} |\Dm^{-1}|_{kk} \max\limits_{k} |(\Em -\Dm)|_{kk}.
\end{align*}
Next is finding an upper bound for the maximum of each of the random elements in the equation above with high probability. To do that we first get an upper bound for each of the diagonal elements with high probability and make use of the union bound method.
This is derived as:
\begin{align*}
  |(\Em -\Dm)_{kk}| & = |\sum \limits_{ij} \delta_{ijk}\ac_{r}^{*}(i) \bc_{r}^{*}(j) \ac_{r}(i)\bc_{r}(j) - \sum \limits_{jk} \delta_{ijk} \ac_{r}^{2}(i)\bc_{r}^{2}(j)|\\
  & = |\sum \limits_{ij} \delta_{ijk}\ac_{r}^{*}(i) \dc_{b_r}(j) \ac_{r}(i)\bc_{r}(j) -\sum \limits_{ij} \delta_{ijk}\dc_{a_r}(i) \bc_{r}^{*}(j) \ac_{r}(i)\bc_{r}(j) \\
  & -\sum \limits_{ij} \delta_{ijk}\dc_{a_r}(i) \dc_{b_r}(j) \ac_{r}(i)\bc_{r}(j) |.
  \end{align*}
  The expression on the right side of the equality are obtained from the fact that $\ac_{r}(i)= \ac_{r}^{*}(i)+ \dc_{a_r}(i)$ and $\bc_{r}(j)= \bc_{r}^{*}(j)+ \dc_{b_r}(j)$. Next Lemma \ref{lem:lema3} is used to bound the three random elements inside the absolute value. Combined with the triangle inequality and the fact that $ |\langle 
  \dc_{a_r},\ac_{r}^{*} \rangle |=\frac{1}{2}\|\dc_{a_r} \|_2^2$ (Lemma \ref{aulem:inner}) yields the following,
  \begin{align*}
 |(\Em -\Dm)_{kk}|  & \leq p \left (|\langle \ac_{r}^{*} {,} \ac_{r} \rangle \langle  \dc_{b_r} {,} \bc_{r} \rangle| + |\langle \dc_{a_r}{,} \ac_{r} \rangle \langle  \bc_{r}^{*}  {,} \bc_{r} \rangle | +  |\langle \dc_{a_r}{,} \ac_{r} \rangle \langle  \dc_{b_r}  {,} \bc_{r} \rangle|\right) \\
  & + p\gamma \left(\|\dc_{a_r} \|_2 + \|\dc_{b_r} \|_2 + \|\dc_{a_r} \|_2 \|\dc_{b_r} \|_2 \right)\\
  & \leq 6p \left( \max\limits_{\uc_r  \in \{a_r, b_r\}} \big\{ \sqrt{1-\frac{\|d_{u_r}\|_2}{2}}\|\dc_{u_r} \|_{2}^{2}, \|\dc_{u_r} \|_{2}^{4}, \gamma \|\dc_{u_r} \|_{2} \big\} \right )\\
& = 6p \max\limits_{\uc_r  \in \{a_r, b_r\}}  \left( \sqrt{1-\frac{\|d_{u_r}\|_2}{2}}\|\dc_{u_r} \|_{2}, \|\dc_{u_r} \|_{2}^{3}, \gamma \right) \|\dc_{u_r} \|_2 .  \stepcounter{equation}\tag{\theequation} \label{ceq:1}
\end{align*}
The above inequality holds with probability $ 1- 2n^{-10}$ provided the reveal probability $p\geq \frac{C\mu ^3(1+\gamma/3) \log^2(n^{10}) }{n^{3/2}\gamma^2}$. %Also $\gamma\leq C/\log{n}$, where $C$ is a global constant.\\
Using \eqref{ceq:1} and the bound from Lemma \ref{lem:lema2}, we get
\begin{align}
\label{supeq:errc1}
\|err_1\|_2 \leq\frac{6p \lambda_r^{*} \max\limits_{\uc_r  \in \{a_r, b_r\}} \left( \sqrt{1-\frac{\|d_{u_r}\|_2}{2}}\|\dc_{u_r} \|_{2}, \|\dc_{u_r} \|_{2}^{3}, \gamma \right) \|\dc_{u_r} \|_2 }{p(1-\gamma)},
\end{align}
with probability $1-2n^{-9}$.% provided $ p \geq  \frac{C\mu ^3(1+\gamma/3) \log_2(n^{10}) }{n^{3/2}\gamma^2}$.\\

Next we work on bounding $err_2$. Note that
\begin{align*}
    \|err_2\| &= \|\Dm^{-1} \sum\limits_{m \in [R] \setminus r} (\lambda_{m}^{*}\Fm \cc_{m}^{*} - \lambda_{m}\Gm \cc_{m})\|_2 \\
    & \leq  \max \limits_{kk} |\Dm^{-1}|_{kk}  \sum\limits_{m \in [R] \setminus r}\| (\lambda_{m}^{*}\Fm \cc_{m}^{*} - \lambda_{m}\Gm \cc_{m})\|_2 \\
    & =  \max \limits_{kk} |\Dm^{-1}|_{kk}  \sum\limits_{m \in [R] \setminus r} \lambda_{m}^{*}\| \Fm \cc_{m}^{*} - \Gm \cc_{m}+ \Delta_{\lambda_m}\Gm \cc_{m}\|_2 \\
       & \leq \max \limits_{kk} |\Dm^{-1}|_{kk}  \sum\limits_{m \in [R] \setminus r}  \lambda_{m}^{*}\left( \|(\Fm- \Gm)\cc_{m}^{*}\|_2 + \|\Gm \dc_{\cc_m}\|_2 + \| \Delta_{\lambda_m}\Gm \cc_{m}\|_2 \right). \stepcounter{equation}\tag{\theequation}\label{eq:eq26}
\end{align*}
We focus on bounding each of the four components in the last inequality above as
\begin{align*}
    \|\Fm \cc_{m}^{*} - \Gm \cc_{m}\|_2 & \leq  \|(\Fm- \Gm)\cc_{m}^{*}\| + \|\Gm \dc_{\cc_m}\|_2 \\
    &= \max\limits_{i} |\Fm_{ii}-\Gm_{kk}|\|\cc_{m}^{*}\|_2+ \max\limits_{i}|\Gm_{kk}| \|\dc_{\cc_m}\|_2. \stepcounter{equation}\tag{\theequation}\label{eq:eq27}
\end{align*}
Just like we did for $err_1$ we bound each element $|\Fm_{kk}-\Gm_{kk}|$ then apply the union bound to get the bound its maximum,
\begin{align*}
|\Fm_{kk}-\Gm_{kk}|&= |\sum\limits_{jk}\delta_{ijk} \ac_{m}^{*}(i)\bc_{m}^{*}(j)\ac_{r}(i)\bc_{r}(j) - \sum\limits_{jk}\delta_{ijk}\ac_{m}(i) \bc_{m}(j) \ac_{r}(i)\bc_{r}(j)|\\
& \leq |\sum\limits_{jk}\delta_{ijk} \dc_{a_m}(i)\bc_{m}^{*}(j)\ac_{r}(i)\bc_{r}(j)| + |\sum\limits_{jk}\delta_{ijk}\ac_{m}^{*}(i) \dc_{b_m}(j) \ac_{r}(i)\bc_{r}(j)|\\
& + | \sum\limits_{jk}\delta_{ijk} \dc_{a_m}(i)\dc_{b_m}(j)\ac_{r}(i)\bc_{r}(j)|\\
& \leq p \left( |\langle \dc_{a_m}{,} \ac_{r} \rangle \langle  \bc_{m}^{*}  {,} \bc_{r} \rangle |+ |\langle \ac_{m}^{*} {,} \ac_{r} \rangle \langle  \dc_{b_m} {,} \bc_{r} \rangle| +  |\langle \dc_{a_m}{,} \ac_{r} \rangle \langle  \dc_{b_m}  {,} \bc_{r} \rangle| \right )\\
& + \gamma (\|\dc_{a_m} \|_2 + \|\dc_{b_m} \|_2 + \|\dc_{a_m} \|_2 \|\dc_{b_m} \|_2)\\
&\leq 6p  \max\limits_{\uc \in \{\ac_m,\bc_m,\ac_r,\bc_r, \}}\left( (\frac{c_0}{\sqrt{n}}+\|\dc_{\uc} \|_2) \|\dc_{\uc} \|_2,\gamma \|\dc_{\uc} \|_2 \right).
\end{align*}
The last inequality above holds with probability $ 1- 2n^{-10}$ provided the reveal probability $p\geq \frac{C\mu ^3(1+\gamma/3) \log^2(n^{10}) }{n^{3/2}\gamma^2}$. The second inequality is obtained by using Lemma \ref{lem:lema4} and the last inequality is obtained using the incoherence assumption (1.iii) to get that $\max\{|\langle \ac_{m}^{*} {,} \ac_{r} \rangle|, |\langle \bc_{m}^{*} {,} \bc_{r} \rangle|\}  \leq \frac{c_0}{\sqrt{n}}+\max\{\|\dc_{\ac_r}\|_2,\|\dc_{\bc_r} \|_2 \}.$ 
Using the union bound we get that 
\begin{align}
\label{eq:eq28}
\max\limits_{k}|\Fm_{kk}-\Gm_{kk}| & \leq 6p  \max\limits_{\uc \in \{\ac_m,\bc_m,\ac_r,\bc_r, \}}\left( (\frac{c_0}{\sqrt{n}}+\|\dc_{u} \|_2), \gamma  \right)\|\dc_{u} \|_2,
\end{align}
with probability $ 1- 2n^{-9}$.\\
Similarly using Lemma \ref{lem:lema4}, and applying the union bound and the fact that,
\begin{align}
|\langle\ac_{m} {,}  \ac_{r}\rangle\langle \bc_{m} {,}  \bc_{r}\rangle | & \leq  \max \{\langle\ac_{m} {,}  \ac_{r}\rangle^2, \langle \bc_{m} {,}  \bc_{r} \rangle^2 \} \\
 & \leq \left(\frac{c_0}{\sqrt{n}}+ \max\limits_{\uc_r \in \{\ac_r , \bc_r\}}3\|\dc_{u_r}\|_2\right)^2,
\end{align}
yields the following inequality,
\begin{align}
\label{eq:eq 20}
    \max\limits_{k}|\Gm_{kk}| \leq p \max\limits_{\uc_r \in \{\ac_r , \bc_r,\ac_m , \bc_m\}} \left((\frac{c_0}{\sqrt{n}}+ 3\|\dc_{u_r}\|_2)^2,
    \gamma\right),
\end{align}
with probability $ 1- 2n^{-9}$.%. and $p \geq \frac{C\mu^4(1+\gamma/3) \log(\frac{1}{2}n^{10}) }{n^{2}\gamma^2}$ granted $\gamma\leq C/ \log(n)$.\\

Putting equations \eqref{eq:eq26}, \eqref{eq:eq27}, \eqref{eq:eq 20} and using Lemma \ref{lem:lema2} to bound
$\Dm^{-1}$ yields,
\begin{equation}
\label{supeq:errc2}
    \|err_2\|_2 \leq \frac{ 8p  \sum\limits_{m \in [R] \setminus r}\lambda_m^{*} \max\limits_{\uc \in \{\ac_m,\bc_m,\ac_r,\bc_r, \}}\left( (\frac{c_0}{\sqrt{n}}+\|\dc_{\uc} \|_2),(\frac{c_0}{\sqrt{n}}+3\|\dc_{\uc} \|_2)^2, \gamma  \right)\|\dc_{\uc} \|_2}{p(1-\gamma)},
\end{equation}
with probability $1-2n^{-9}$ provided $ p \geq \frac{C\mu ^3(1+\gamma/3) \log^2(n^{10}) }{n^{3/2}\gamma^2}$.% and $\gamma\leq C/ \log(n)$.\\

Next we use Lemma \ref{lem:lemmatenspec}, combined with Lemma \ref{lem:lema2} to bound the $\|err_3\|_2$. Note that $\|err_3\|_2 = \|\Dm^{-1} P_{\Omega}(\mathcal{E}_T(\ac_r,\bc_r,\I))\|_2 \leq  \max \limits_{kk} |\Dm^{-1}|_{kk} \|P_{\Omega}(\mathcal{E}_T(\ac_r,\bc_r,\I))\|_2$. Denote $\be_k$ as the vector whose entries are zero except that the $k$-th entry is one. Remind that $\delta_{ijk}$ is a Bernoulli random variable with success probability $p$. Note that 
$
\|P_{\Omega}(\mathcal{E}_T(\ac_r,\bc_r,\I))\|_2 = \Big\|\sum_{i,j,k} \delta_{ijk} (\mathcal{E}_T)_{ijk}a_{ri}b_{rj}\be_k\Big\|_2 = \sqrt{\sum_{i,j,k} \delta_{ijk} (\mathcal{E}_T)_{ijk}^2 a_{ri}^2 b_{rj}^2 \be_k^{\top} \be_k}.
$
Since $\delta_{ijk}$ is a Bernoulli random variable with success probability $p$ and using a similar concentration argument to Lemma \ref{lem:lema2}, we have that $\sqrt{\sum_{i,j,k} \delta_{ijk} (\mathcal{E}_T)_{ijk}^2 a_{ri}^2 b_{rj}^2 \be_k^{\top} \be_k} \le \sqrt{p}(1+\gamma) \sqrt{\sum_{i,j,k} (\mathcal{E}_T)_{ijk}^2 a_{ri}^2 b_{rj}^2 \be_k^{\top} \be_k} \le \sqrt{p}(1+\gamma)\|\mathcal{E}_T\|$. Therefore, we have
\begin{equation}
\label{supeq:errc3}
\|err_3\|_2   \leq \frac{\sqrt{p}(1+\gamma)\|\mathcal{E}_T\|}{p(1-\gamma)},
\end{equation}
with probability $1-2n^{-9}$ provided $ p \geq \frac{C\mu ^3(1+\gamma/3) \log^2(n^{10}) }{n^{3/2}\gamma^2}$.%  and $\gamma\leq C/ \log(n)$.\\
Combining the error bounds results of $\|err_1\|_2$, $\|err_2\|_2$, $\|err_3\|_2$ in equations \eqref{supeq:errc1}, \eqref{supeq:errc2} and \eqref{supeq:errc3} respectively, yields
\begin{align*}
    &\|\tilde{\cc}_r-\lambda_{r}^{*}\cc_{r}^{*} \|_2 \\ &\leq  \frac{ 8pR \lambda_{max}^{*}  \max\limits_{\uc \in \{\ac_m,\bc_m,\ac_r,\bc_r, \}}\left(  \sqrt{1-\frac{\|d_{u}\|_2}{2}}\|\dc_{u} \|_{2}, (\frac{c_0}{\sqrt{n}}+\|\dc_{u} \|_2),(\frac{c_0}{\sqrt{n}}+3\|\dc_{u} \|_2)^2,\|\dc_{u} \|_2^3, \gamma  \right)\|\dc_{u} \|_2}{p(1-\gamma)}\\
    & + \frac{\sqrt{p}(1+\gamma)\|\mathcal{E}\|}{p(1-\gamma)}\stepcounter{equation}\tag{\theequation}\label{supeq:boundcu},
    \end{align*}
with probability $1-2n^{-9}$. The proof of Lemma \ref{cor:contrac} is then completed by applying the results of Lemma \ref{aulem:normunnorm} which shows that $\|\cc_r -\cc_r^*\|_2 \leq \frac{2}{\lambda_r^{*}} \|\tilde{\cc}_r -\lambda_r^{*} \cc_r^*\|_2$ and Lemma \ref{lem:lema5} ($|\lambda_r -\lambda_r^{*}| \leq \|\tilde{\cc_r} - \lambda_r^{*}\cc_r^{*} \|_2$) and by letting $\max\{\|\dc_{\uc}\|_2\}=\epsilon_T$. \eop

\subsubsection{Proof of Lemma \ref{cor:contraa}}
\label{proof:contraa} 
We now prove the contraction result in one iteration of Algorithm \ref{alg:jALS} for the shared components of the tensor and matrix $\ac_r$ in the special case where the tensor and matrix weights are equal and both tensor and matrix are dense. 
When the tensor and matrix weight are assumed to be equal, the close form solution for the update of the shared tensor component derived in Lemma \ref{lem:update} simplifies to  
$\ac_r = \Tfrac{(\text{res}_T(\mathbf{I}, \bc_r, \cc_r)  +  \text{res}_M \vc_r)}{\lambda_r( P_\Omega(\mathbf{I}, (\bc_r).^2, (\cc_r).^2 ) + 1 )}$. In this special case we can still employ the same technique used in bounding the non-shared components by using the intermediate step of bounding the expression $\|\tilde{\ac}_r -\lambda_r^{*} \ac_r^*\|_2$ where 
$\tilde{\ac}_r =  \Tfrac{(\text{res}_T(\mathbf{I}, \bc_r, \cc_r)  +  \text{res}_M \vc_r)}{ P_\Omega(\mathbf{I}, (\bc_r).^2, (\cc_r).^2 ) + 1 }$. \\
This is the main advantage of restricting the problem to the equal tensor matrix weight case as it allows the proof technique derived for the non-shared component to be easily extended to the case of the shared component. As we will show in the analysis of Lemma \ref{cor:contraas} this advantage disappears when the weight of the tensor ans matrix are allowed to be different.

Let $\Dm$, $\Em$, $\Fm$, $\Gm$, $\Hm$, $\Jm$, $\Pm$ be $n\times n$ diagonal matrices with diagonal elements,\\
\[ \Dm_{ii}= \sum\limits_{j,k}{\delta_{ijk}\bc_{r}^{2}(j) \cc_{r}^{2}(k)}+ 1 \text{ ;} \text{ } \text{ } \text{ }
 \Em_{ii}= \sum\limits_{j,k} \delta_{ijk} \bc_{r}^{*}(j) \cc_{r}^{*}(k) \bc_r(j) \cc_r(k); \]
\[ \Fm_{ii}= \sum\limits_{j,k} \delta_{ijk} \bc_{m}^{*}(j) \cc_{m}^{*}(k) \bc_r(j) \cc_r(k)  \text{ ;} \text{ } \text{ } \text{ } \Gm_{ii}= \sum\limits_{j,k} \delta_{ijk} \bc_{m}(j) \cc_{m}(k) \bc_r(j) \cc_r(k) ; \]
\[ \Hm_{ii}= \sum\limits_{l}\vc_{r}^{*}(l)\vc_r(l) \text{ ;} \text{ } \text{ } \text{ } \Jm_{ii}= \sum\limits_{l}\vc_{m}^{*}(l)\vc_r(l) \text{ ;} \text{ } \text{ } \text{ } \Pm_{ii}= \sum\limits_{l}\vc_{m}(l)\vc_r(l).\] 
Then the vector $\tilde{\ac}_{r}$ obtained after one pass of the inner loop of Algorithm \ref{alg:jALS} can be written as
\begin{align*}
\tilde{\ac}_{r} &= \Dm^{-1} \left(\lambda_{r}^{*}\Em \ac_{r}^{*} + \sum\limits_{m \in [R] \setminus r} (\lambda_{m}^{*}\Fm \ac_{m}^{*} - \lambda_{m}\Gm \ac_{m}) +P_\Omega(\mathcal{E}_T(\I,\bc_r,\cc_r)) \right)\\
& + \Dm^{-1} \left(\lambda_{r}^{*} \Hm \ac_{r}^{*}+ \sum\limits_{m\in [R] \setminus r} (\lambda_{m}^{*}\Jm \ac_{m}^{*}- \lambda_{m}\Pm \ac_{m})+ \mathcal{E}_M\vc_r  \right).\stepcounter{equation}\tag{\theequation}\label{supeq:aar}
\end{align*}
In the next steps we bound 
\begin{align*}
\|\tilde{\ac}_{r}-\lambda_r^{*} \ac_{r}^{*} \|_2 & \leq  \|\underbrace{ \lambda_r^{*}\Dm^{-1}\left(\Em + \Hm -\Dm \right) a_{r}^{*}}_{err_1}\|_2 + \|\underbrace{ \Dm^{-1} \sum\limits_{m \in [R] \setminus r} (\lambda_{m}^{*}\Fm a_{m}^{*} - \lambda_{m}\Gm a_{m}}_{err_2})\|_2 \\
& + \|\underbrace{ \Dm^{-1} \sum\limits_{m\in [R] \setminus r} (\lambda_{m}^{*}\Jm a_{m}^{*}- \lambda_{m}\Pm a_{m})}_{err_3}\|_2 
 +\|\underbrace{ \Dm^{-1} (P_\Omega(\mathcal{E}_T(\I,\bc_r,\cc_r)) + \mathcal{E}_M\vc_r)}_{err_4}\|_2. \stepcounter{equation}\tag{\theequation}\label{supeq:2}
\end{align*}

In the shared component case, the right hand side of equation \eqref{supeq:2} can be characterized as the sum of 4 sources of errors, where $err_1  = \lambda_r^{*}\Dm^{-1}\left(\Em + \Hm -\Dm \right) a_{r}^{*}$ can be characterized as the error due to the power method applied to both the tensor and matrix. This error is similar to $err_1$ discussed in the proof of Lemma \ref{cor:contrac} with the exception that it factors in the contribution of the matrix. Again, if $\Ten^*$ was a rank $1$,  noiseless tensor, then proving Lemma \ref{cor:contraa} would reduce to bounding this term.
The second and third sources of error  $err_2  = \Dm^{-1} \sum\limits_{m \in [R] \setminus r} (\lambda_{m}^{*}\Fm a_{m}^{*} - \lambda_{m}\Gm a_{m})$ and $err_3 = \Dm^{-1} \sum\limits_{m\in [R] \setminus r} (\lambda_{m}^{*}\Jm a_{m}^{*}- \lambda_{m}\Pm a_{m}) $ again represents the main challenge in the proof. The challenge in bounding these two errors are very similar to those exposed for $err_2$ in the analysis of Lemma \ref{cor:contrac} in addition to the fact that we have an extra residual due to the matrix. If both the tensor and matrix components were orthogonal this error would be non existent. We therefore partly control these errors magnitude through the bound imposed on the components vector inner product namely Assumption (1.\textit{iii})the incoherence assumption.
The fourth error term $err_4  =  \Dm^{-1} (P_\Omega(\mathcal{E}_T(\I,\bc_r,\cc_r)) + \mathcal{E}_M\vc_r$ is simply the error due to the noise of the tensor and the matrix and can be easily bounded after standard Assumptions are made about the spectral norms of $\Er_T$ and $\Er_M$. At first glance it might seem that right hand-side of the inequalities in equation \eqref{supeq:2} is larger than that found in equation  \eqref{supeq:1} making therefore the bound on the shared component larger than that of the that of the non-shared component. However as we demonstrate in the proof below, the component $\Dm^{-1}$ plays the role of a weight which averages the tensor and matrix sources of error in equation \eqref{supeq:2}.

We start with bounding the first error term,
\begin{align*}
\|err_1\|_2 & = \|\lambda_r^{*}\Dm^{-1}\left(\Em + \Hm -\Dm \right) a_{r}^{*}\|_2 \\
& \leq \lambda_r^{*} \|\Dm^{-1}\left(\Em +\Hm -\Dm \right)\|_2 \|a_{r}^{*}\|_2 \\
& \leq \lambda_r^{*} \max \limits_{i} |\Dm^{-1}_{ii}| |\left( \Em +\Hm -\Dm \right)_{ii}|,
\end{align*}
where last inequality above is obtained by observing that $\Dm^{-1}\left( \Em + \Hm -\Dm \right)$ is a diagonal matrix whose spectral norm is the maximum absolute value of its diagonal elements and that $\|a_{r}^{*}\|_2 =1$. We proceed to getting an upper bound for each of the maximum of each of the random variable elements in the equation above with high probability. To do that we first get an upper bound on each of the
diagonal elements with high probability and make use of the union bound method to get a high
probability bound on the maximums.
\begin{align*}
    |( \Em + \Hm -\Dm)_{ii}| & \leq   | \langle \vc_{r}^{*} {,}  \vc_{r} \rangle -  1 | + |\sum \limits_{jk} \delta_{ijk} \bc_{r}^{*}(j)\cc_{r}^{*}(k)\bc_{r}(j)\cc_{r}(k)- \sum \limits_{jk} \delta_{ijk} \bc_{r}^{2}(j)\cc_{r}^{2}(k)|\\
  & = \frac{1}{2} \|\dc_v\|_2^2+ |\sum \limits_{ij} \delta_{ijk}\ac_{r}^{*}(i) \dc_{b_r}(j) \ac_{r}(i)\bc_{r}(j) -\sum \limits_{ij} \delta_{ijk}\dc_{a_r}(i) \bc_{r}^{*}(j) \ac_{r}(i)\bc_{r}(j) \\
  & -\sum \limits_{ij} \delta_{ijk}\dc_{c_r}(i) \dc_{b_r}(j) \cc_{r}(i)\bc_{r}(j) |\\
  &\leq \frac{1}{2} \|\dc_v\|_2^2+ p \left (|\langle \cc_{r}^{*} {,} \cc_{r} \rangle \langle  \dc_{b_r} {,} \bc_{r} \rangle| + |\langle \dc_{c_r}{,} \cc_{r} \rangle \langle  \bc_{r}^{*}  {,} \bc_{r} \rangle | +  |\langle \dc_{c_r}{,} \cc_{r} \rangle \langle  \dc_{b_r}  {,} \bc_{r} \rangle|\right) \\  
  & + p\gamma \left(\|\dc_{c_r} \|_2 + \|\dc_{b_r} \|_2 + \|\dc_{c_r} \|_2 \|\dc_{b_r} \|_2 \right).
  \end{align*}
  The expression on the right side of the equality is obtained by combining the triangle inequality to the fact that $\cc_{r}(i)= \cc_{r}^{*}(i)+ \dc_{c_r}(i)$ $\bc_{r}(j)= \bc_{r}^{*}(j)+ \dc_{b_r}(j)$ and using the results from Lemma \ref{aulem:inner}. We then use Lemma \ref{lem:lema3} to bound the three random elements inside the absolute value. Hence, provided the reveal probability $ p \geq \frac{C\mu ^3(1+\gamma/3) \log^2(n^{10}) }{n^{3/2}\gamma^2}$ we get,
  \begin{align*}
    |( \Em + \Hm -\Dm)_{ii}| 
   |( \Em + \Hm -\Dm)_{ii}| & \leq \frac{1}{2} \|\dc_v\|_2^2+ 6p \left( \max\limits_{\uc_r  \in \{c_r, b_r\}} \big\{ \sqrt{1-\frac{\|d_{u_r}\|_2}{2}}\|\dc_{u_r} \|_{2}^{2}, \|\dc_{u_r} \|_{2}^{4}, \gamma \|\dc_{u_r} \|_{2} \big\} \right )\\
& \leq \frac{1}{2} \|\dc_v\|_2^2+ 6p \max\limits_{\uc_r  \in \{c_r, b_r\}}  \left( \sqrt{1-\frac{\|d_{u_r}\|_2}{3}}\|\dc_{u_r} \|_{2}, \|\dc_{u_r} \|_{2}^{3}, \gamma \right) \|\dc_{u_r} \|_2, \stepcounter{equation}\tag{\theequation}\label{supeq:a1}
  \end{align*}
with probability $ 1- 2n^{-10}$.
Using the union bound on the result in equation \eqref{supeq:a1} combined with the results of  Lemma \ref{lem:lema2}. We get,
\begin{equation}
\label{supeq:erra1}
\|err_1\|_2 \leq  \frac{ \lambda_r^{*}\left(6p \max\limits_{\uc_r  \in \{a_r, b_r\}} \left( \sqrt{1-\frac{\|d_{u_r}\|_2}{2}}\|\dc_{u_r} \|_{2}, \|\dc_{u_r} \|_{2}^{3}, \gamma \right) \|\dc_{u_r} \|_2 + 1/2 \|\dc_v\|_2^2\right)}{p(1-\gamma)+1}
\end{equation}
with probability  $ 1- 2n^{-9}$.\\
Next we proceed to bound $\|err_3\|_2$ before coming back to $\|err_2\|_2$,
\begin{align*}
    \|err_3\|_2 = \|\Dm^{-1} \sum\limits_{m\in [R] \setminus r} (\lambda_{m}^{*}\Jm a_{m}^{*}- \lambda_{m}\Pm a_{m})\|_2.
\end{align*}
We start by bounding the component inside the summation.
\begin{align*}
    \|\lambda_{m}^{*}\Jm a_{m}^{*}- \lambda_{m}\Pm a_{m}\|_2 &= \|\lambda_{m}^{*}\langle \vc_{m}^{*} {,}  \vc_{r} \rangle \ac_{m}^{*}- \lambda_{m}\langle \vc_{m} {,}  \vc_{r} \rangle \ac_{m}\|_2\\
     & = \lambda_{m}^{*}\|(\langle \vc_{m}^{*} {,}  \vc_{r} \rangle - \langle \vc_{m} {,}  \vc_{r} \rangle)\ac_{m}^{*} + \langle \vc_{m} {,}  \vc_{r} \rangle \dc_{a_m} + \Delta_{\lambda_{m}}\langle \vc_{m} {,}  \vc_{r} \rangle \ac_{m}  \|_2\\
    & \leq  3\lambda_m^{*}\max \left(  \|\dc_{\vc_m}\|_2,\frac{c_0}{\sqrt{n}}+3\|\dc_{v_r} \|_2 \right) \|\dc_{v_r} \|_2,
 \stepcounter{equation}\tag{\theequation}\label{eq:eq17}
\end{align*}
where the last inequality is due to the fact that $\langle \vc_{m} {,}  \vc_{r} \rangle \leq (\frac{c_0}{\sqrt{n}}+ 3\|\dc_{v_r}\|_2)$. This, combined with the results of Lemma \ref{lem:lema2} to bound $|\Dm^{-1}|$ yields,
\begin{equation}
\label{supeq:erra3}
    \|err_3\|_2 \leq \frac{3\sum \limits_{m\in [R] \setminus r} \lambda_m^{*}\max ( \|\dc_{\vc_m}\|_2 ,\frac{c_0}{\sqrt{n}}+3\|\dc_{v_r} \|_2 ) \|\dc_{v_r} \|_2}{p(1-\gamma)+1},
\end{equation}
with probability $1-2n^{-9}$.\\
The technique used to bound $\|err_2\|_2$ in this section is very similar to the one used to bound expression in section. We therefore provide the bound and incite the reader to review the section mention to understand the process involved. The main difference recedes in substituting the components $\cc$ for $\ac$ and finding a lower bound for $D^{-1}$ using Lemma \ref{lem:lema2}. This yields, 
\begin{equation}
\label{supeq:erra2}
    \|err_2\|_2 \leq \frac{ 8p  \sum\limits_{m \in [R] \setminus r}\lambda_m^{*} \max\limits_{\uc \in \{\cc_m,\bc_m,\cc_r,\bc_r, \}}\left( (\frac{c_0}{\sqrt{n}}+\|\dc_{u} \|_2),(\frac{c_0}{\sqrt{n}}+3\|\dc_{u} \|_2)^2\|, \gamma  \right)\|\dc_{u} \|_2}{p(1-\gamma)+1},
\end{equation}
with probability $1-2n^{-9}$.\\
Next $\|err_4\|_2$ is bounded using Lemma \ref{lem:lemmatenspec}, Lemma \ref{lem:lema2} and the fact that $\|\mathcal{E}_M \vc_r\|_2 \leq \|\mathcal{E}_M\|$ since $\|\vc_r\|_2$=1 and by definition $\|\mathcal{E}_M\|=\sup_{\substack{\| \uc \| = 1}} \|{\Er_M \uc }\|_2$. Similar to the proof of $(\ref{supeq:errc3})$, we obtain
\begin{equation}
\label{supeq:erra4}
\|err_4\|_2 \leq \frac{\sqrt{p}(1+\gamma)\|\mathcal{E}_T\|+ \|\mathcal{E}_M\| }{p(1-\gamma)+1 }\\
\end{equation}
with probability $1-2n^{-9}$.\\
Combining the error bounds results of $\|err_1\|_2$, $\|err_3\|_2$, $\|err_2\|_2$, $\|err_4\|_2$ in equations \eqref{supeq:erra1}, \eqref{supeq:erra2}, \eqref{supeq:erra3} and \eqref{supeq:erra4} respectively, we get
\begin{align*}
    &\|\tilde{\ac}_r-\lambda_{r}^{*}\ac_{r}^{*} \|_2 \\
    &\leq  \frac{ 8pR \lambda_{max}^{*}  \max\limits_{\uc \in \{\cc_m,\bc_m,\cc_r,\bc_r, \}}\left(  \sqrt{1-\frac{\|d_{u}\|_2}{2}}\|\dc_{u} \|_{2}, (\frac{c_0}{\sqrt{n}}+\|\dc_{u} \|_2),(\frac{c_0}{\sqrt{n}}+3\|\dc_{u} \|_2)^2,\|\dc_{u} \|_2^3, \gamma  \right)\|\dc_{u} \|_2}{p(1-\gamma)+1} \\
    & +\frac{3R \lambda_{max}^{*}\max \left( \|\dc_{v_r} \|_2,\frac{c_0}{\sqrt{n}}+3\|\dc_{v_r} \|_2 \right) \|\dc_{v_r} \|_2  +\sqrt{p}(1+\gamma)\|\mathcal{E}_T\|+ \|\mathcal{E}_M\|}{p(1-\gamma)+1} \stepcounter{equation}\tag{\theequation}\label{supeq:boundau}
\end{align*}
with probability $1-2n^{-9}$.\\
The proof of Lemma \ref{cor:contraa} is then completed by applying the results of Lemma \ref{aulem:normunnorm} which shows that $\|\ac_r -\ac_r^*\|_2 \leq \frac{2}{\lambda_r^{*}} \|\tilde{\ac}_r -\lambda_r^{*} \ac_r^*\|_2$ and letting $\max\{\|\dc_{\uc}\|_2\}=\epsilon_T$ and $\max\{\|\dc_{\vc}\|_2\}=\epsilon_M$. \eop

\subsubsection{Proof of Lemma \ref{cor:contraas}}
We now prove Lemma \ref{cor:contraas} which establishes an error contraction result for the shared tensor components in one iteration of Algorithm \ref{alg:jALS} when the input tensor and matrix are assumed to be sparse and their respective components weight are allowed to differ.
First, we introduce some notation below in order reveal how we address the sparse components in the analysis .\\
Define $F_a := \textrm{supp}(\ac_r^*) \cup \textrm{supp}(\ac_r)$, $F_b := \textrm{supp}(\bc_r^*) \cup \textrm{supp}(\bc_r)$,  $F_c := \textrm{supp}(\cc_r^*) \cup \textrm{supp}(\cc_r)$ and $F_v := \textrm{supp}(\vc_r^*) \cup \textrm{supp}(\vc_r)$ where $\textrm{supp}(\uc)$ refers to the set of indices in a vector $\uc$ that are nonzero. Then let $F$ and $F^{'}$ be compositions of support sets defined as  $F := F_a \circ F_b \circ F_c$ and $F^{'} := F_1 \circ F_v$ respectively. 
We use the notation $\Ten^{\bks r}:=\sum_{m\in [R] \setminus r} \lambda_m \ac_m \otimes\bc_m \otimes \cc_m$ to represent the CP decomposition of the tensor $\Ten$ minus its $r^{th}$ rank $1$ tensor element $(\lambda_r \ac_r \otimes\bc_r \otimes \cc_r)$.\\
Denote the truncated vectors $\uc_{r}^*$ and $\uc_{r}$ to be $\bar{\uc}_{r}^*= \textrm{Truncate}(\uc_{r}^*, {F_{\uc}})$ and $\bar{\uc}_{r} =\textrm{Truncate}(\uc_{r}, {F_{\uc}})$ with $\uc \in \{\ac, \bc,\cc, \vc\}$ and $r=1,\ldots,R$.

Note that in the update of $a_r$ in our algorithm, we first obtain non-sparse estimator $\ac_r$ in line (8) of algorithm \ref{alg:jALS} then update it by applying the truncation  method and normalization method in (9). We let $\dot{\ac}_r$ be the update on line (8) of algorithm \ref{alg:jALS} before the truncation and $\ac_r$ be the truncated update on line (9) of the algorithm. That is ${\ac}_r =\frac{\dot{\ac}_r}{\|\dot{\ac}_r\|_2}$ with,
\[\dot{\ac}_r = \Tfrac{(\lambda_r\text{res}_{T_F}(\mathbf{I}, \bc_r, \cc_r)  + \sigma_r \text{res}_{M_{F'}} \vc_r)}{(\lambda_r ^2 P_\Omega(\mathbf{I}, (\bc_r).^2, (\cc_r).^2 ) + \sigma_r^2  )}
\]
where $\text{res}_{T_F}$ denotes the restriction of the residual tensor $\text{res}_{T}$ on the three modes indexed by $F_a$, $F_b$ and $F_c$ and  $\text{res}_{T_F}$ is the equivalent for the residual matrix $\text{res}_{M}$.  That is
\begin{align*}
\text{res}_{T_F} &= \sum_{m\in [R]} \lambda_m^* \bar{\ac}_m^* \otimes \bar{\bc}_m^* \otimes \bar{\cc}_m^*
-\sum_{m\in [R] \setminus r} \lambda_m \bar{\ac}_m \otimes \bar{\bc}_m \otimes \bar{\cc}_m,\\
\text{res}_{M_F'} &= \sum_{m\in [R]} \sigma_m^* \bar{\ac}_m^* \otimes \bar{\vc}_m^* -\sum_{m\in [R] \setminus r} \lambda_m \bar{\ac}_m \otimes \bar{\vc}_m.
\end{align*}
Proving Lemma \ref{cor:contraas} involves bounding $\|\ac_r-\ac_r^* \|_2$ which we do in two steps. First we notice that $\|\ac_r-\ac_r^* \|_2 \leq \|\ac_r-\dot{\ac}_r \|_2 + \|\dot{\ac}_r-\ac_r^* \|_2$ using the triangle inequality. Then we bound each of the two norms in the expression above. As will be demonstrated in the proof,
$$\|\ac_r-\ac_r^* \|_2 \leq \|\ac_r-\dot{\ac}_r \|_2 + \|\dot{\ac}_r-\ac_r^* \|_2 \leq  2\|\dot{\ac}_r-\ac_r^* \|_2.$$ 
While bounding $\|\ac_r-\ac_r^* \|_2$  directly is a challenge, getting relatively tight upper bounds for $\|\ac_r-\dot{\ac}_r \|_2$ and $ \|\dot{\ac}_r-\ac_r^* \|_2$ although challenging is feasible.\\
\textbf{Step1}: We begin with bounding $\|\dot{\ac}_r-\ac_r^* \|_2$.
% Recovering the tensor and matrix component support necessitate the truncation parameter $s$ be larger or equal to the true support parameter $d$. Hence, we make that assumption all along the proof to follows. 

%The aforementioned assumption then allows us to %rewrite $\text{res}_{T_F}$ and %$\text{res}_{M_F'}$ as
%\begin{align*}
%\text{res}_{T_F} &= \sum_{m\in [R]} \lambda_m^* {\ac}_m^* \otimes {\bc}_m^* \otimes {\cc}_m^*
%-\sum_{m\in [R] \setminus r} \lambda_m \bar{\ac}_m \otimes \bar{\bc}_m \otimes \bar{\cc}_m.\\
%\text{and}\\
%\text{res}_{M_F'} &= \sum_{m\in [R]} \sigma_m^* {\ac}_m^* \otimes {\vc}_m^* -\sum_{m\in [R] \setminus r} \lambda_m \bar{\ac}_m \otimes \bar{\vc}_m
%\end{align*}
 
%%%%%%%%%%%%%%%%%%%%%%
Let $\Dm$, $\Em$, $\Fm$, $\Gm$, $\Hm$, $\Jm$, $\Pm$ be $n\times n$ diagonal matrices with diagonal elements,\\
\[ \Dm_{ii}= \lambda_r^2 \sum\limits_{j,k}{\delta_{ijk}\bc_{r}^{2}(j) \cc_{r}^{2}(k)}+ \sigma_r^2 \text{ ;} \text{ } \text{ } \text{ }
 \Em_{ii}= \sum\limits_{j,k} \delta_{ijk} \bar{\bc}_{r}^{*}(j) \bar{\cc}_{r}^{*}(k) \bc_r(j) \cc_r(k); \]
\[ \Fm_{ii}= \sum\limits_{j,k} \delta_{ijk} \bar{\bc}_{m}^{*}(j) \bar{\cc}_{m}^{*}(k) \bc_r(j) \cc_r(k)  \text{ ;} \text{ } \text{ } \text{ } \Gm_{ii}= \sum\limits_{j,k} \delta_{ijk} \bar{\bc}_{m}(j) \bar{\cc}_{m}(k) \bc_r(j) \cc_r(k) ; \]
\[ \Hm_{ii}= \sum\limits_{l}\bar{\vc}_{r}^{*}(l)\vc_r(l) \text{ ;} \text{ } \text{ } \text{ } \Jm_{ii}= \sum\limits_{l}\bar{\vc}_{m}^{*}(l)\vc_r(l) \text{ ;} \text{ } \text{ } \text{ } \Pm_{ii}= \sum\limits_{l}\bar{\vc}_{m}(l)\vc_r(l).\] 
Then the vector ${\ac}_{r}$ obtained after one pass of the inner loop of Algorithm \ref{alg:jALS} and before normalization can be written as
\begin{align*}
\dot{\ac}_{r} &=  \lambda_r \Dm^{-1} \left(\lambda_{r}^{*}\Em \bar{\ac}_{r}^{*} + \sum\limits_{m \in [R] \setminus r} (\lambda_{m}^{*}\Fm \bar{\ac}_{m}^{*} - \lambda_{m}\Gm \bar{\ac}_{m}) +P_{\Omega}(\mathcal{E}_{T_{F}} \times_2 \bc_r \times_3 \cc_r) \right)\\
& + \sigma_r \Dm^{-1} \left(\sigma_{r}^{*} \Hm \bar{\ac}_{r}^{*}+ \sum\limits_{m\in [R] \setminus r} (\sigma_{m}^{*}\Jm \bar{\ac}_{m}^{*}- \sigma_{m}\Pm \bar{\ac}_{m})+ \mathcal{E}_{M_{F'}}\vc_r \right). \stepcounter{equation}\tag{\theequation}\label{sup:adot}
\end{align*}
This means that
\begin{align*}
&\|\dot{\ac}_r-\ac_{r}^{*} \|_2  = \|\underbrace{ \Dm^{-1}\left(\lambda_r \lambda_r^{*} \Em + \sigma_r \sigma_r^{*}\Hm -\Dm\I\right) \bar{\ac}_{r}^{*}}_{err_1}\|_2
 + \|\underbrace{\lambda_r \Dm^{-1} \sum\limits_{m \in [R] \setminus r} (\lambda_{m}^{*}\Fm \bar{\ac}_{m}^{*} - \lambda_{m}\Gm \bar{\ac}_{m})}_{err_2}\|_2\\ 
 & + \|\underbrace{\sigma_r \Dm^{-1} \sum\limits_{m\in [R] \setminus r} (\sigma_{m}^{*}\Jm {\ac}_{m}^{*}- \sigma_{m}\Pm \bar{\ac}_{m})}_{err_3}\|_2 + \|\underbrace{\Dm^{-1} (\lambda_rP_{\Omega}( \mathcal{E}_{T_{F}} \times_2 \bc_r \times_3 \cc_r) + \sigma_r \mathcal{E}_{M_{F'}}\vc_r) }_{err_4}\|_2. \stepcounter{equation}\tag{\theequation}\label{sup:adot2}
\end{align*}
The right hand side of the inequality above is split into four sources of errors where $err_2$ and $err_3$ are due to tensor rank being greater than one, $err_3$ is the error associated tot the tensor and matrix noise and $err_1$ is the error from the power iteration used in the algorithm. We notice in the case where the tensor and matrix have different weight expression of $\ac_r $ contains the estimated weights unlike when the tensor weights can be assumed to be equal. This main difference requires careful derivation of the error bound for the update of the shared components.

We start with bounding the first error term
\begin{align*}
\|err_1\|_2 & = \|\Dm^{-1}\left(\lambda_r \lambda_r^{*} \Em + \sigma_r \sigma_r^{*}\Hm -\Dm\I\right) \bar{\ac}_{r}^{*}\|_2 \\
& \leq \| \Dm^{-1}\left(\lambda_r \lambda_r^{*} \Em + \sigma_r \sigma_r^{*}\Hm -\Dm\I\right)\|_2 \|\bar{\ac}_{r}^{*}\|_2 \\
& \leq \max \limits_{i} \underbrace{|\Dm^{-1}_{ii}|}_{err_{11}} \underbrace{|\left(\lambda_r \lambda_r^{*} \Em + \sigma_r \sigma_r^{*}\Hm -\Dm\I\right)_{ii}|}_{err_{12}},
\end{align*}
where the third inequality is due to the fact that $\|\bar{\ac}_{r}^{*}\|_2 \leq \|{\ac}_{r}^{*}\|_2=1$ and since, \\ $\Dm^{-1}\left(\lambda_r \lambda_r^{*} \Em + \sigma_r \sigma_r^{*}\Hm -\Dm\I\right)$ is a diagonal matrix hence its spectral norm is obtained by taking the maximum absolute value of its diagonal elements. We therefore proceed to getting an upper bound each of the maximum of each of the random variable elements in the equation above with high probability. To do that we first get an upper bound on each of the diagonal elements with high probability and make use of the union bound method to get a high probability bound on the maximums.
\begin{align*}
  err_{12} & = |\lambda_r \lambda_{r}^{*} \sum \limits_{jk} \delta_{ijk} \bar{\bc}_{r}^{*}(j)\bar{\cc}_{r}^{*}(k)\bc_{r}(j)\cc_{r}(k)+ \sigma_r \sigma_{r}^{*} \langle \bar{\vc}_{r}^{*} {,}  \vc_{r} \rangle - (\lambda_{r}^{2} \sum \limits_{jk} \delta_{ijk} \bc_{r}^{2}(j)\cc_{r}^{2}(k)+ \sigma_{r}^{2})|\\
    & \leq \underbrace{ |\lambda_r \lambda_{r}^{*}  \sum \limits_{jk} \delta_{ijk} \bar{\bc}_{r}^{*}(j)\bar{\cc}_{r}^{*}(k)\bc_{r}(j)\cc_{r}(k)- \lambda_{r}^{2} \sum \limits_{jk} \delta_{ijk} \bc_{r}^{2}(j)\cc_{r}^{2}(k)|}_{I_{121}} + \underbrace{|\sigma_r \sigma_{r}^{*} \langle {\vc}_{r}^{*} {,}  \vc_{r} \rangle -  \sigma_{r}^{2})|}_{I_{122}}.
\end{align*}
We can bound $I_{121}$ and $I_{122}$ next  
\begin{align*}
I_{122} & = |\sigma_r \sigma_{r}^{*} \langle \bar{\vc}_{r}^{*} {,}  \vc_{r} \rangle -  \sigma_{r}^{2})|\\
& \leq \sigma_r \sigma_{r}^{*}(| \langle \vc_{r}^{*} {,}  \vc_{r} \rangle - 1| + \Delta_{\sigma_r})\\
& \leq \sigma_r \sigma_{r}^{*} ( \frac{1}{2} \|d_v\|_2^2, +\Delta_{\sigma_r})\stepcounter{equation}\tag{\theequation}\label{eq:eq14}
\end{align*}
where the first inequality is due to using the triangle inequality, the fact that $\sigma_r =\sigma_r-\sigma_r^* +\sigma_r^*$ and Lemma \ref{aulem:supportvec} by noting that $supp({\vc}_{r}) \subseteq F_b$.
%$\langle \bar{\vc}_{r}^{*} {,}  \vc_{r} \rangle %\leq \langle \vc_{r}^{*} {,}  \vc_{r} \rangle$ . 
The second inequality is obtained from the results of Lemma \ref{aulem:inner}. Next we also bound $I_{121}$.
\begin{align*}
 I_{121}&=|\lambda_r \lambda_{r}^{*} \sum \limits_{jk} \delta_{ijk} \bar{\bc}_{r}^{*}(j)\bar{\cc}_{r}^{*}(k)\bc_{r}(j)\cc_{r}(k)- \lambda_{r}^{2} \sum \limits_{jk} \delta_{ijk} \bc_{r}^{2}(j)\cc_{r}^{2}(k)|\\
 &\leq \lambda_r \lambda_{r}^{*} (|\sum \limits_{jk} \left(\delta_{ijk} \bar{\bc}_{r}^{*}(j)\bar{\cc}_{r}^{*}(k)\bc_{r}(j)\cc_{r}(k)-  \delta_{ijk} \bc_{r}^{2}(j)\cc_{r}^{2}(k) \right)| + \Delta_{\lambda_r} \sum \limits_{jk} \delta_{ijk} \bc_{r}^{2}(j)\cc_{r}^{2}(k)) \\
 & \leq |\sum \limits_{jk}\delta_{ijk} {\bc}_{r}^{*}(j)\dc_{c_r}^{*}(k)\bc_{r}(j)\cc_{r}(k)|+ |\sum \limits_{jk}\delta_{ijk}\dc_{b_r}^{*}(j){\cc}_{r}^{*}(k)\bc_{r}(j)\cc_{r}(k)| \\
 &+ |\sum\limits_{jk}\delta_{ijk}\dc_{b_r}^{*}(j)\dc_{c_r}^{*}(k)\bc_{r}(j)\cc_{r}(k)|,
\end{align*}
where the last inequality is obtained using the triangle inequality and the fact that $\bc_{r}(j)= \bc_{r}^{*}(j)+ \dc_{b_r}(j)$ and $\cc_{r}(j)= \cc_{r}^{*}(j)+ \dc_{c_r}(j)$ combined with the fact that $F_b=supp({\bc}_{r}^*) \subseteq supp(\bar{\bc}_{r}^*)=F$ and $F_c=supp({\cc}_{r}^*) \subseteq supp(\bar{\cc}_{r}^*)=F$ which means that $\bar{\bc}_{r}^*(k)-{\bc}_{r}^*(k)= 0$ and $\bar{\cc}_{r}^*(k)-{\cc}_{r}^*(k)= 0$. Next applying the results of Lemma \ref{lem:lema2} and Lemma \ref{lem:lema5}, we get 
\begin{align*}
     I_{121}& \leq \lambda_{r}^{*}\lambda_{r} p\left(|\langle \bc_{r}^{*} {,}  \bc_{r}\rangle \langle\dc_{c_r} {,}  \cc_{r}\rangle| + |\langle \dc_{b_r} {,}  \bc_{r}\rangle \langle \cc_{r}^{*} {,}  \cc_{r}\rangle| + |\langle \dc_{b_r} {,}  \bc_{r}\rangle \langle \dc_{c_r} {,}  \cc_{r}\rangle| + \Delta_{\lambda_r}\right)\\
    & + p\gamma( \|\dc_{c_r}\|_2+ \|\dc_{b_r}\|_2 + \|\dc_{c_r}\|_2\|\dc_{b_r}\|_2+ \Delta_{\lambda_r})\\
    & \leq 8\lambda_{r}^{*}\lambda_{r}p \left( \max\limits_{\uc_r  \in \{c_r, b_r\}} \big\{ \sqrt{1-\frac{\|d_{u_r}\|_2}{2}}\|\dc_{u_r} \|_{2}^{2}, \|\dc_{u_r} \|_{2}^{4},\Delta_{\lambda_r}, \gamma \|\dc_{u_r} \|_{2},\gamma \Delta_{\lambda_r} \big\} \right ),\stepcounter{equation}\tag{\theequation}\label{eq:eq15}
\end{align*}
where the last inequality above holds with probability $ 1- 2d^{-10}$ provided the reveal probability $ p \geq \frac{C\mu ^3(1+\gamma/3) \log^2(d^{10}) }{d^{3/2}\gamma^2}$. Combining equations \eqref{eq:eq14} and  (\ref{eq:eq15}) followed by making use of lemma (\ref{lem:lema2}) to bound the denominator of $\|err_{1}\|_2$, we get
\begin{equation}
\label{eq:err1at}
   \|err_1\|_2 \leq  \frac{8\lambda_{r}^{*}\lambda_{r}p \left( \max\limits_{\uc_r  \in \{c_r, b_r\}} \big\{ \sqrt{1-\frac{\|d_{u_r}\|_2}{2}}\|\dc_{u_r} \|_{2}^{2}, \|\dc_{u_r} \|_{2}^{4},\Delta_{\lambda_r}, \gamma \|\dc_{u_r} \|_{2},\gamma \Delta_{\lambda_r} \big\} \right )+\sigma_r \sigma_{r}^{*} ( \frac{1}{2} \|d_v\|_2^2 +\Delta_{\sigma_r})}{ \lambda_{r}^2p(1-\gamma ) +\sigma_{r}^2}, 
\end{equation}
with probability $1-2d^{-9}$.\\
We now move on to bounding the expression $\|err_3\|_2$. 
\begin{align*}
    \|err_3\|_2 &\leq \sigma_r \|\Dm^{-1} \sum\limits_{m\in [R] \setminus r} (\sigma_{m}^{*}\Jm \bar{\ac}_{m}^{*}- \sigma_{m}\Pm \bar{\ac}_{m})\|_2\\
    & \leq \sigma_r\max \limits_{i} |\Dm^{-1}_{ii}| \sum\limits_{m\in [R] \setminus r}\|\sigma_{m}^{*}\langle \bar{\vc}_{m}^{*} {,}  \vc_{r} \rangle \bar{\ac}_{m}^{*}- \sigma_{m}\langle \bar{\vc}_{m} {,}  \vc_{r} \rangle \bar{\ac}_{m}\|_2\\
    & \leq  \sigma_r\max \limits_{i} |\Dm^{-1}_{ii}| \sum\limits_{m\in [R] \setminus r} \sigma_{m}^{*} \left(|\langle \bar{\vc}_{m}^{*} {,}  \vc_{r} \rangle - \langle \bar{\vc}_{m} {,}  \vc_{r} \rangle| \|\bar{\ac}_{m}^{*}\|_2 +|\langle \bar{\vc}_{m} {,}  \vc_{r} \rangle| \|\dc_{a_m}\|_2\right) \\
    &+ \sigma_r\max \limits_{i} |\Dm^{-1}_{ii}| \sum\limits_{m\in [R] \setminus r} \sigma_{m}^{*} \left(\Delta_{\sigma_m}|\langle \bar{\vc}_{m} {,}  \vc_{r} \rangle |\|\bar{\ac}_{m}  \|_2\right), \stepcounter{equation}\tag{\theequation}\label{eq:eq17b}
\end{align*}
where for inequality three, we use the fact that $\|\langle \bar{\vc}_{m}^{*} {,}  \vc_{r} \rangle \bar{\ac}_{m}^{*}\|_2 \leq \|\langle \vc_{m}^{*} {,}  \vc_{r} \rangle \ac_{m}^{*}\|_2$ since $\|\bar{\ac}_{m}^{*}\|_2 \leq 1$ and that the truncation process is invariant to scaling. We also used the fact that $\sigma_r =\sigma_r-\sigma_r^* +\sigma_r^*$.
Next, since $\{\textrm{supp}({\vc}_{m}^{*}), \textrm{supp}({\vc}_{m}) \}\subseteq F$ it follows that $\langle \bar{\vc}_{m}^{*} {,}  \vc_{r} \rangle - \langle \bar{\vc}_{m} {,}  \vc_{r} \rangle=\langle \dc_{\vc_m} {,} \vc_r \rangle $. Then noticing that $\langle \vc_{m} {,}  \vc_{r} \rangle \leq (\frac{c_0}{\sqrt {d}}+ 3\|\dc_{v_r}\|_2)$ and using the results of Lemma \ref{lem:lema2} to bound $\max \limits_{i} |\Dm^{-1}_{ii}|$ yields
\begin{equation}
\label{eq:err3at}
    \|err_3\|_2 \leq \frac{\sigma_r \sum\limits_{m\in [R] \setminus r}\sigma_{m}^{*}\left (\|\dc_{\vc_m}\|_2 + (\frac{c_0}{\sqrt{d}}+ 3\|\dc_{v_r}\|_2)\|\dc_{\ac_m}\|_2 + \Delta_{\sigma_m} (\frac{c_0}{\sqrt{d}}+ 3\|\dc_{v_r}\|_2) \right)}{\lambda_{r}^2p(1-\gamma ) +\sigma_{r}^2},
    \end{equation}
with probability  $ 1- 2d^{-9}$ provided the reveal probability $ p \geq \frac{C\mu ^3(1+\gamma/3) \log^2(d^{10}) }{d^{3/2}\gamma^2}$.

Next we bound the expression  $\|err_2\|_2$ as
\begin{align*}
    \|err_2\|_2 &= \| \lambda_r \Dm^{-1} \sum\limits_{m \in [R] \setminus r} \lambda_{m}^{*}(\Fm \bar{\ac}_{m}^{*} - \Gm \bar{\ac}_{m}+ \Delta_{\lambda_{m}}\Gm \bar{\ac}_{m})\|_2\\
    &\leq  \lambda_r \|\Dm^{-1}\|_2 \sum\limits_{m \in [R] \setminus r}  \lambda_{m}^{*} \left(\|(\Fm - \Gm) \bar{\ac}_{m}^{*}\|_2+ \|\Gm \dc_{a_m}\|_2 + \|\Delta_{\lambda_{m}}\Gm \bar{\ac}_{m}\|_2\right)\\
        &\leq  \lambda_r \|\Dm^{-1}\|_2 \sum\limits_{m \in [R] \setminus r}  \lambda_{m}^{*} \left(\underbrace{\max\limits_{i}|(\Fm - \Gm)_{ii}|}_{I_{21}}+  (\|\dc_{a_m}\|_2 + \Delta_{\lambda_{m}})\underbrace{\max\limits_{i}|\Gm_{ii}|}_{I_{22}}\right), \stepcounter{equation}\tag{\theequation}\label{sup:er2}
\end{align*}
where the second inequality is due to the triangle inequality and the third inequality is due to the fact that $\|\bar{\ac}_m^*\|_2 \leq \|\ac_m^*\|_2=1$ and $\|\bar{\ac}_m\|_2 \leq \|\ac_m\|_2 =1$ as well as the fact that the matrices $\|\Fm - \Gm\|_2$ and $\|\Gm\|_2$ are diagonal matrices hence there spectral norm is their maximum absolute diagonal value.
We focus on bounding bounding ${I_{21}}$ and ${I_{22}}$ next. 
\begin{align*}
{I_{21}}&= |\sum\limits_{jk}\delta_{ijk} \bar{\cc}_{m}^{*}(k)\bar{\bc}_{m}^{*}(j)\cc_{r}(k)\bc_{r}(j) - \sum\limits_{jk}\delta_{ijk}\bar{\cc}_{m}(k) \bar{\bc}_{m}(j) \cc_{r}(k)\bc_{r}(j)|\\
& \leq |\sum\limits_{jk}\delta_{ijk} \dc_{c_m}(k)\bar{\bc}_{m}^{*}(j)\cc_{r}(k)\bc_{r}(j)| + |\sum\limits_{jk}\delta_{ijk}\bar{\cc}_{m}^{*}(k) \dc_{b_m}(j) \cc_{r}(k)\bc_{r}(j)|\\
& + | \sum\limits_{jk}\delta_{ijk} \dc_{c_m}(k)\dc_{b_m}(j)\cc_{r}(k)\bc_{r}(j)|\\
& \leq p \left( |\langle \dc_{c_m}{,} \cc_{r} \rangle \langle  \bar{\bc}_{m}^{*}  {,} \bc_{r} \rangle |+ |\langle \bar{\cc}_{m}^{*} {,} \cc_{r} \rangle \langle  \dc_{b_m} {,} \bc_{r} \rangle| +  |\langle \dc_{c_m}{,} \cc_{r} \rangle \langle  \dc_{b_m}  {,} \bc_{r} \rangle| \right )\\
& + \gamma(\|\dc_{c_m} \|_2 + \|\dc_{b_m} \|_2 + \|\dc_{c_m} \|_2 \|\dc_{b_m} \|_2)\\
&\leq 6p  \max\limits_{\uc \in \{\cc_m,\bc_m,\cc_r,\bc_r, \}}\left( (\frac{c_0}{\sqrt{d}}+\|\dc_{u} \|_2),\|\dc_{u} \|_2, \gamma  \right)\|\dc_{u} \|_2 \stepcounter{equation}\tag{\theequation}\label{sup:eqi21}.
\end{align*}
The last inequality above holds with probability $ 1- 2d^{-10}$ provided the reveal probability $ p \geq \frac{C\mu ^3(1+\gamma/3) \log^2(d^{10}) }{d^{3/2}\gamma^2}$. The third inequality is due to Lemma \ref{lem:lema3} by noting that since $\textrm{supp}({\bc}_{m}^{*}) \subseteq F_b$ then $\bar{\bc}_{m}^{*}(j) \leq \frac{\mu}{\sqrt{d}} $. Similarly using Lemma  \ref{lem:lema4}, and applying the union bound and the fact that $
|\langle\cc_{m} {,}  \cc_{r}\rangle\langle \bc_{m} {,}  \bc_{r}\rangle | \leq  \max \{\langle\cc_{m} {,}  \cc_{r}\rangle^2, \langle \bc_{m} {,}  \bc_{r} \rangle^2 \} $, $\leq \left(\frac{c_0}{\sqrt{d}}+ \max\limits_{\uc_r \in \{\cc_r , \bc_r\}}3\|\dc_{u_r}\|_2\right)^2 $
 yields the following inequality
\begin{align}
\label{sup:eqi22}
   {I_{22}} \leq p \max\limits_{\uc_r \in \{\cc_r , \bc_r,\cc_m , \bc_m\}} \left((\frac{c_0}{\sqrt{d}}+ 3\|\dc_{u_r}\|_2)^2,
    \gamma\right),
\end{align}
with probability $ 1- 2d^{-9}$.\\\
Putting equations \eqref{sup:er2}, \eqref{sup:eqi21},\eqref{sup:eqi22}, and Lemma \ref{lem:lema2} together yields
\begin{equation}
\label{eq:err2at}
    \|err_2\|_2 \leq \frac{ \lambda_r8p  \sum\limits_{m \in [R] \setminus r}\lambda_m^{*} \max\limits_{\uc \in \{\ac_m,\bc_m,\ac_r,\bc_r, \}}\left( (\frac{c_0}{\sqrt{d}}+\|\dc_{u} \|_2),(\frac{c_0}{\sqrt{d}}+3\|\dc_{u} \|_2)^2,\|\dc_{u} \|_2, \gamma  \right)\|\dc_{u} \|_2}{\lambda_r^2 p(1-\gamma)+\sigma_r^2},
\end{equation}
with probability $1-2d^{-9}$ provided $ p \geq \frac{C\mu ^3(1+\gamma/3) \log^2(d^{10}) }{d^{3/2}\gamma^2}$.\\
Next, we bound the error matrix and error matrix through $\|err_4\|_2$ which is bounded by applying Lemma\ref{lem:lemmatenspec}, Lemma \ref{lem:lema2} and the fact that $\|\mathcal{E}_M \vc_r\|_2 \leq \|\mathcal{E}_M\|$ since $\|\vc_r\|_2$=1 and by definition $\|\mathcal{E}_M\|=\sup_{\substack{\| \uc \| = 1}} \|{\Er_M \uc }\|_2$. Following a similar proof of $(\ref{supeq:errc3})$, we have, 
\begin{equation}
\label{eq:err4at}
\|err_4\|_2 \leq \frac{\lambda_r \sqrt{p}(1+\gamma)\| \Er_T \|_{<d+s>}+ \sigma_r \| \Er_M \|_{<d+s>} }{\lambda_r^2p(1-\gamma)+\sigma_r^2},
\end{equation}
with probability $1-2d^{-9}$ provided $ p \geq \frac{C\mu^4 (1+\gamma/3) \log^2(d^{10}) }{d^2\gamma^2}$. Combining the error bounds results of $\|err_1\|_2$, $\|err_3\|_2$, $\|err_2\|_2$, $\|err_4\|_2$  in equations \eqref{eq:err1at}, \eqref{eq:err2at}, \eqref{eq:err3at} and \eqref{eq:err4at}, lettings $\|\dc_{\uc}\|_2 =\epsilon_T$, for $\uc \in\{\ac_r,\bc_r,\cc_r \}$, $\|\dc_v\|_2= \epsilon_M$, $\Delta_{\lambda_r}= \frac{\epsilon_T}{\lambda_r^*}$ and  $\Delta_{\sigma_r}= \frac{\epsilon_M}{\sigma_r^*}$ $\forall r\in[R]$ 
and using the fact that $ \lambda_{r}^{*}-\epsilon_T\leq  \lambda_{r}\leq \lambda_{r}^{*}+\epsilon_T $ and  $ \sigma_{r}^{*}-\epsilon_T\leq  \sigma_{r}\leq \lambda_{r}^{*}+\epsilon_T $ for all $r \in [R]$, yields
\begin{align*}
    &\|\dot{\ac}_r-\ac_{r}^{*} \|_2 \\
    &\leq  \frac{8pR \lambda_{max}^{*} (\lambda_{r}^{*}+\epsilon_T) \max\limits_{\uc \in \{\cc_m,\bc_m,\cc_r,\bc_r, \}}\left(  \sqrt{1-\frac{\epsilon_T}{2}}\epsilon_T, (\frac{c_0}{\sqrt{d}}+\epsilon_T),(\frac{c_0}{\sqrt{d}}+3\epsilon_T)^2,\epsilon_T, \gamma, 1/\lambda_{min}^{*} \right)\epsilon_T }{(\lambda_{min}^{*}-\epsilon_T)^2 p(1-\gamma)+(\sigma_{min}^{*}-\epsilon_M)^2} \\
    & +\frac{3R\sigma_{max}(\sigma_{r}^{*}+\epsilon_M)\max \left( \epsilon_M,1/\sigma_{min}^*,\frac{c_0}{\sqrt{d}}+3\epsilon_M \right) \epsilon_M}{(\lambda_{min}^{*}-\epsilon_T)^2 p(1-\gamma)+(\sigma_{min}^{*}-\epsilon_M)^2}\\ &+\frac{(\lambda_{r}^{*}+\epsilon_T)\sqrt{p}(1+\gamma)\| \Er_T \|_{<d+s>}+(\sigma_{r}^{*}+\epsilon_T)\| \Er_M \|_{<d+s>}}{(\lambda_{min}^{*}-\epsilon_T)^2 p(1-\gamma)+(\sigma_{min}^{*}-\epsilon_M)^2},
     \stepcounter{equation}\tag{\theequation}\label{sup:fat}
\end{align*}
with probability $1-2d^{-9}$. Simplifying the expression completes the proof for step 1 of the Lemma \ref{cor:contraas}.

\textbf{Step2}: We now get an upper bound for $\|
\ac_r-\dot{\ac}_r\|_2$. Note that 
$$
\|\ac_r-\dot{\ac}_r\|_2  = \|
\frac{\dot{\ac}_r}{\|\dot{\ac}_r\|_2}-\dot{\ac}_r\|_2 = \|\frac{\dot{\ac}_r}{\|\dot{\ac}_r\|}\|_2 |1-\|\dot{\ac}_r \|_2| = |1-\|\dot{\ac}_r \|_2|.$$ Hence bounding $\|\ac_r-\dot{\ac}_r\|_2$ simplifies to bounding $|1-\|\dot{\ac}_r \|_2|$.Using the expression of $\dot{\ac}_{r}$ in \eqref{sup:adot} and applying the triangle inequality we get,
\begin{align*}
|1-\|\dot{\ac}_{r}\|_2| & \leq \underbrace{|1- \|\lambda_r \Dm^{-1} \lambda_{r}^{*}\Em \bar{\ac}_{r}^{*} + \sigma_r \Dm^{-1} \sigma_{r}^{*} \Hm \bar{\ac}_{r}^{*} \|_2|}_{I} +\underbrace{\|\lambda_r \Dm^{-1} \sum\limits_{m \in [R] \setminus r} (\lambda_{m}^{*}\Fm \bar{\ac}_{m}^{*} - \lambda_{m}\Gm \bar{\ac}_{m})\|_2}_{II} \\
& +\underbrace{\| \sigma_r \Dm^{-1} \sum\limits_{m\in [R] \setminus r} (\sigma_{m}^{*}\Jm \bar{\ac}_{m}^{*}- \sigma_{m}\Pm \bar{\ac}_{m})\|_2}_{III} +\underbrace{\|P_{\Omega}(\mathcal{E}_{T_{F}}) \times_2 \bc_r \times_3 \cc_r +\mathcal{E}_{M_{F'}}\vc_r \|_2}_{IV}.
\stepcounter{equation}\tag{\theequation}\label{sup:adotminus1}
\end{align*}
Bounds for elements $(II)$ $(III)$ and $(IV)$ in the equation above are derived in \eqref{eq:err2at}, \eqref{eq:err3at} and \eqref{eq:err4at} respectively. Hence we only focus on bounding elements $(I)$.
\begin{align*}
I & = |\|{\ac}_{r}^{*}\|_2 - \|\lambda_r \Dm^{-1} \lambda_{r}^{*}\Em \bar{\ac}_{r}^{*} + \sigma_r \Dm^{-1} \sigma_{r}^{*} \Hm \bar{\ac}_{r}^{*} \|_2|\\
& \leq \|{\ac}_{r}^{*} - \Dm^{-1}(\lambda_r  \lambda_{r}^{*}\Em  + \sigma_r \sigma_{r}^{*} \Hm) \bar{\ac}_{r}^{*} \|_2\\
&=\| \Dm^{-1}\left(\lambda_r \lambda_r^{*} \Em + \sigma_r \sigma_r^{*}\Hm -\Dm\I\right) \bar{\ac}_{r}^{*}\|_2\\
& =\|err_1\|_2, \stepcounter{equation}\tag{\theequation}\label{sup:step22}
\end{align*}
where $err_1$ is the error component defined in \eqref{sup:adot2} and bounded in \eqref{eq:err1at}.
The first equality is obtained by using the fact that $\|{\ac}_{r}^{*}\|_2=1$, vector norm property is then use to get the first inequality and finally second equality is due to  $\ac_r^{*} =\Dm^{-1}\Dm \ac_r^{*} $ and the fact that $\bar{\ac}_{r}^*= {\ac}_{r}^*$ since $F_a=supp({\ac}_{r}^*) \subseteq supp(\bar{\ac}_{r}^*)=F$.
Hence combining above results yields,
\begin{align*}
\|\ac_r-\dot{\ac}_r\|_2  & \leq I + II + III + III + IV \\
& \leq \|\dot{\ac}_r-\ac_{r}^{*} \|_2,    \stepcounter{equation}\tag{\theequation}\label{sup:step23}
\end{align*}
which ends step 2 of the proof. The proof of Lemma \ref{cor:contraas} is completed by combining results of step 1 and step 2 which shows that $\|
\ac_r-\ac_{r}^{*} \|_2 \leq 2\|\dot{\ac}_r-\ac_{r}^{*} \|_2,$ and taking the maximum over all $r$.\eop

%%%%%%%%%%%%%%%%%%%%%%%%%%%%%%%%%%%%%%%%%%%%%%%%%%%%%%%%%%%%%%%%%%%
%%%%%%%%%%%%%%%%%%%%%%%%%%%%%%%%%%%%%%%%%%%%%%%%%%%%%%%%%%%%%%%%%%%%%%%%
\subsection{Auxillary Lemmas} 
\label{sup:aulemma}
\begin{lemma}
\label{lem:lema2}
Let $\uc$ and $\wc$ be unit vectors in $\mathbb{R}^n $ such that $|\uc(i)| \leq \frac{\mu}{\sqrt d}$ and $|\wc(j)| \leq \frac{\beta}{\sqrt d}$. Also let $\delta_{i,j,k}$ be $i.i.d.$ Bernoulli random variables with $P(\delta_{ijk}=1)=p$ and  $ 1\leq i \leq n$, $ 1\leq j \leq n$, $ 1\leq k \leq n$. \\
Then provided $p \geq \frac{C\mu^2 \beta^2(1+\gamma/3) \log(d^{10}) }{d^2\gamma^2}$ we have
\[ \left|\ \sum\limits_{j,k}\delta_{ijk}\uc_{r}^{2}(j) \wc_{r}^{2}(k) - p\langle \uc {,} \uc \rangle \langle \wc {,} \wc \rangle   \right|\leq p\gamma,\]
with probability $1-d^{-10}$.
\end{lemma}
\paragraph{Proof:}
Let $X_{jk}= \frac{1}{p} \left(\delta_{ijk}\uc^{2}(j)\wc^{2}(k)-E(\delta_{ijk}\uc^{2}(j)\wc^{2}(k))\right)$. Using the bound on the elements of $\uc$ and $\wc$, we have $|X_{jk}|=|\frac{1}{p}(\delta_{ijk}-p)\uc^{2}(j)\wc^{2}(k)|\leq \frac{\mu^2\beta^2}{pd^2} $.
Also 
$$\sum\limits_{j,k}E[X_{jk}^{2}]=\frac{1}{p}(1-p)\sum\limits_{j,k}\uc_{r}^{4}(j)\wc_{r}^{4}(k)\leq \frac{\mu^2 \beta^2}{pd^2}.$$
Applying Bernstein tail bound inequality we get: 
$$P\left(| \sum\limits_{j,k}\delta_{ijk}\uc_{r}^{2}(j) \wc_{r}^{2}(k)-p\langle \uc {,} \uc \rangle \langle \wc {,} \wc \rangle |\geq pt \right )\leq \exp{(\frac{-d^2p t^2/2}{\mu^2 \beta^2(1+\frac{1}{3}t)})}. $$
Setting the right side of the inequality to be less than $q$ yields:
$$P\left(| \sum\limits_{j,k}\delta_{ijk}\uc_{r}^{2}(j) \wc_{r}^{2}(k) -p\langle \uc {,} \uc \rangle \langle \wc {,} \wc \rangle |\leq  p\gamma \right ) \geq 1-q, $$
for $ p \geq \frac{\mu^2 \beta^2(1+\gamma/3) \log(1/q) }{d^2\gamma^2}$.
Choosing $q \leq d^{-10}$ completes the proof of Lemma \ref{lem:lema2}. \eop

\begin{lemma}
\label{lem:lema3}
Let $\uc^{*}$, $\uc$ and $\wc$ be unit vectors in $\mathbb{R}^n $ such that $|\uc^{*}_i| \leq \frac{\mu}{\sqrt d}$,  $|\uc|$ and $ |\wc| \leq \frac{\beta}{\sqrt d}$. Let $\dc$ be another vector with $\|\dc \|_2 \leq 1$. Also let $\delta_{i,j,k}$ be $i.i.d.$ Bernoulli random variables with $P(\delta_{ijk}=1)=p$ and $ 1\leq i \leq n$, $ 1\leq j \leq n$, $ 1\leq k \leq n$.
Provided $p \geq \frac{C\mu \beta^2(1+\gamma/3) \log^2(\frac{1}{2}d^{10}) }{d^{3/2}\gamma^2}$, with probability greater than $ 1-2d^{-10}$, we have
\[ |\ \sum\limits_{j,k}\delta_{ijk}\uc^{*}(j) \dc(k) \uc(j) \wc(k)-p \langle \uc^{*} {,} \uc \rangle  \langle \dc {,} \wc \rangle |\leq p\gamma\|\dc\|_2. \]
\end{lemma}

\paragraph{Proof:} 
Let $X_{jk}= \frac{1}{p} \left(\delta_{ijk}\uc^{*}(j)\dc(k)\uc(j)\wc(k)-E(\delta_{ijk}\uc^{*}(j)\dc(k)\uc(j)\wc(k))\right)$. Then we have That is $ |X_{jk}|=\frac{1}{p}(\delta_{ijk}-p)\uc^{*}(j)\dc(k)\uc(j)\wc(k)\leq \frac{1}{p}(1-p)\frac{\mu \beta^2}{d^{3/2}}\|\dc\|_2$. Also,
$$\sum\limits_{j,k}E[X_{jk}^2]=\frac{1}{p}\sum\limits_{j,k}(\uc(j)^2 \dc(k)^2 \uc(j)^2 \wc(k)^2)\leq\frac{\mu \beta^2 \|d\|_{2}^{2}}{pd^{3/2}}.$$

Applying Bernstein tail bound inequality we get: 
\begin{equation}
P\left(|\sum\limits_{j,k}\delta_{ijk}\uc^{*}(j)\dc(k)\uc(j)\wc(k)-p \langle \uc^{*} {,} \uc \rangle  \langle \dc {,} \wc \rangle |\geq pt \right )\leq 2 \exp{(\frac{-d^{3/2}p t^2}{\mu \beta^2 \|\dc\|_2(\|\dc\|_2+\frac{1}{3}t)})}.\\
\end{equation}
Setting the right side of the inequality to be less than $q$ and choosing $t \leq \gamma \|\dc\|_2 $ then solving for $p$ yields:
$$P \left(|\sum\limits_{j,k}\delta_{ijk}\uc^{*}(j)\dc(k)\uc(j)\wc(k)-p \langle \uc^{*} {,} \uc \rangle  \langle \dc {,} \wc \rangle |\leq p\gamma\|b\|_2 \right ) \geq 1-2q,$$ 
for $ p \geq \frac{\mu \beta^2(1+\gamma/3) \log(\frac{1}{q}) }{d^{3/2}\gamma^2}$.
Choosing $q \leq d^{-10}$ completes the proof of Lemma \ref{lem:lema3}.\eop

\begin{lemma}
\label{lem:lema4}
Let $\uc^{*}$, $\wc^{*}$, $\uc$ and  $\wc$ be unit vectors in $\mathbb{R}^n $ such that $|\uc^{*}(i)|$ and $|\wc^{*}(j)| \leq \frac{\mu}{\sqrt d}$,  $|\uc_{i}|$ and $ |\wc_{i}| \leq \frac{\beta}{\sqrt d}$. Let $\delta_{i,j,k}$ be $i.i.d.$ Bernoulli random variables with $P(\delta_{ijk}=1)=p$ and $ 1\leq i,j,k \leq n$. Provided $p \geq \frac{C\mu^2 \beta^2(1+\gamma/3) \log(\frac{1}{2}d^{10}) }{d^{2}\gamma^2}$, with probability greater than $ 1-2d^{-10}$, we have
\[ | \sum\limits_{j,k}\delta_{ijk}\uc^{*}(j) \wc^{*}(k) \uc(j) \wc(k)| \leq p|\langle \uc^{*} {,} \uc \rangle  \langle \wc^{*} \wc \rangle |+ p\gamma.\]
\end{lemma}

\paragraph{Proof:} 
Let $X_{jk}= \frac{1}{p} \left(\delta_{ijk}\uc^{*}(j)\wc^{*}(k)\uc(j)\wc(k)-E(\delta_{ijk}\uc^{*}(j)\wc^{*}(k)\uc(j)\wc(k))\right)$. Then we have $ |X_{jk}|=\frac{1}{p}(\delta_{ijk}-p)\uc^{*}(j)\wc^{*}(k)\uc(j)\wc(k)\leq \frac{1}{p}(1-p)\frac{\mu^2 \beta^2}{d^2}$.Also 
$$\sum\limits_{j,k}E[X_{jk}^2]=\frac{1}{p}(1-p)\sum\limits_{j,k}(\uc(j)^{2*} \wc(k)^{*2} \uc(j)^2 \wc(k)^2)\leq \frac{1}{p}(1-p)\frac{\mu^2 \beta^2}{d^2}.$$
Applying Bernstein tail bound inequality we get: 
$$P\left(|\sum\limits_{j,k}\delta_{ijk}\uc^{*}(j)\dc(k)\uc(j)\wc(k)-p \langle \uc^{*} {,} \uc \rangle  \langle \dc {,} \wc \rangle |\geq pt \right )\leq 2 \exp{(\frac{-d^2p t^2}{\mu^2 \beta^2 (1-p)(1+\frac{1}{3}t)})}. $$
Setting the right side of the inequality to be less than $q$ and choosing $t \leq \gamma $ then solving for $p$ yields:
$$P \left( | \sum\limits_{j,k}\delta_{ijk}\uc^{*}(j) \wc^{*}(k) \uc(j) \wc(k)- \langle \uc^{*} {,} \uc \rangle  \langle \wc^{*} \wc \rangle |\leq  p\gamma \right )\geq 1-2q,$$ and $ p \geq \frac{\mu^2 \beta^2(1+\gamma/3) \log(\frac{1}{q}) }{d^{2}\gamma^2}$.
Letting $q \leq d^{-10}$ completes the proof of Lemma  \ref{lem:lema4}.\eop

\begin{lemma}
\label{lem:lema5}
Let $ \lambda_r$ be the update of the $\text{r}^{th}$ weight of the tensor after one iteration of Algorithm \ref{alg:jALS} and let $\lambda_{r}^{*}$ be the true $\text{r}^{th}$ weight of the tensor decomposition in the dense tensor and dense matrix case. Let $\tilde\cc$ be as defined in \eqref{eq:unorm_update} and $\cc$ as defined in \eqref{subeq:updatebc} then with probability greater than  $1-2n^{-9}$ we have
\[|\lambda_r -\lambda_r^{*}| \leq \|\tilde{\cc_r} - \lambda_r^{*}\cc_r^{*} \|_2.  \]
\end{lemma}

\paragraph{Proof:} 
We know that $\|{\cc_r^{*}}\|_2  = \|\cc_r\|_2=1$ hence we can write,
\begin{align*}
    |\lambda_r -\lambda_r^{*}| & =|\|\lambda_r{\cc_r}\|_2-\|\lambda_r^{*}{\cc_r}^{*}\|_2| \\
    & \leq \|\lambda_r{\cc_r}-\lambda_r^{*}{\cc_r}^{*}\|_2\\
    & = \|\tilde{\cc_r} -\lambda_r^{*}{\cc_r}^{*}\|_2
\end{align*}
The last equality above is obtained by observing that $\tilde{\cc_r} =\lambda_r{\cc_r} $ as shown in the proof of Lemma\ref{lem:update}. This complete the proof of the Lemma. 
Notice that the above Lemma can also be applied on $\sigma_r$ to obtain $|\sigma_r -\sigma_r^{*}| \leq \|\tilde{\vc_r} - \sigma_r^{*}\vc_r^{*} \|_2$. \eop

\begin{lemma}
\label{aulem:normunnorm}
Let $\tilde\cc$ be as defined in \eqref{eq:unorm_update} and $\cc$ as defined in \eqref{subeq:updatebc}. Also let $ \lambda_r$ be the update of the $\text{r}^{th}$ weight of the tensor after one iteration of Algorithm \ref{alg:jALS} and let $\lambda_{r}^{*}$ be the true $\text{r}^{th}$ weight of the tensor decomposition in the dense tensor and dense matrix case. Then with probability greater than $1-2n^{-9}$ we have
 \begin{align*}
    \|\cc_r -\cc_r^{*}\|_2 \leq \frac{2}{\lambda_r^*} \|\tilde{\cc_r} - \lambda_r^{*}{\cc_r}^{*} \|_2, \\
 \|\cc_r -\cc_r^{*}\|_2 + \Delta_{\lambda_r} \leq \frac{3}{\lambda_r^*} \|\tilde{\cc_r} -\lambda_r^{*}{\cc_r}^{*}\|_2,
 \end{align*}
where $\Delta_{\lambda_r}$ is as defined in  \eqref{supeq:relatdistance}.
\end{lemma}

\paragraph{Proof:} 
\begin{align*}
    \lambda_r^{*}\|\cc_r -\cc_r^{*}\|_2 & =  \|\lambda_r^{*}\cc_r -\lambda_r^{*}{\cc_r}^{*}\|_2 \\
    & =  \|\lambda_r\cc_r -\lambda_r^{*}{\cc_r}^{*} - \epsilon_{\lambda_r}\cc_r\|_2
    \leq \|\lambda_r\cc_r -\lambda_r^{*}{\cc_r}^{*} \|_2 + \| \epsilon_{\lambda_r}\cc_r\|_2\\
    & =  \|\tilde {\cc_r} -\lambda_r^{*}{\cc_r}^{*} \|_2 +  |\lambda_r -\lambda_r^{*}| \\
    & \leq 2 \|\tilde {\cc_r} -\lambda_r^{*}{\cc_r}^{*} \|_2,\stepcounter{equation}\tag{\theequation} \label{sup:bridge}\\
\end{align*}
which proves the first inequality of the Lemma. The proof of the second inequality in the lemma is obtained by combining \eqref{sup:bridge} with the results of Lemma \ref{lem:lema5}. \eop
%%%%%%%%%%%%%%%%%%
\begin{lemma}
\label{lem:lemmatenspec}
For any tensor $\Er_T \in \mathbb{R} ^{n \times n \times n}$ and any vectors $\uc$ and $\vc \in \mathbb{R}^n$ with \\ $\|\uc\|_2=\|\vc\|_2=1$, we have $$\|\Er_T  \times_1 \uc \times_2 \vc\|_2 \leq \|\Er_T\|,$$ where $\|\Er_T\|$ represents the spectral norm of the tensor defined in \eqref{eq:spectnorm}.
\end{lemma}

\paragraph{Proof:}
\begin{align*}
\|\Er_T  \times_1 \uc \times_2 \vc\|_2 &=\frac{\|\Er_T  \times_1 \uc \times_2 \vc\|_2^2 }{\|\Er_T  \times_1 \uc \times_2 \vc\|_2 }\\
& =\Big|\Er_T  \times_1 \uc \times_2 \vc \times_3 \left(\frac{\Er_T  \times_1 \uc \times_2 \vc}{\|\Er_T  \times_1 \uc \times_2 \vc\|_2}\right) \Big|\\
&\geq \sup_{\substack{\| \uc \| = \|\vc\| = \|\wc\| = 1}} \Big|{\Er_T \times_1 \uc \times_2 \vc \times_3 \wc }\Big|\\
&= \|\Er_T  \|.
\end{align*}
The first inequality is due to $\|\uc\|_2=\|\vc\|_2=1$ and the fact that $\frac{\Er_T  \times_1 \uc \times_2 \vc}{\|\Er_T  \times_1 \uc \times_2 \vc\|_2}=1$. The last equality is obtained by applying the definition of the tensor spectral norm provided in \eqref{eq:spectnorm}. \eop
%%%%%%%%%%%%%%%%%
\begin{lemma}
\label{aulem:inner}
Let $\uc$ and $\wc$ be unit vectors and let $\dc$ be a vector such that $\dc= \uc-\wc$ then
$$ |\langle \wc,\dc  \rangle |=\frac{1}{2}\|\dc\|_2^2. $$
\end{lemma}
\paragraph{Proof:}
Note that $\|\uc \|_2^2 =\sum\left(\wc(i)+\dc(i) \right)^2$. Hence given that $\uc$ is a unit vector we get 
\begin{align*}
 \sum\wc(i)^2 +2\sum\wc(i)\dc(i)+\sum\dc(i)^2&=1\\
  2\sum\wc(i)\dc(i)+\sum\dc(i)^2&=0\\
 2\sum\wc(i)\dc(i)&=-\sum\dc(i)^2\\ 
 |\langle\wc,\dc\rangle| &= \frac{1}{2}\|\dc\|_2^2,
\end{align*}
Which completes the proof of the lemma. \eop

\begin{lemma}
\label{aulem:supportvec}
Let $\uc$ and $\wc$ be unit vectors define $F_1 := \textrm{supp}(\uc)$, $F_2:= \textrm{supp}(\wc)$ be the support sets for $\uc$ and $\wc$ respectively with $F_i \subseteq \{1,\cdots d\}$ and $F:= F_u \cup F_w$ be the union of the two vectors' support sets. Let $\bar{\uc}:=\textrm{Truncate}(\uc, F)$ then it follows that 
$$ \langle \bar{\uc}, \wc \rangle =\langle {\uc}, \wc \rangle.$$
\end{lemma}
\paragraph{Proof:}
Since by definition, $\bar{\uc}:=\textrm{Truncate}(\uc, F)$, then we can write $ \langle \bar{\uc}, \wc \rangle$ explicitly as $\langle \bar{\uc}, \wc \rangle=\sum\limits_{i \in [d]}\bar{\uc}(i)\wc(i)$. Since $\bar{\uc}(i) \neq 0$ only when $i \in F_1$ and $i \in F_2$, we get $\sum\limits_{i \in [d]}\bar{\uc}(i)\wc(i)=\sum\limits_{i \in F}{\uc}(i)\wc(i)$. However, we know that $\textrm{supp}(\wc)=F_2 \subseteq F$ hence we get $$\langle \bar{\uc}, \wc \rangle=\sum\limits_{i \in F}{\uc}(i)\wc(i)= \sum\limits_{i \in [d]}{\uc}(i)\wc(i)=\langle \uc, \wc \rangle. $$ \eop

%\change{
%\begin{lemma}
%\label{lem:lemma_TS14}
%\cite[Theorem 1]{tomioka2014spectral}
%For an order-K tensor ${\cal X} \in \mathbb R^{n_1\times n_2 \times \cdots \times n_K}$ whose entries are i.i.d. sub-Gaussian random variables with zero-mean and variance proxy $\sigma^2$, its spectral norm is bounded as
%$$
%\| {\cal X} \| \le \sqrt{8 \sigma^2 \left( (\sum_{k=1}^K n_k) \log(2K/K_0) + \log(2/\delta) \right)},
%$$
%with probability at least $1-\delta$ and $K_0 = log(3/2)$.
%\end{lemma}
%}

%\change{
%\begin{lemma}
%\label{lem:lemma_sparse_spectral_norm_bound}
%For an error tensor ${\cal X} \in \mathbb R^{n_1\times n_2 \times n_3}$ whose entries are i.i.d. sub-Gaussian random variables with zero-mean and variance proxy $\sigma^2$, and a sparsity parameter $s_i \le n_i$, its sparse spectral norm is bounded as
%$$
%\| {\cal X} \|_{\langle s \rangle} = C
%$$
%with probability at least $1-n^{-10}$, where $n = \max\{n_1, n_2, n_3\}$ and $s = \max\{s_1, s_2, s_3\}$. 
%\end{lemma}
%}

\subsection{Additional Simulations} 
\label{sup:simulation}

The two additional simulations, we focus solely on the recovery accuracy of the shared and non-shared tensor components under our \texttt{COSTCO} to investigate the practical effect of component dimensions size and the rank on our algorithm. 

%%%%%%%%%%%%%%%%%%%%%%%%%%%%%%
%Tensor couple component size
%%%%%%%%%%%%%%%%%%%%%%%%%%%%%%%
\noindent{\textbf{Component Size}:} This part of the simulation considers the effect of varying the size of the coupled components $\Am^*$ of the true tensor on the tensor recovery.  We set the tensor missing entry percentage to be $ 90\% $; the noise level parameters are set to be $\eta_T = 0.001$ and $\eta_M = 0.001$ respectively and the sparsity level is kept at $60\%$. The complete simulation results are presented in Table \ref{tab:dim}. The tensor completion error improves with increasing size of the shared dimension since there is more information provided by the covariate matrix. With more and more information provided from the covariate matrix, the latent structure of the shared component dominates those of the non-shared components, making it easier to complete the whole tensor.

%%%%%%%%%%%%%%%%%%%%%%%%%%%%%%%
%%%% tensor COMPONENT SIZE TABLE
%%%%%%%%%%%%%%%%%%%%%%%%%%%%%%%

\begin{table}[h!]
\centering
\begin{small}
\caption{Estimation errors of \texttt{COSTCO} with varying coupled dimension $d_1$.}%; symbol $(\ddot{\Am})$ used to put shared tensor-matrix component $\Am$ in emphasis.}
\label{tab:dim}
\begin{tabular}{c|cccccc} \hline
 &  \multicolumn{6}{c}{Estimation Error} \\ 
 \cline{2-7}
Coupled Dimension $d_1$  & $\Ten$  & Comp $\ddot{\Am}$  & Comp ${\Bm}$  & Comp ${\Cm}$ & Comp ${\Dm}$ & $\bm{\lambda}$ \\
\hline
%	%\hspace{0.05in}\\
20	&	5.64e-05 &	1.77e-05	&	3.67e-05		&	3.51e-05	&	3.68e-05	&	1.60e-06	\\
	&	(1.24e-11) &	(6.09e-12)	&	(1.41e-11)		&	(1.88e-11)	&	(2.20e-11)	&	(6.09e-13)	\\
	%\hspace{0.05in}\\
	  \cline{2-7}
	%\hspace{0.05in}\\
%30	&	4.52e-05 	&	1.74e-05	&	3.13e-05	&	3.07e-05	&	3.16e-05	&	1.59e-06	\\
%	&	(4.13e-12)	&	(5.63e-12)	&	(7.83e-12)	&	(1.01e-11)	&	(6.19e-12)	&	(5.11e-13)	\\
	%\hspace{0.05in}\\
%	  \cline{2-7}
	%\hspace{0.05in}\\	
50	&	3.71e-05	&	1.72e-05	&	2.35e-05	&	2.39e-05	&	2.44e-05	&	1.25e-06	\\
	&	(3.29e-12)&	(2.66e-12)	&	(2.59e-12)		&	(4.06e-12)	&	(4.72e-12)	&	(5.14e-13)	\\
	%\hspace{0.05in}\\
	  \cline{2-7}
	%\hspace{0.05in}\\
100	&	2.66e-05 &	1.73e-05 &	1.72e-05		&	1.76e-05	&	1.77e-05	&	7.65e-07	\\
	&	(1.43e-12) &	(5.69e-13)	&	(2.86e-12)		&	(3.50e-12)	&	(1.96e-12)	&	(1.34e-13)	\\
		%\hspace{0.05in}\\
\hline
\end{tabular}
\end{small}
\end{table}

%%%%%%%%%%%%%%%%%%%%%%%%%%%%%%%
\noindent\textbf{Rank:} In this case we investigate the impact of the rank of the tensor and matrix on the tensor recovery performance of our \texttt{COSTCO} algorithm. We set the missing percentage of the tensor to $90\%$, the sparsity to be $60\%$ and the tensor and matrix noise levels $\eta_T$ and $\eta_M$ to be both $0.001$. We still tune the rank and cardinality using the procedure in Section \ref{sup:tuning}. As shown in Table \ref{tab:rank}, the recovery error is an increasing function of the tensor rank. It is well documented that the noisy tensor completion problem in general gets harder as the rank increases \citep{Song2019}. %This result also aligns well with the theoretical derivation provided in section \ref{sec:analtheo}. %In assumption 6, we see that the initialization error is a decreasing function of the rank $R$. Hence tensor with larger $R$ requires the initialization algorithm to be more accurate than tensors with smaller ranks.

%%%%%%%%%%%%%%%%%%%%%%%%%%%%%%%
%%%% tensor Rank TABLE
%%%%%%%%%%%%%%%%%%%%%%%%%%%%%%%

\begin{table}[h!]
\centering
\begin{small}
\caption{Estimation errors of \texttt{COSTCO} with varying rank.}%; symbol $(\ddot{\Am})$ used to put shared tensor-matrix component $\Am$ in emphasis.}
\label{tab:rank}
%\vskip 1 em
\begin{tabular}{c|cccccc} \hline
 &  \multicolumn{6}{c}{Estimation Error} \\ 
 \cline{2-7}
Tensor Rank  & $\Ten$  & Comp $\ddot{\Am}$  & Comp ${\Bm}$  & Comp ${\Cm}$ & Comp ${\Dm}$ & $\bm{\lambda}$ \\
\hline
	%\hspace{0.05in}\\
1	&	4.78e-05	&	2.76e-05	&	1.97e-06	&	2.77e-05	&	2.62e-05	&	5.31e-06	\\
	&	(1.34e-11)	&	(1.67e-11)	&	(6.72e-14)	&	(7.50e-12)	&	(1.38e-11)	&	(1.29e-11)	\\
	%\hspace{0.05in}\\
 \cline{2-7}
	%\hspace{0.05in}\\
2	&	6.50e-05	&	6.78e-05	&	1.39e-05	&	6.63e-05	&	6.66e-05	&	1.26e-05	\\
	&	(1.04e-11)	&	(6.82e-11)	&	(4.67e-11)	&	(5.07e-11)	&	(7.16e-11)	&	(3.76e-11)	\\
	%\hspace{0.05in}\\
 \cline{2-7}
	%\hspace{0.05in}\\
3	&	8.57e-05	&	7.82e-05	&	2.76e-05	&	7.99e-05	&	7.81e-05	&	1.32e-05	\\
	&	(2.52e-11)	&	(5.27e-11)	&	(1.11e-10)	&	(8.10e-11)	&	(5.97e-11)	&	(4.14e-11)	\\
	%	\hspace{0.05in}\\
\hline
\end{tabular}
\end{small}
\end{table}

\subsection{\change{Implementations of Covariate-assisted Neural Tensor Factorization}} 
\label{sec: sup_neural}

\change{
In Section \ref{sec:realdata}, we include a new competitive method, a covariate-assisted version of the neural tensor factorization \citep{wu2019neural}, and compare it with our \texttt{COSTCO} in the CTR prediction task. The original neural tensor factorization framework \citep{wu2019neural} takes a three-mode tensor as input and learns the latent embeddings for each mode of the tensor via a multi-layer perceptron (MLP). As a fair comparison, we implement a covariate-assisted neural tensor factorization method via Tensorflow. Specifically, user id, ad id, and device id are first converted to one-hot encodings, which are then fed into three parallel embedding layers. The concatenation of these and the covariates of the corresponding advertisement is then fed into a 3-layer perceptron to learn its representation, which is subsequently used as features to predict the associated CTR entries. 
 }
 
\change{    
Figure \ref{ifig:neuralmethod} demonstrates the recovery error of this covariate-assisted neural tensor factorization. For the structure of neural network, we fix the embedding dimension of device as $5$ and the hidden units of all layers of MLP as $20$ and consider cases $(d_1, d_2) \in  \{16,32,64,128,256,512 \} \times \{ 10,20,40,80\}$, where $d_1$ and $d_2$ denote the embedding dimension for user and advertisement, respectively. In our implementation, we have also varied the embedding dimensions of the device mode and the number of hidden units of MLP, and the prediction performance is robust to these parameters. We initialize all parameters of the neural network from $N(0, 0.1)$ and set the learning rate and the batch size as $0.005$ and $200$, respectively. As shown in Figure \ref{ifig:neuralmethod}, the best tensor recovery error of this new method is about $0.910$ and is stabilized even when the embedding dimensions are very large. This prediction performance is better than the baseline \texttt{tenALSsparse} whose recovery error is $1.083$, but is still inferior to our \texttt{COSTCO} whose recovery error is $0.825$.
}

\begin{figure}[h!]
\centering
\includegraphics[scale=0.6]{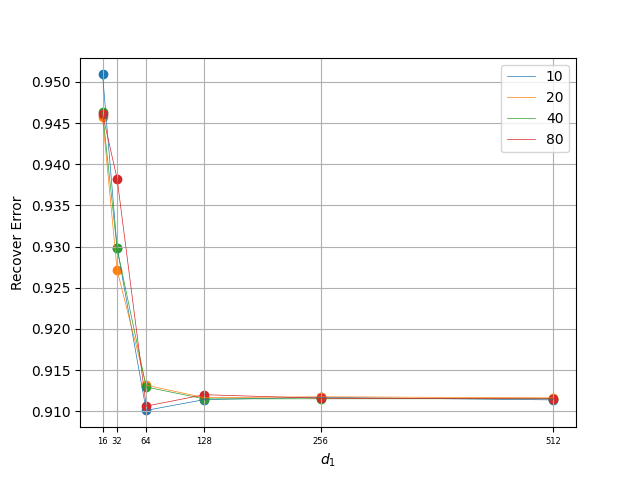}
        \caption{\change{The tensor recovery error of the covariate-assisted neural tensor factorization method. The X-axis $d_1$ refers to the embedding dimension for the user mode, and the four colorful lines refer to different embedding dimensions for the advertisement mode. Note that the tensor recovery error for our \texttt{COSTCO} is $0.825$, and the recovery error for \texttt{tenALSsparse} is $1.083$.}}
        \label{ifig:neuralmethod}
\end{figure}

\end{document}

%% file: def-smile.tex
\let\hat\widehat
\let\tilde\widetilde
\newcommand{\be}{\bm{e}}

\newcommand{\cS}{{\mathcal{S}}}

\newcommand{\blambda}{\bm{\lambda}}

\newcommand{\bsigma}{\bm{\sigma}}

\DeclareMathOperator{\ind}{\mathrm{1}}  % Indicator

\newtheorem{lemma}{{\bf Lemma}}
\newtheorem{theorem}{{\bf Theorem}}
\newtheorem{remark}{{\bf Remark}}
\def\eop
{\hfill $\Box$ \\

}